\documentclass[10pt,twocolumn,letterpaper]{article}
\pdfoutput=1

\usepackage{times}
\usepackage{epsfig}
\usepackage{graphicx}
\usepackage{amsmath}
\usepackage{amssymb}
\usepackage{algorithm}
\usepackage{algorithmic}

\usepackage{booktabs, multirow, multicol}
\usepackage{tabularx}

\usepackage{amsmath}
\usepackage{amssymb}
\usepackage{mathtools}
\usepackage{enumerate}
\usepackage{graphicx}

\newcommand{\half}{\frac{1}{2}}

\newcommand{\Prob}{\mathbb{P}}

\usepackage{latex/cvpr}
\cvprfinalcopy

\begin{document}

%%%%%%%%% TITLE
% \twocolumn[
\title{CProp: Adaptive Learning Rate Scaling from Past Gradient Conformity}

\author{
Konpat Preechakul\\
Chulalongkorn University\\
{\tt\small the.akita.ta@gmail.com}
\and
Boonserm Kijsirikul\\
Chulalongkorn University\\
{\tt\small boonserm.k@gmail.com}
}

\date{}
\maketitle
% ]
%\thispagestyle{empty}

%%%%%%%%% ABSTRACT
\begin{abstract}
Most optimizers including stochastic gradient descent (SGD) and its adaptive gradient derivatives face the same problem where an effective learning rate during the training is vastly different. A learning rate scheduling, mostly tuned by hand, is usually employed in practice. In this paper, we propose CProp, a gradient scaling method, which acts as a second-level learning rate adapting throughout the training process based on cues from past gradient conformity. When the past gradients agree on direction, CProp keeps the original learning rate. On the contrary, if the gradients do not agree on direction, CProp scales down the gradient proportionally to its uncertainty. Since it works by scaling, it could apply to any existing optimizer extending its learning rate scheduling capability. We put CProp to a series of tests showing significant gain in training speed on both SGD and adaptive gradient method like Adam. Codes are available at \url{https://github.com/phizaz/cprop}.
\end{abstract}
\begin{figure*}[t]
\centerline{
\includegraphics[width=0.36\textwidth]{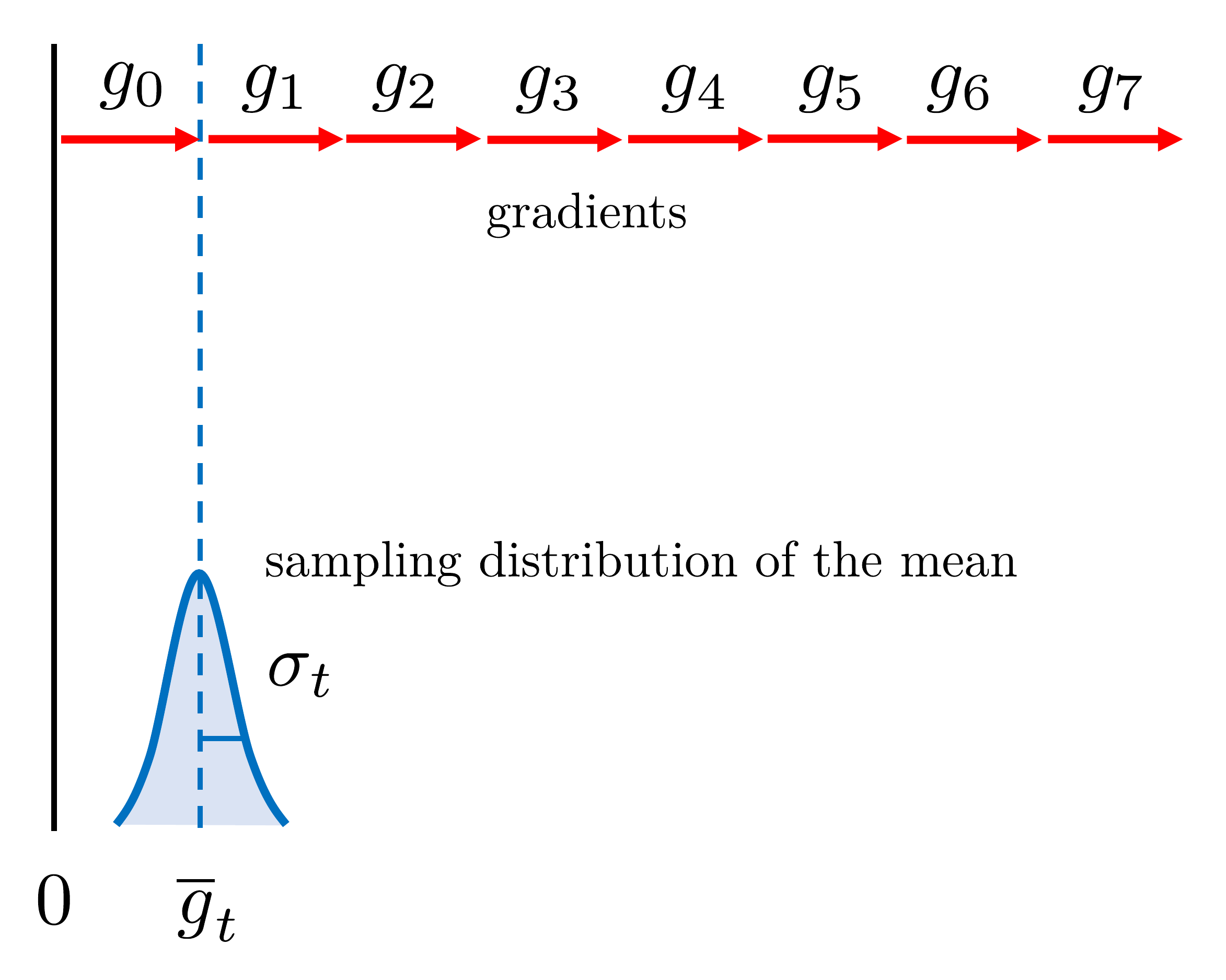}
\includegraphics[width=0.42\textwidth]{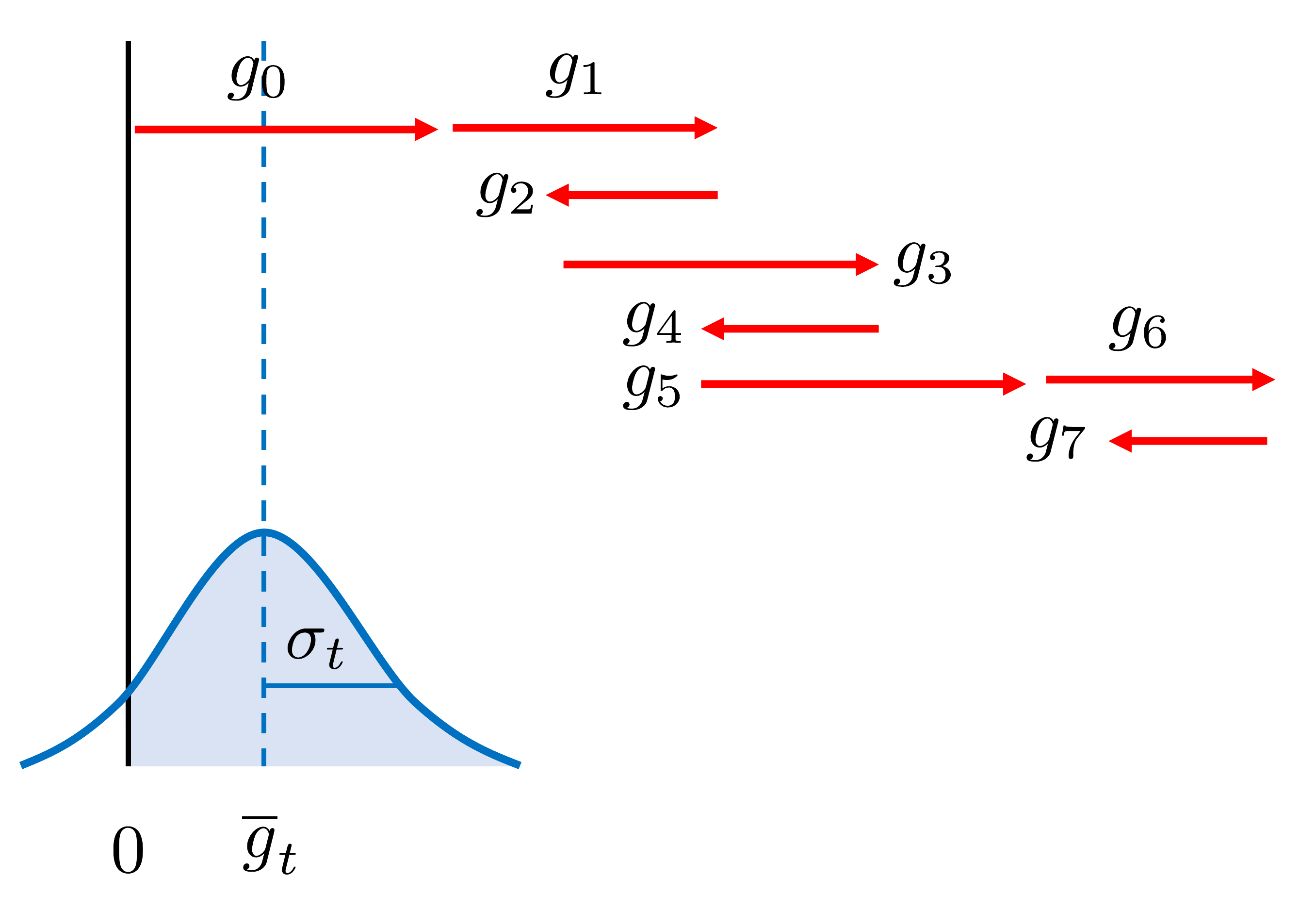}
\includegraphics[width=0.21\textwidth]{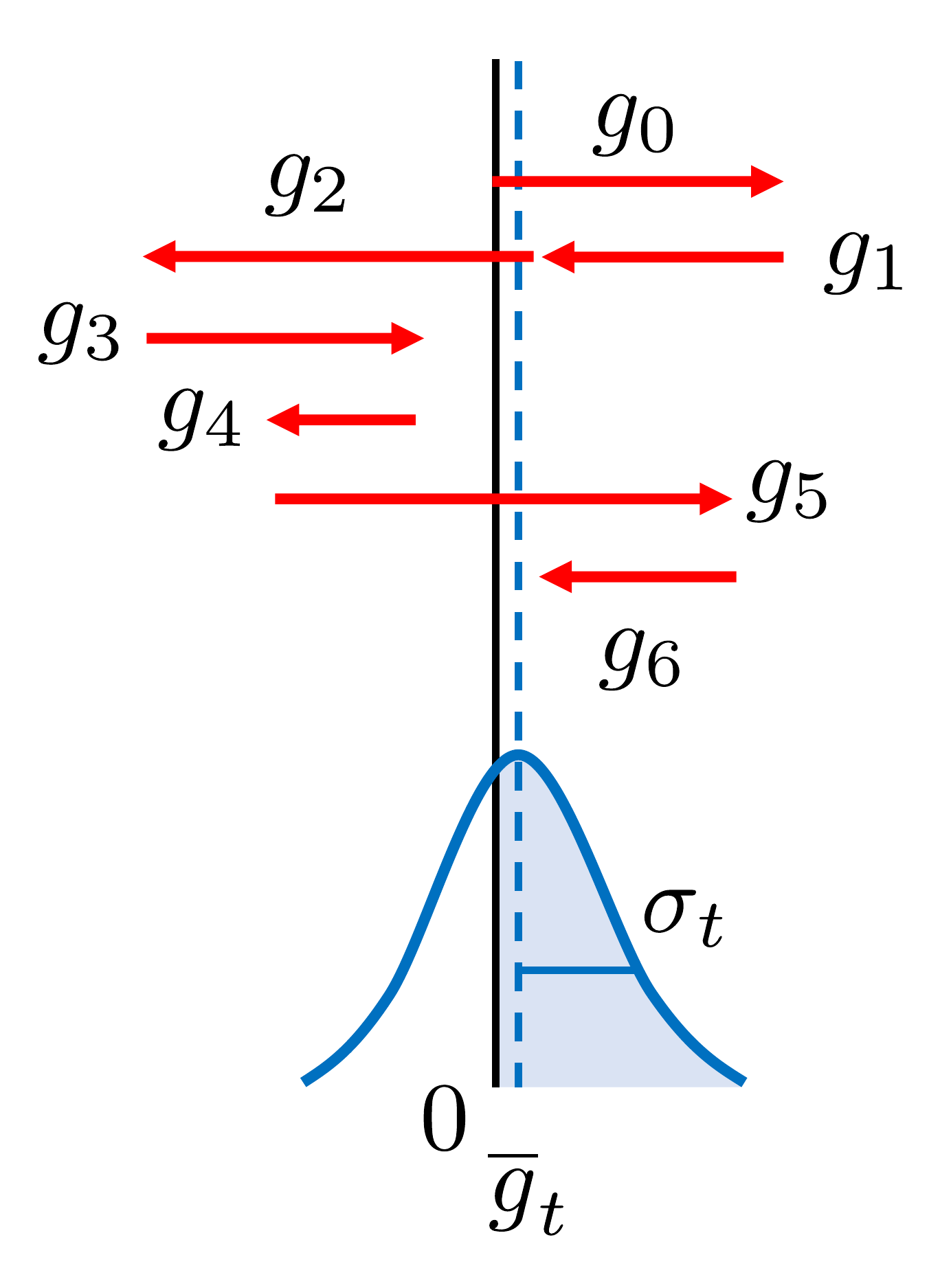}
}
% \caption{\label{fig:landscape} Conceptual difference between too large learning rate (left) and too small learning rate (right). Too large learning rate is characterized by low conformity of the past gradients (different directions). Too small learning rate is characterized by high conformity (same directions). This principle was introduced in RPROP \cite{Riedmiller1993-ng} and is used in this work.}
\caption{\label{fig:landscape} Intuitive understanding of CProp. CProp considers each parameter dimension independently. (Left) Total conformity of gradients owing to its large mean and small variance, the collective sign of gradients is clearly positive. (Mid) High conformity of gradients, the collective sign is positive with high confidence. (Right) Low conformity of gradients, either signs are almost equally probable. CProp scales the original learning rate according to the conformity where the conformity is measured from the area under curve (beyond or below 0) of the sampling distribution of the mean of gradients.}
\vspace{-0.25cm}
\end{figure*}

\section{Introduction\label{introduction}}

Training a neural network with SGD requires a properly tuned learning rate (step size) which should be the largest possible while not being too large to overstep. However, there is usually no single best learning rate throughout the training process because the loss landscape is not equally sensitive i.e. near a local optimum the margin of error is lower, overstepping becomes much easier \cite{Li2017-ds}. Hence, a reduced learning rate is advised and required for convergence guarantee \cite{Kingma2014-jt}. Techniques like learning rate decay are usually employed to deal with such problems \cite{Tan2019-yb}. This, turns out, to exacerbate the problem of tuning since we introduce even more hyperparameters. If a step decay learning rate is used, when to decay and how much to decay need to be determined beforehand, worse yet many decay steps might be employed as usually seen in literature \cite{Simonyan2014-wm,Krizhevsky2012-zm,He2016-wj}. This problem is referred as a learning rate scheduling problem. 

There is a another related problem regarding the learning rate. A neural network comprises of millions of weights and hundreds of layers is not likely to share a single global learning rate owing to the fact that the size of the gradient could be vastly different (especially the early and the later layers \cite{Yedida2019-my}). Adaptive gradient techniques have been developed to address this problem. They focus on forming a new gradient which is easier to share a single learning rate. For example, in the case of Adam \cite{Kingma2014-jt}, the gradient is divided by the running average of its square. This, on average, makes the adaptive gradient smaller than one for all parameters. It is now possible to use a single shared learning rate since we have some guarantee on the size of the gradients. 

However, the adaptive gradient does not address the problem of learning scheduling explicitly. It might or might not reduce the learning rate over time. It is the duty of a practitioner to reduce the learning rate as one sees fit. One could then characterize that learning rate scheduling concerns itself on the temporal-view of the learning rate, and adaptive gradient concerns itself on the spatial-view of the learning rate. In this paper, we aim to address both problems at the same time.

Optimal learning rate is another line of work that deals with both problems at the same time. We could derive the optimal learning rate from the second-order information of the local curvature of the loss function. This is compute and memory intensive requiring approximation which was proposed in \cite{Schaul2013-lf} where it approximates only the diagonal Hessian. Another way is to estimate the Lipschitz constant of the loss function which will hint the learning rate as in \cite{Yedida2019-my}. However, estimating a Lipschitz constant is not trivial. It could estimate the constant only once per epoch. Probably the most related work, is RPROP \cite{Riedmiller1993-ng} and its variants \cite{Anastasiadis2003-gm,Igel2003-ts}. RPROP is an iterative algorithm to determine the optimal learning rate using the signs of consecutive gradients. It has an assumption that if the gradient signs invert, the optimizer has overstepped a local minima, then the learning rate should be reduced. In many variants, the overstep should be undone on the next iteration. Otherwise, RPROP increases the learning rate for the next iteration. However, RPROP is a batch gradient algorithm. It requires correct gradients to make decision whether to increase or decrease the learning rate. A naive adaptation to mini-batch setting would drastically underestimate the learning rate since the gradient signs could easily invert due to their randomness.

Inspired by RPROP but for mini-batch setting, we propose \textbf{CProp}, \textit{Conformity Propagation}, to scale the the gradient according to its past conformity. In a mini-batch setting, we cannot determine the sings of consecutive gradients easily. Instead, we use a long history of gradients to determine how much the gradients conform, scoring from 0 to 1, the higher conformity the larger the scale. This score is suitable to use as a scaling factor for the gradient and is compatible to any optimizer by adding just one multiplication term. CProp gives a scale for each individual parameter and adapts throughout the learning process. 

The paper is structured as follows. In section \ref{method}, we describe CProp and how to use it with any optimizer. In section \ref{experiment}, we conducted a series of experiments to show that it performs well in a variety of settings. In section \ref{discussion}, we try to answer a few possible questions regarding CProp, and cases where it does not work as well. 
% In section \ref{analysis}, we show the proof of convergence of CProp.  

\newcommand{\Pg}{\Prob_{g, t}}
\newcommand{\Pgbar}{\Prob_{\overline{g}, t}}
\newcommand{\g}{g_t}
\newcommand{\gbar}{\overline{g}_t}
\newcommand{\s}{s_t}
\newcommand{\Plt}{p_{t,<0}}
\newcommand{\Pgt}{p_{t,>0}}
\newcommand{\erf}{\mathrm{erf}}

\section{Adaptive gradient scaling from past gradient conformity\label{method}}

CProp is a shorthand for \textit{Conformity Propagation} which gets fits name from Resilient Propagation (RPROP) \cite{Riedmiller1993-ng}. The idea of CProp is to look at a long history of past gradients and measure their \textit{conformity} scoring from 0 to 1. Why from 0 to 1? This is because CProp works as a scaling term to an original learning rate. It allows CProp to augment any optimizer while the scaling down property does away the need for re-tuning the learning rate. Figure \ref{fig:landscape} overviews CProp. One might think that the \textit{conformity} should be measured from the directions of the past gradients. In reality, the gradients are of different sizes, some might be noises. Considering only the signs of gradients would amplify those small noisy gradients making the \textit{conformity} score not useful. Instead, CProp measures the \textit{conformity} by asking a question \textit{"Does the past gradients conform enough to collectively show a clear sign, positive or negative?"} Imagine two extreme cases as follows. If past gradients have clear positive average and small variance (figure \ref{fig:landscape}, left), it suggests that the learning rate is not too large and we are making consistent progress, in other words, \textbf{high conformity}. On the contrary, if the past gradients have near-zero average and large variance (figure \ref{fig:landscape}, right), it suggests that the learning rate is too large, in other words, \textbf{low conformity}. The goal of CProp is to propose a way to quantify the conformity into a scale from 0 to 1. 

Interestingly, scaling between 0 and 1 corresponds well with probability framework. Hence, CProp derives its scaling term from the \textit{confidence} that the average of the past gradients has a clear sign either positive or negative. If the past gradient collectively have a positive sign (or negative) with high confidence, it corresponds to high conformity, and vice versa. We denote the \textit{confidence} of having positive sign and negative sign as  $\Plt^i$ and $\Pgt^i$ respectively where $t$ denotes time (a specific iteration). Since CProp scales the learning rate fore each parameter independently, we shall see $i$ almost everywhere which regards to the $i$-th element of the parameter. By estimating this \textit{confidence}, CProp could derive its scales. More specifically, we say that the \textit{conformity} of the $i$-th element, $\s^i$, is proportional to:
\begin{equation}
\s^i \propto \max\left(\Pgt^i,  \Plt^i \right)
\end{equation}
because the conformity does not care for a specific sign. 
We use the notation:

\begin{equation*}
\Pgt^i = \Pgbar^i(\gbar^i > 0)
\end{equation*}
\begin{equation*}
\Plt^i = \Pgbar^i(\gbar^i < 0)
\end{equation*}
where $\gbar^i = \frac{1}{t}\sum_{\tau=0}^t g^i_\tau$ is an empirical mean of the past gradients, and $g^i_t$ is an $i$-th element of a \textit{mini-batch} gradient at time $t$. From now on we will omit $i$ in the superscript to reduce clutter.
Note that $\max\left(\Plt,  \Pgt \right) \in [0.5, 1]$. To keep $\s \in [0, 1]$, we rescale it, and define the CProp scaling factor as:
\begin{equation}\label{eq:main}
\s = 2 \left(\max\left(\Pgt,  \Plt \right) - \half \right)
\end{equation}
which is equivalent to $2\left| \Plt - 0.5 \right|$. 

Generally, $\Pgbar$ is an unknown distribution, and we have only one sample of it (we have only a single history of gradients). It is difficult to estimate the CDF of $\Pgbar$ which is required to get the scaling factor. 
To make progress, we assume that $\g$ is identically independently sampled from some distribution $\Pg$\footnote{This assumption is incorrect because $\Pg$  depends on all previous $t$, neither independent nor identical over $t$. However, it enables us to estimate the CDF.}.
As will be further discussed, in practice, we use a moving average to limit how far back we care about the gradients. This ameliorates our \textit{identical} assumption to some extent. 
We now approach to estimate the CDF from the fact that $\Pgbar$ is a sampled mean distribution of $\Pg$ which can be estimated from resampling techniques like bootstrapping \cite{Bickel1981-ss}. However, in a mini-batch setting, we favor speed over the fidelity of the CDF. Bootstrapping will be too slow for that matter. We look for more efficient computation of the CDF which is possible if we make further assumption that $\Pgbar$ is a Normal distribution. As we deal with a longer history of past gradients i.e. $t \gg 0$, it is more likely that $\Pgbar$ is closer to a Normal distribution.

\subsection{Estimating the CDF of $\Pgbar$}

We assumed $\Pgbar$ to be a Normal distribution which has its CDF defined as:
\begin{equation}
\Phi(x; \mu, \sigma) = \half  \left[ 1 + \erf(\frac{x-\mu}{\sigma\sqrt{2}}) \right]
\end{equation}
where $\erf(x)$ is an error function defined as:
\begin{equation}\label{eq:err_function}
\erf(x) = \frac{2}{\sqrt{\pi}} \int_0^x e^{-t^2} dt
\end{equation}
the CDF of a Normal distribution can be computed efficiently on GPU, and is implemented in major deep learning frameworks. Note that one could make further approximation using a Logistic distribution, which resembles a Normal distribution, but has a closed form CDF. We leave this as a future work. 

The confidences that the past gradients collectively have negative and positive sign are in the form $\Plt = \Phi(0; \mu_t, \sigma_t)$ and $\Pgt = 1 - \Plt$ respectively.

We now need to estimate the $\mu_t$ and $\sigma_t$ of $\Pgbar$. An obvious choice for estimating $\mu_t$ is an empirical average $\gbar$. 
In case of $\sigma_t$, the most efficient way is to estimate it using a standard error (with Bessel's correction). Although it is biased with small $N$, it is less biased with large $N$. Here, $N = t$. We write:
\begin{equation}
\sigma_t \approx \sqrt{\frac{1}{t} \mathrm{var}_t}
\end{equation}
where:
\begin{equation}\label{eq:variance}
\begin{split}
\mathrm{var}_t 
&= \frac{1}{t-1} \sum_{\tau=0}^t (g_\tau - \gbar)^2 \\
&= \frac{t}{t-1} \left[ \underbrace{\left(\frac{1}{t}\sum_{\tau=0}^t g_\tau^2 \right)}_\text{mean squared gradient} - \underbrace{\gbar^2}_\text{mean gradient squared} \right]
\end{split}
\end{equation}
\noindent Note that a better estimate like Student's T-statistic \cite{Student1908-rt} could be used, but it should not be much of an improvement since we usually work with a long history of gradients i.e. $t \gg 0$. 
\subsection{Exponential moving average}

One argument against equation \ref{eq:main} is that incorporating gradients since $t=0$ might interfere with the scaling when $t \gg 0$. We might want to focus more on recent gradients than the long past gradients. A moving average is more suitable. We define $\beta$ to be a hyperparameter to control how far back the gradients CProp looks, the larger the $\beta$ the further back CProp looks, $\beta \in [0, 1)$. We define $m_t$ to be the moving average of gradients, and $v_t$ to be the moving average of squared gradients.  Both are bias corrected as proposed in \cite{Kingma2014-jt}. $m_t$ could be used instead of $\gbar$. For equation \ref{eq:variance}, we substitute $v_t$ for the first term in the bracket, we obtain:
\begin{equation}
\mathrm{var}_t \approx \frac{N}{N-1} \left[v_t- m_t^2 \right]
\end{equation}
and obtain:
\begin{equation}
\sigma_t \approx \sqrt{\frac{1}{N} \frac{N}{N-1} \left(v_t- m_t^2 \right)}
\end{equation}
With moving average, we effectively limit how far back we look at the gradients, hence $N \leq t$. A reasonable estimate of $N$ is $1/(1-\beta)$. If $\beta = 0.9$, $\beta^{10}$ is small, the effect of $g_{t-10}$ becomes negligible. However, it is incorrect to assume that $N$ is a constant. Early in training, $N_t$ is potentially much smaller, and is growing until it converges to $1/(1-\beta)$ at $t \rightarrow \infty$. Finally, we use $N_t = (1-\beta^t)/(1-\beta)$. It is easily observed that $1-\beta^t$ converges to $1$ at $t \rightarrow \infty$. We found that $\beta = 0.999$ worked well in most cases.

The full CProp algorithm is provided in algorithm \ref{algo1}. CProp has a linear complexity with the number of parameters.

\begin{algorithm}[t]
\caption{\label{algo1}CProp with any optimizer. Any vector-to-vector multiplication is element-wise.}
\begin{algorithmic}
\small
\STATE {Given $\beta \in [0,1)$ CProp gradient horizon, $c$ overconfident coefficient}
\STATE {Given $f(\theta)$ objective function with parameters $\theta$}

\STATE {$m_0 \leftarrow \bf{0}$ \quad (Initialize $1^\text{st}$ moment vector)}
\STATE {$v_0 \leftarrow \bf{0}$ \quad (Initialize $2^\text{nd}$ moment vector)}
\STATE $t \leftarrow 0$
\WHILE {$\theta_t$ not converged}
  \STATE $t \leftarrow t+1$
  \STATE $m_t \leftarrow \beta m_{t-1} + (1-\beta) g_t$
  \STATE $v_t \leftarrow \beta v_{t-1} + (1-\beta) g_t^2$
  \STATE $\hat{m}_t \leftarrow m_t/(1-\beta^t)$ \quad (Bias correction)
  \STATE $\hat{v}_t \leftarrow v_t/(1-\beta^t)$ \quad (Bias correction)
  
  \STATE $n_t \leftarrow (1-\beta^t)/(1-\beta)$ \quad (Estimated number of items)
  \STATE $\sigma_t \leftarrow \sqrt{\max(\hat{v}_t - \hat{m}_t^2, 0) / (n_t-1+\epsilon)} + \epsilon$
  \STATE $p_{t,<0} \leftarrow \mathrm{CDF}_{\mathcal{N}(\hat{m}_t, \sigma_t)}(\bf{0})$
  \STATE $\hat{s}_t \leftarrow \min(2c \left|(p_{t,<0} - 0.5\right|, 1)$ \quad (Conformity)
  \STATE $\hat{g}_t \leftarrow \text{optimizer update direction}$
  \STATE $\theta_t \leftarrow \theta_{t-1} + \hat{s}_t \cdot \hat{g}_t$
  
\ENDWHILE
\end{algorithmic}
\label{alg:crude_algo}
\end{algorithm}

\subsection{Overconfidence coefficient}
It is observed in many previous works that larger learning rate tends to generalize better \cite{Keskar2016-zv, Jastrzebski2018-mz, Xing2018-oa}. Since CProp scales down the learning rate, CProp could have harmful effects on the generalization performance. To mitigate this problem, we could make CProp more confident than usual by multiplying the \textit{conformity} score by some constant $c$ as follows:
\begin{equation}\label{eq:tuning}
\hat{s}_t = \min (c \times  \s , 1)
\end{equation}
$c$ is a tunable hyperparameter dictating how overconfident should CProp be. Now, $\hat{s}_t$ is used as the scaling factor instead. While $c>1$ is likely to saturate previously large \textit{conformity} to 1, it keeps very small \textit{conformity} small which is the main principle of CProp. Based on empirical evidence, $c=1$ is optimal regarding training loss, yet larger $c$'s trade off training performance with generalization favorably. Note that when $c \rightarrow \infty $, CProp converges to the original optimizer. We provide empirical evidence in section \ref{dis:generalization}.

\subsection{Other possible heuristics\label{method:alternatives}}
How does CProp differ from just adding another adaptive term? Here we consider some other heuristics which capture more or less the same concept as CProp including: 
\begin{enumerate}
\item \textbf{Relative rate:} $s_t = \frac{\left| \sum_{\tau=0}^t g^i_\tau \right|}{\sum_{\tau=0}^t \left| g_\tau \right|}$ 
\item \textbf{Moment rate:} $s_t = \frac{\left| \frac{1}{t} \sum_{\tau=0}^t g_\tau \right|}{ \sqrt{\frac{1}{t}\sum_{\tau=0}^t  (g_\tau)^2 }}$ which is the same as Adam's adaptive gradient with absolute while using the same betas $\beta_1 = \beta_2$. Under this formulation, the conformity scale almost never gets close to 1 underestimating the learning rate. 
\end{enumerate}
We have conducted experiments and come to a conclusion that CProp is a better choice in terms of effectiveness and ease of tuning. The results are shown in supplementary material (SM). The two methods required an extensive tuning of $c$ in equation \ref{eq:tuning} to be competitive with our CProp. Since both do not have clear interpretations comparing to CProp, tuning is a much harder task and may not transfer across settings.
\section{Experiments\label{experiment}}

In this section, we conducted a series of experiments to empirically show the reach of CProp across multiple tasks and optimizers.

Our aim is to show the relative improvement over traditional optimizers. We considered mainly SGD and Adam \cite{Kingma2014-jt}. However, we also conducted a few experiments with RMSProp \cite{Tieleman2012-rf}, AMSGrad \cite{Reddi2018-mk} and AdaBound\footnote{From \url{https://github.com/Luolc/AdaBound}} \cite{Luo2019-kx}. Default parameters are used for each corresponding optimizer (see specifics in SM). We mostly concern ourselves with training performance i.e. training loss. Note that reaching the state of the art is not our goal. All of our code including architectures, training loop were implemented\footnote{We aim to make the code regarding the algorithm publicly available.} on Pytorch \cite{Paszke2017-dt}. All experiments were run with 3 different random seeds and plotted with the area as one standard deviation. Plots were smoothed to improve readability.

Since the starting learning rate is important for the trainability, we have conducted separate experiments to find the learning rate which reduces the training loss fastest. More specifically, we run experiments with a given learning rate for a short interval, 5,000 iterations, the best learning rate is the one with lowest training loss at 5,000-th iteration\footnote{In some cases, a few learning rates performed equally well, we chose one subjectively.}. We search by increasing/decreasing by a factor of 3 until the best is found. For example, we start with two guesses, say, 0.1 and 0.3. If later we find that 0.1 is better, we continue to search with 0.03. If 0.03 is even better than 0.1, we continue to search with 0.01, and so on. This is repeated until the best is found. Note that we only tuned for the learning rates for traditional optimizers, not for CProp. We provide our learning rate search results in SM.

Unless stated otherwise, we used the batch size of 32, we set our CProp's $\beta$ (gradient horizon) to be $0.999$, $c$ (overconfidence coefficient) to be 1, and $\epsilon$ to be $10^{-8}$ which worked well.

\subsection{Fashion MNIST}

\begin{figure}[t]
\centerline{
\includegraphics[width=0.5\textwidth]{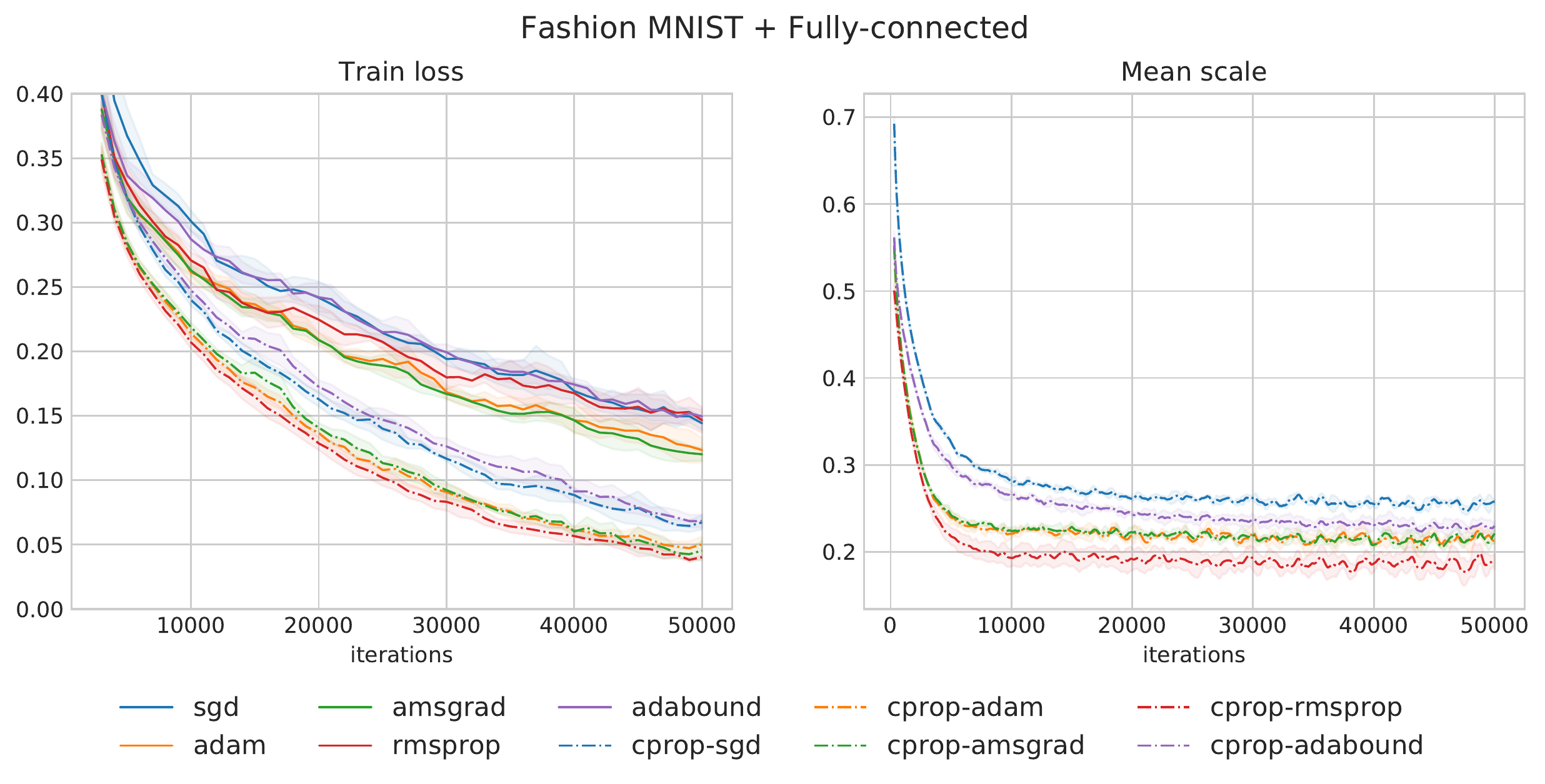}
}
\vspace{-0.15cm}
\caption{\label{fig:fmnist-fc} Results on Fashion MNIST with fully-connected architecture. \textit{Mean scale} denotes the mean of conformity scale. Each plot is from 3 different random seeds. The area is one standard deviation. \textit{CProp} is a respective optimizer augmented with CProp. Best viewed in color.}
\vspace{-0.25cm}
\end{figure}

Fashion MNIST \cite{Xiao2017-sx} is an MNIST-like dataset having the same dimensions and size but being a bit harder than MNIST \cite{LeCun2010-ba}. This dataset is small enough to solve with a multi-layer perceptron (MLP). We experimented with two architectures: an MLP with 784-300-100-10 neurons, and a CNN with 16-32-64-128-avgpool-10 output channels. Details of these architectures are provided in SM. The results are shown in figure \ref{fig:fmnist-fc} for fully-connected. To reduce redundancy, we combined the plot of CNN with varying batch size experiments shown in figure \ref{fig:batchsize}. CProp-augmented optimizers performed favorably in all settings.

% \begin{figure}[t]
% \centerline{
% \includegraphics[width=0.5\textwidth]{latex/sections/figures/fmnist+cnn.pdf}
% }
% \vspace{-0.15cm}
% \caption{\label{fig:fmnist-cnn} Results on Fashion MNIST with CNN architecture.}
% \vspace{-0.25cm}
% \end{figure}
\subsection{Cifar100}

\begin{figure}[t]
\centerline{
\includegraphics[width=0.5\textwidth]{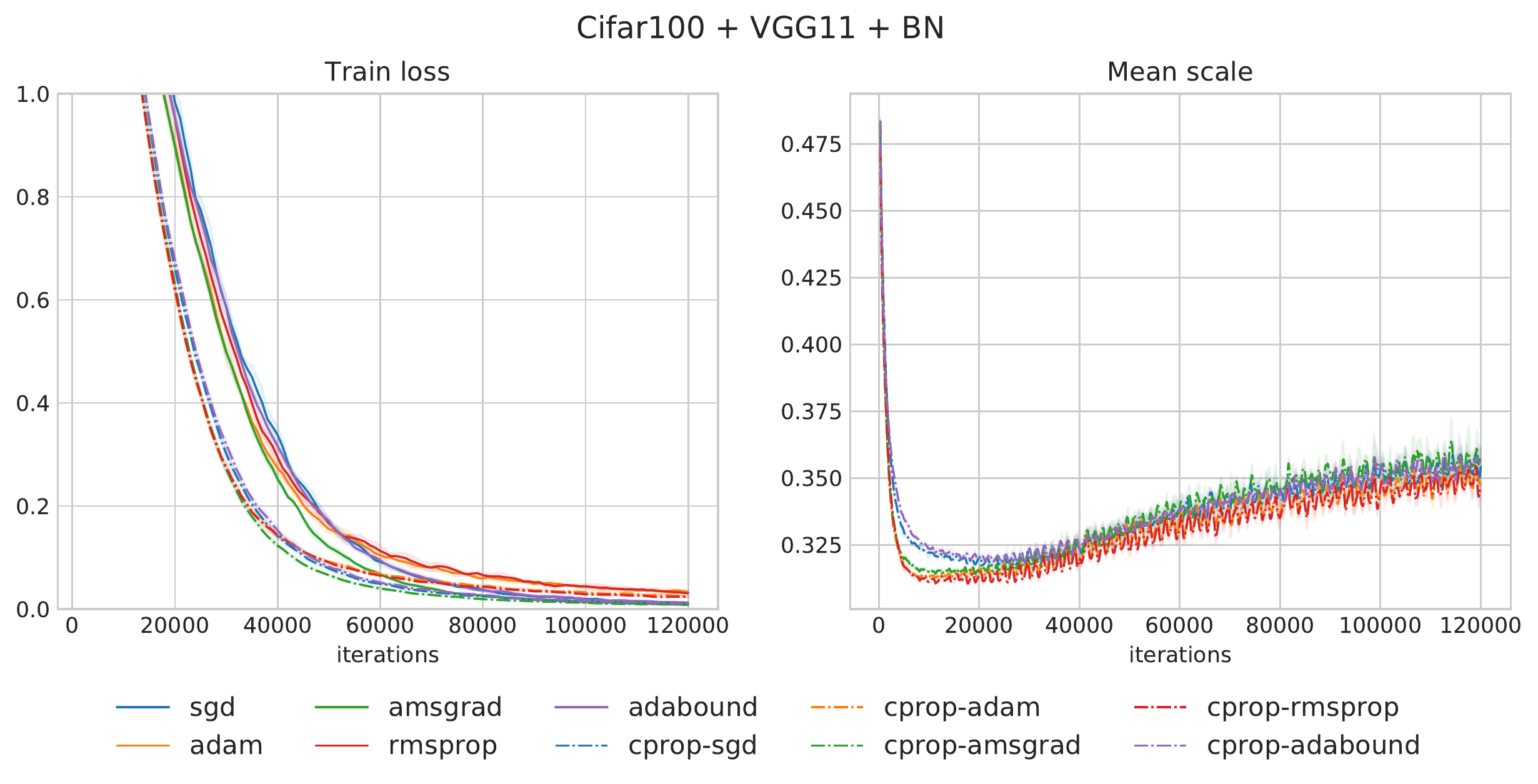}
}
\vspace{-0.15cm}
\caption{\label{fig:cifar-vgg} Results on Cifar100 with VGG architecture.}
\vspace{-0.25cm}
\end{figure}

\begin{figure}[t]
\centerline{
\includegraphics[width=0.5\textwidth]{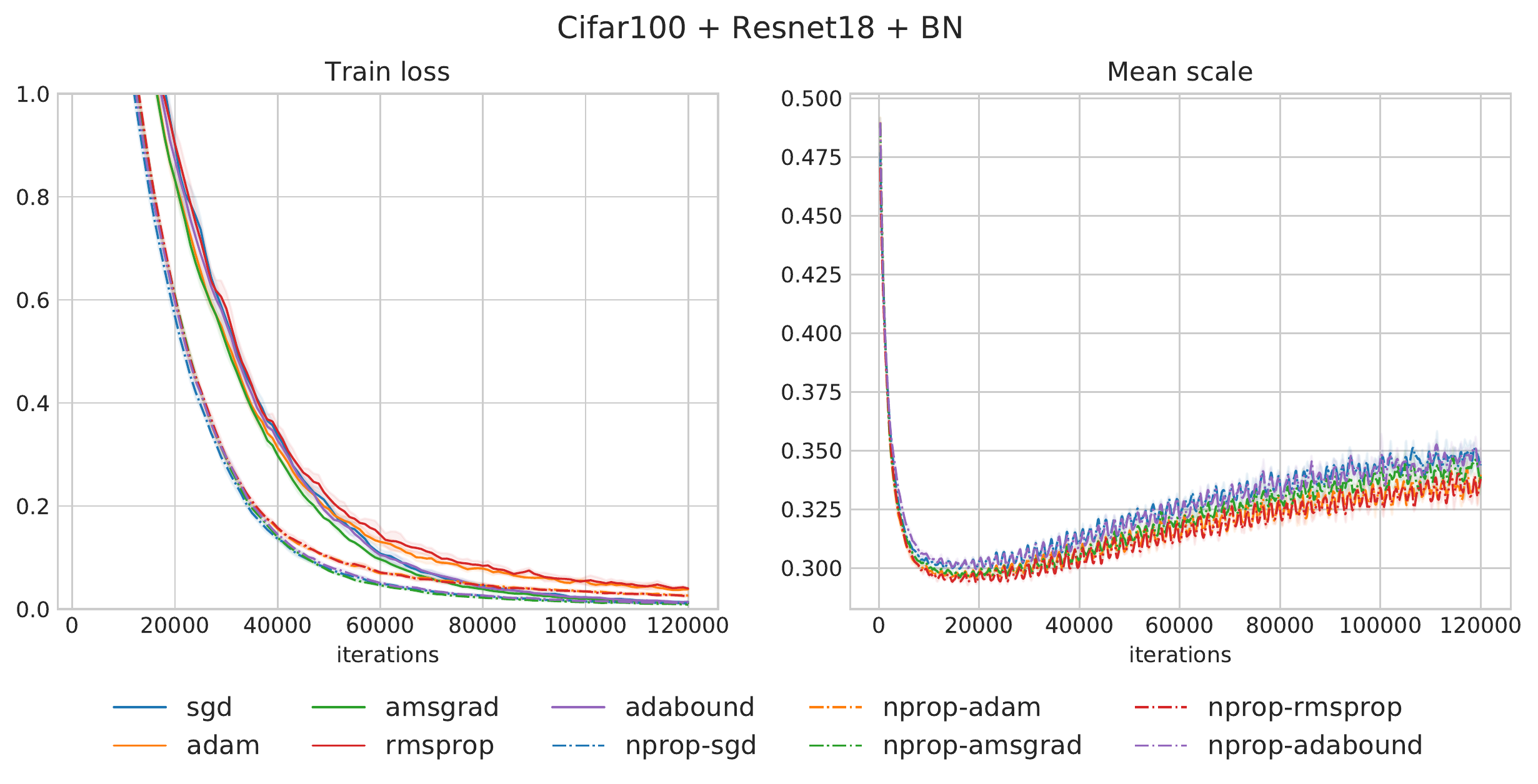}
}
\vspace{-0.15cm}
\caption{\label{fig:cifar-resnet} Results on Cifar100 with Resnet architecture.}
\vspace{-0.25cm}
\end{figure}

Cifar100 \cite{Krizhevsky2009-ar} is a 100-way small image classification task with RGB color space. It contains 50,000 images of training data, each is 32x32 pixels. Though small, Cifar100 is not trivial. We prepared the dataset as follows: random crop of size 32x32 with zero padding of size 4, random horizontal flip, and normalize by means and variances. Here, we compared based on two kinds of architectures: a VGG11-like \cite{Simonyan2014-wm} and a Resnet18-like \cite{He2016-wj}. Details of these architectures are provided in SM. We z-normalized the images before feeding to the networks. The results are shown in figure \ref{fig:cifar-vgg} for VGG, and \ref{fig:cifar-resnet} for Resnet. CProp-augmented optimizers improved over their unaugmented peers in all cases. 

\subsection{Tiny Imagenet}

\begin{figure}[t]
\centerline{\includegraphics[width=0.5\textwidth]{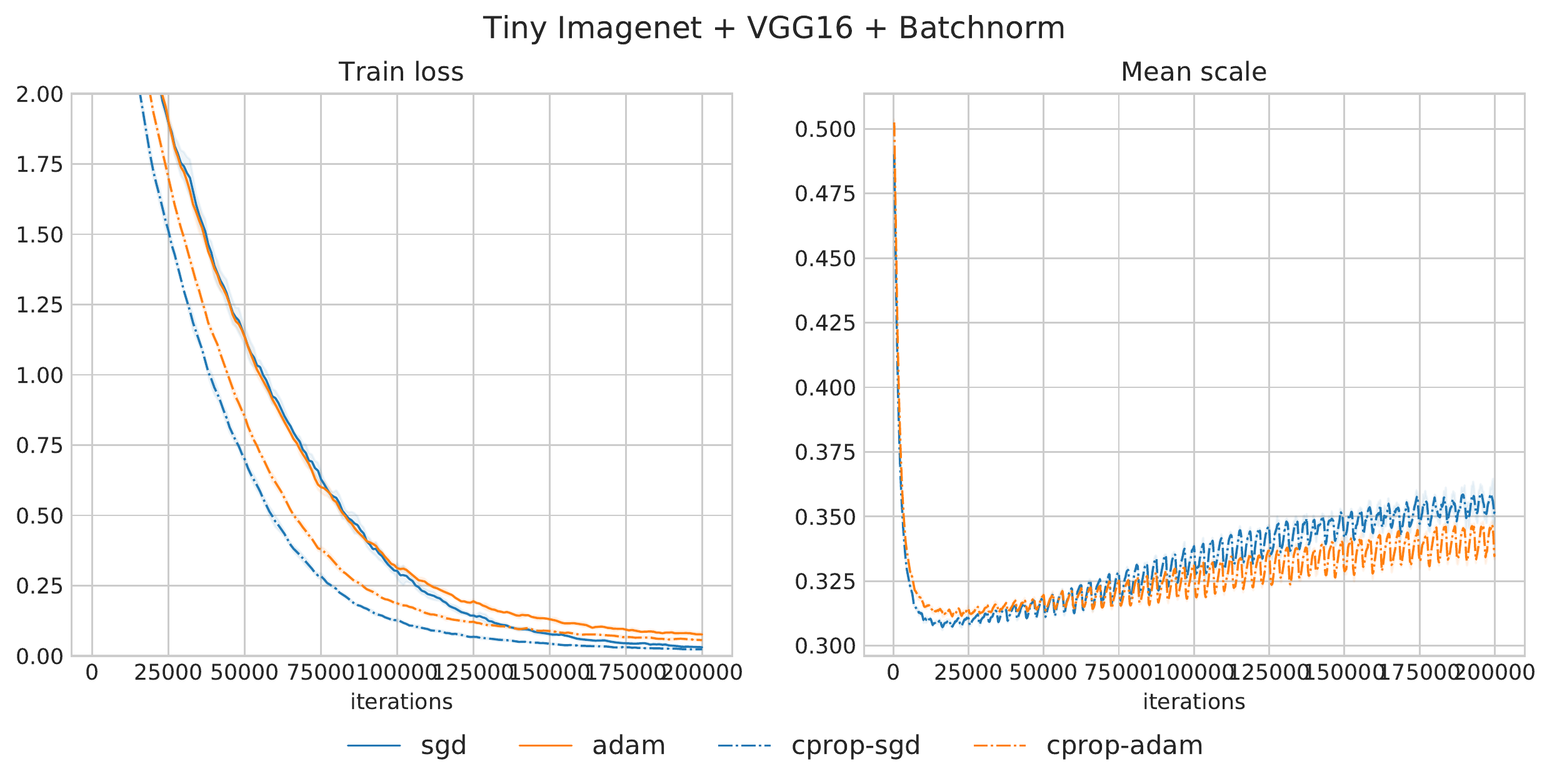}}
\vspace{-0.15cm}
\caption{\label{fig:tinyimagenet-vgg} Results on Tiny Imagenet with VGG architecture.}
\vspace{-0.25cm}
\end{figure}

\begin{figure}[t]
\centerline{\includegraphics[width=0.5\textwidth]{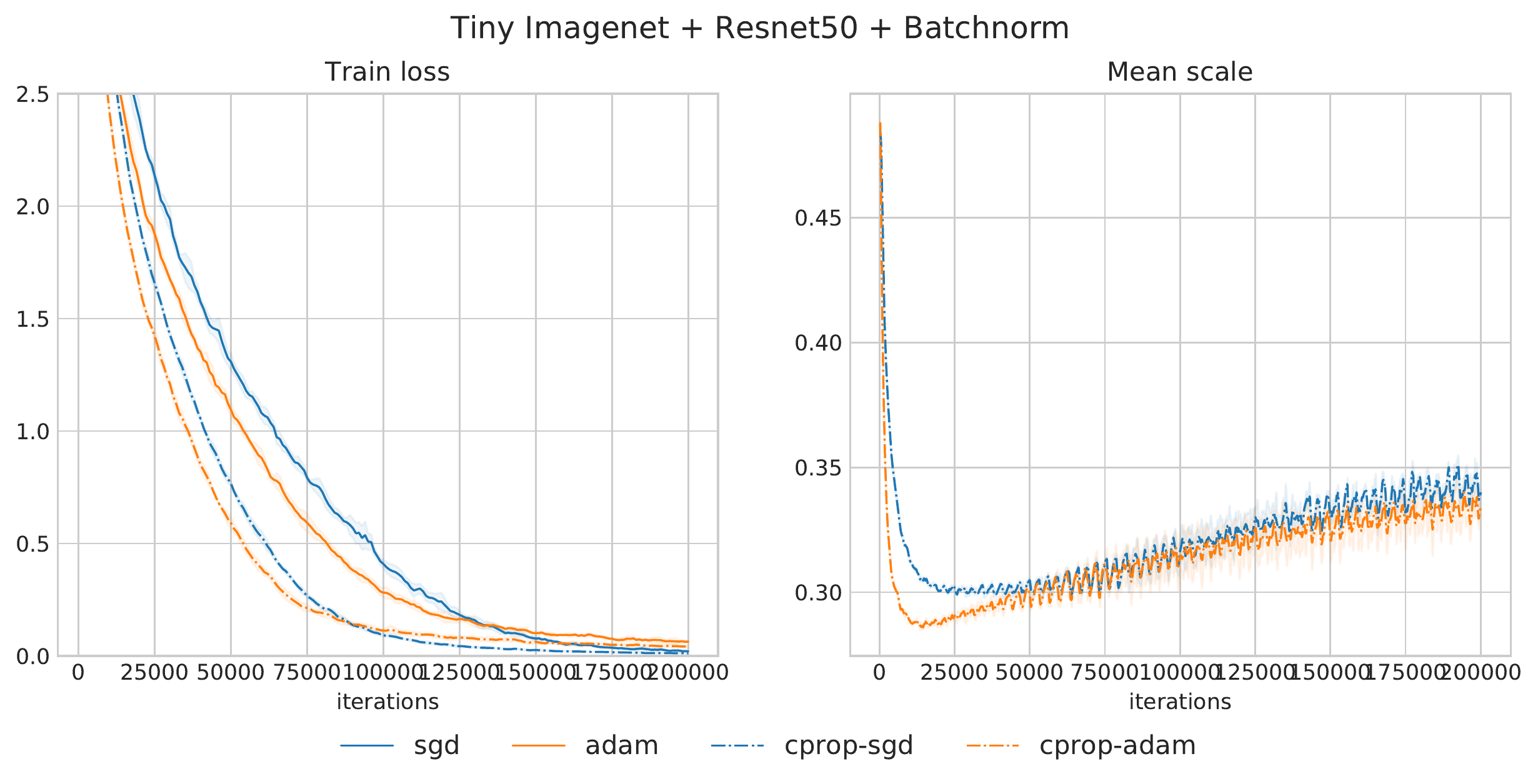}}
\vspace{-0.15cm}
\caption{\label{fig:tinyimagenet-resnet} Results on Tiny Imagenet with Resnet architecture.}
\vspace{-0.25cm}
\end{figure}

Tiny Imagenet \cite{Stanford_CS231N_undated-wl} aims to provide a smaller version of Imagenet classification task. It has 200 classes of 64x64 images with the total of 100,000 training images. We prepared the dataset as follows: random crop of size 64x64 with zero padding of size 8, random horizontal flip, and normalize by means and variances. To evaluate the train loss, we created a smaller subset of the train dataset containing 10,000 images due to the training dataset being too large to evaluate frequently. We compared two architectures: VGG16-like and Resnet50-like architectures, both have batch normalization applied \cite{Ioffe2015-ik}. Details of these architectures are provided in SM. The results are shown in figure \ref{fig:tinyimagenet-vgg} for VGG and \ref{fig:tinyimagenet-resnet} for Resnet. CProp provides significant improvement over traditional SGD and Adam. 

\subsection{Language modeling}

There are two goals of this section: showing how CProp performs with LSTM models, and how CProp performs with language modeling task. As will be discussed later, CProp does not perform as well with the presence of strong dropout which is a crucial part of a language model e.g. AWD-LSTM \cite{Merity2017-hd}. We provide readers with two parts of the experiment: first, we demonstrate that CProp works well with LSTM models without dropout, second, we show how CProp fares against more realistic language modeling.

\begin{figure}[t]
\centerline{
\includegraphics[width=0.5\textwidth]{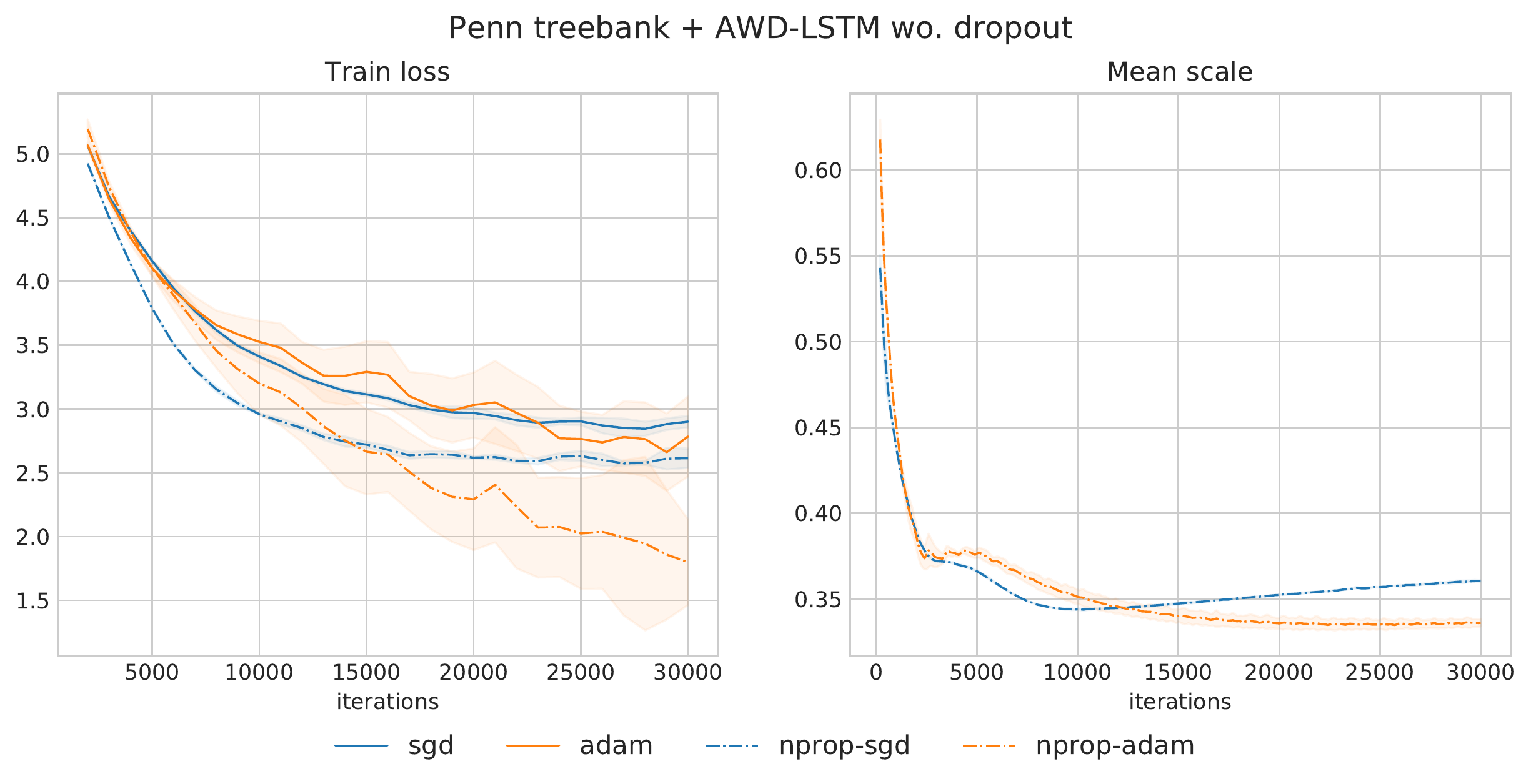}
}
\vspace{-0.15cm}
\centerline{
\includegraphics[width=0.5\textwidth]{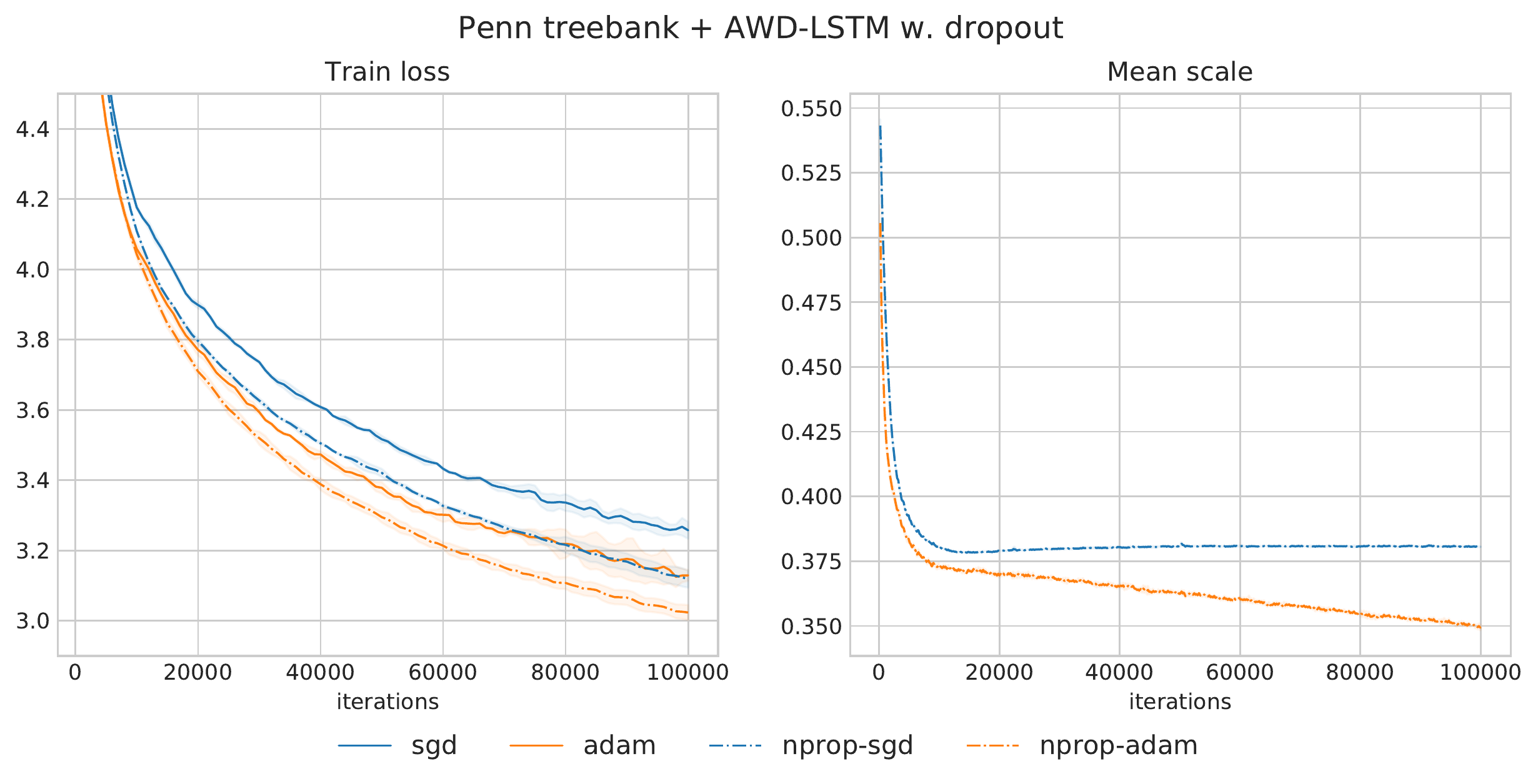}
}
\vspace{-0.15cm}
\caption{\label{fig:ptb-lstm} Results on Penn treebank with AWD-LSTM. (Top) AWD-LSTM without dropout. (Bottom) AWD-LSTM with dropout.}
\vspace{-0.25cm}
\end{figure}

We trained LSTM \cite{Hochreiter1997-uo} models on Penn treebank \cite{Marcus1993-vh} on word-level language modeling task. The dataset was prepared using Torchtext library \cite{Pytorch_undated-xz} which is described briefly here. We first chopped a long text into smaller texts to match the batch size of 32. A batch of input is a sliding window over these texts with the size and shift of truncated backpropagation (tbptt) which is 35 in our case. We applied gradient norm scaling of 0.25 to stabilize training. We used two kinds of networks: with and without dropout, both are based on AWD-LSTM \cite{Merity2017-hd} with tying embedding and output layer \cite{Inan2016-vd, Press2017-xg}. Details of the architecture are provided in SM. The results are shown in figure \ref{fig:ptb-lstm}. We see large improvement over traditional optimizers when no dropout is applied. The result is a bit less dramatic with dropout where the gaps are narrower. The hardships with dropout are further discussed in section \ref{dis:dropout}. However, these small gaps translate to large difference in terms of test perplexity shown in \ref{fig:generalization} in case of SGD.

\section{Discussion\label{discussion}}

\subsection{Learning rate selection\label{dis:learning_rate}}

CProp does not scale \textit{up} the learning rate. That means if the learning rate is too small to begin with, there would be very limited improvement to be gained from using CProp, in some cases, CProp could hurt the overall performance. This is the reason behind our learning rate selection philosophy which favors fastest reduction in training loss. True, the philosophy encourages large learning rates, yet we argue that smaller learning rates could still be applied via learning rate step decays. Ideally, if we select the largest learning rate for each moment, an improvement from CProp is expected. 

\subsection{Relationship to traditional learning rate scheduling\label{dis:scheduling}}
CProp and traditional scheduling techniques should be thought of as complementary more than adversary. It is true that CProp adjusts learning rate adaptively. However, it relies on gradient signals to make adjustments which are limited in long-term planning. For example, it is believed that temporary increasing the learning rate is beneficial by encouraging the optimization to explore more, reducing premature convergence to poor minima \cite{Smith2018-mo,Xing2018-oa}. This behavior cannot be derived from gradient signal, and CProp cannot do this. Hence, it is advisable to incorporate prior knowledge about the learning dynamics into the learning rate scheduler on top of CProp. 

% \begin{figure}[t]
% \centerline{
% \includegraphics[width=0.25\textwidth]{latex/sections/figures/cifar100-resnet-step-decay-test.pdf}
% \includegraphics[width=0.25\textwidth]{latex/sections/figures/cifar100-vgg-bn-decay-test.pdf}
% }
% \vspace{-0.15cm}
% \caption{\label{fig:scheduling} Applying traditional learning rate decay with CProp. The learning rate is divided 5.}
% \vspace{-0.25cm}
% \end{figure}

\subsection{How does the scale evolve over time?\label{dis:scaling}}
\begin{figure}[t]
\centerline{
\includegraphics[width=0.3\textwidth]{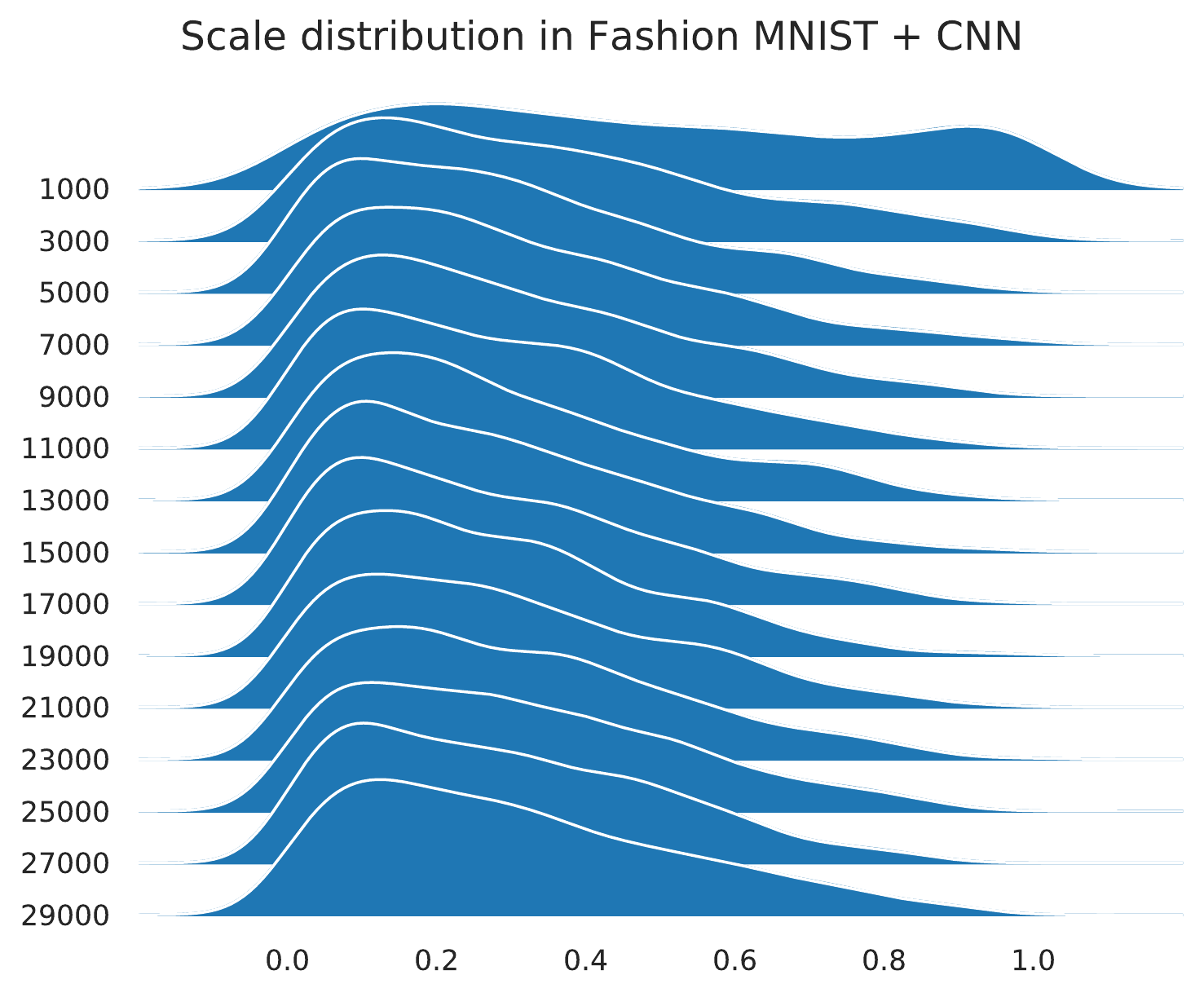}
\includegraphics[width=0.2\textwidth]{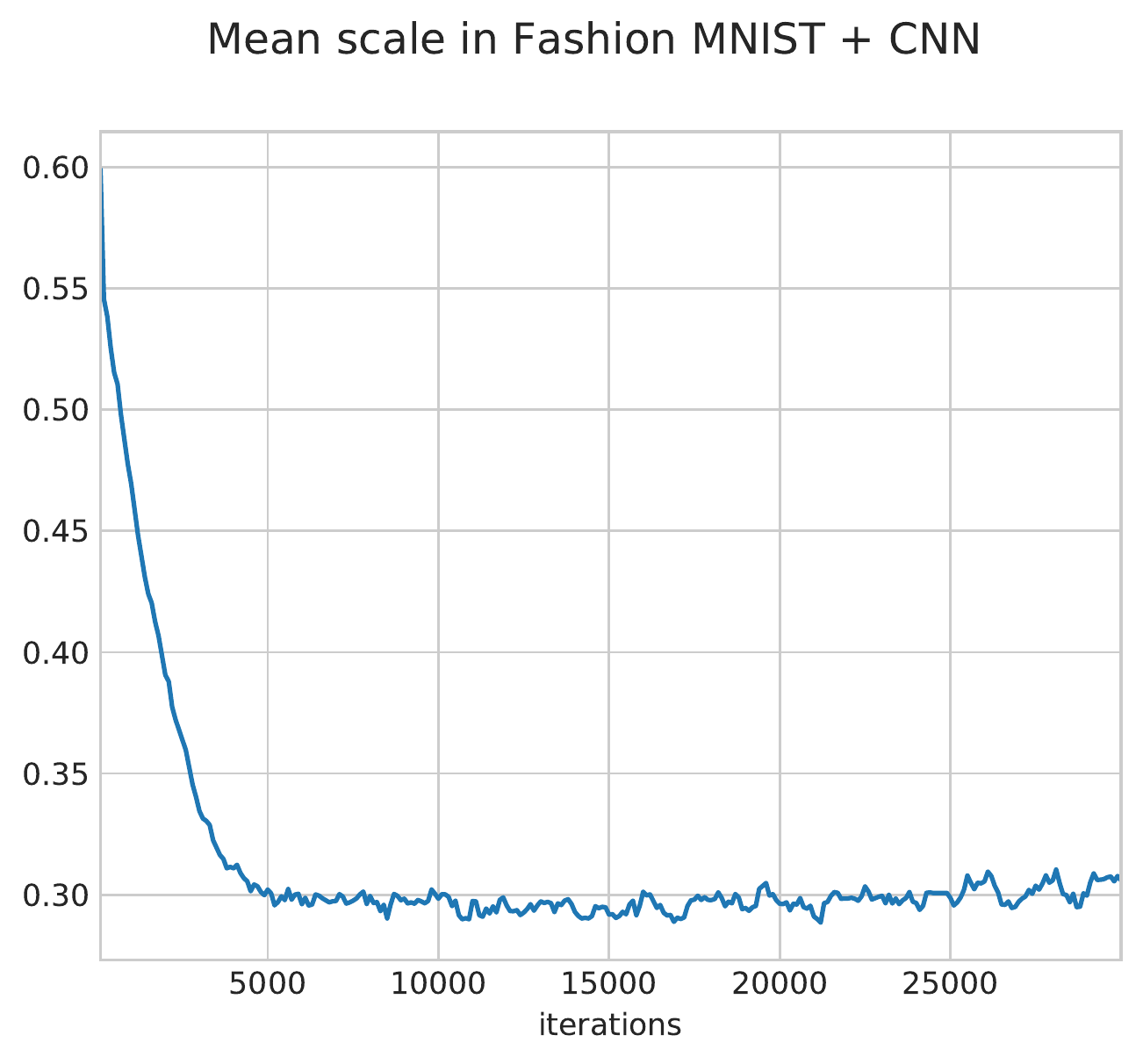}
}
\vspace{-0.15cm}
\caption{\label{fig:fmnist-cnn-scaling} CProp scaling distribution evolution over time in Fashion MNIST with CNN architecture. (left) The histogram of scaling plotted over time. The y-axis is the iterations. (right) The mean scale over time.}
\vspace{-0.25cm}
\end{figure}

In figure \ref{fig:fmnist-cnn-scaling}, we plotted the histograms of the scale evolving over time. CProp tends to scale down most of the learning rates quickly while maintaining high learning rate for a small portion throughout the training. During this moment, we see most of the improvement from CProp suggesting that, by reducing overly large learning rates, optimization progresses much easier. After that the scaling plateaus somewhat. It implies that the statistics derived from the gradients itself become more stationary. CProp is not likely to scale the gradient towards zero cleanly at $t \rightarrow \infty$ suggesting that CProp does not solve the problem of learning rate scheduling entirely.

\subsection{Effect of batch sizes\label{dis:batch_size}}
\begin{figure}[t]
\begin{center}
\includegraphics[width=0.5\textwidth]{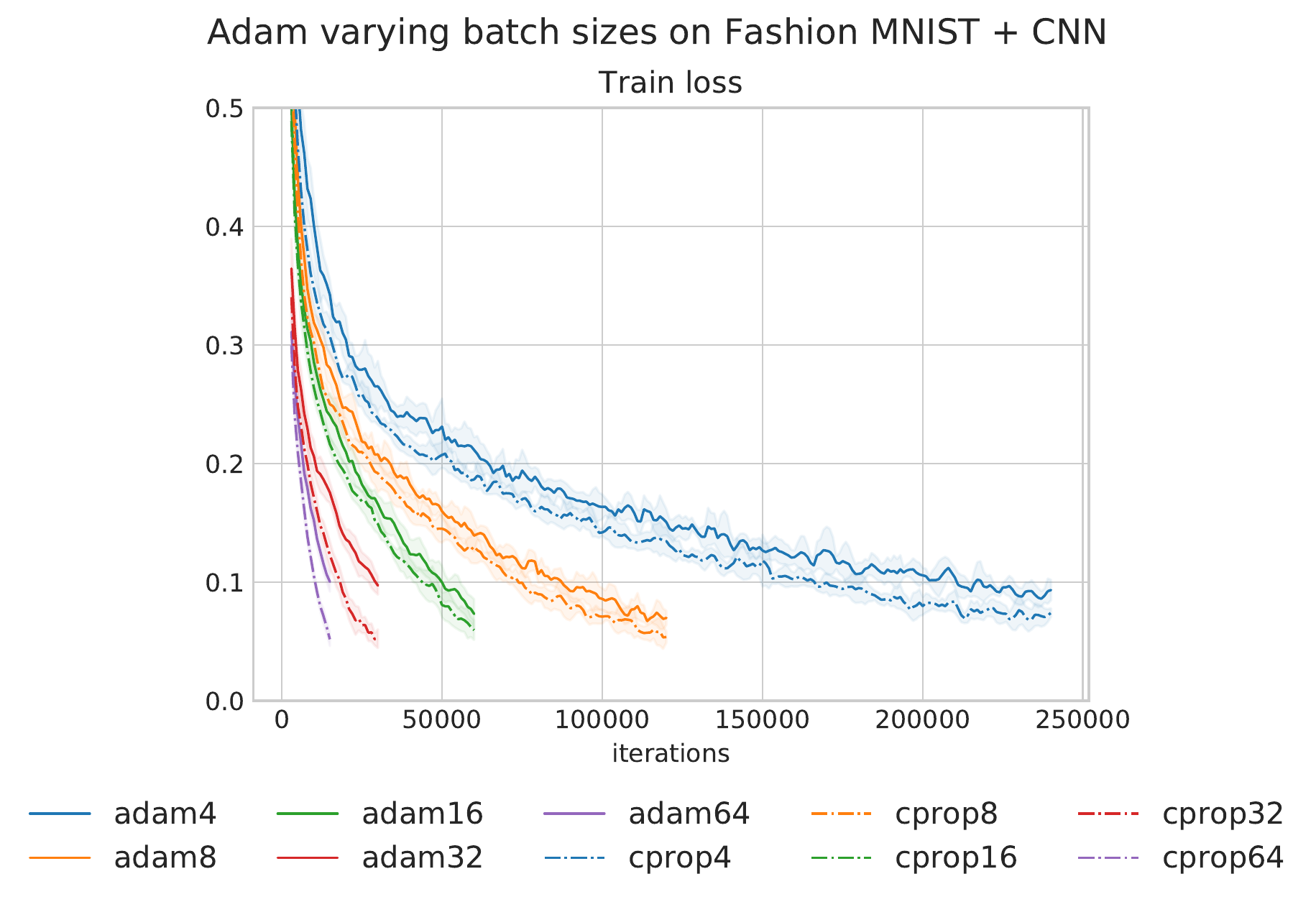}
\end{center}
\vspace{-0.25cm}
\caption{\label{fig:batchsize} Comparing different batch sizes from 4 to 64. CProp works well across batch sizes. Due to the space limitation, this of SGD is not presented here, yet CProp improved over SGD as well.}
\vspace{-0.45cm}
\end{figure}

CProp works well across batch sizes. In figure \ref{fig:batchsize}, we conducted experiments on different batch sizes, ranging from 4 to 64 for both SGD and Adam, to show the consistency of CProp. Reducing batch size is usually paired with reducing learning rate proportionally \cite{Smith2017-ne}. We used this principle with SGD where we scaled the learning rate proportionally to the change of batch size. However, this did not work well with Adam where the principle would underestimate the learning rate. We conducted learning rate selection separately for each batch size in Adam. The leaning rate selection results are provided in SM.

\subsection{Regarding generalization\label{dis:generalization}}

\begin{figure*}[t]
\begin{center}
\includegraphics[width=0.33\textwidth]{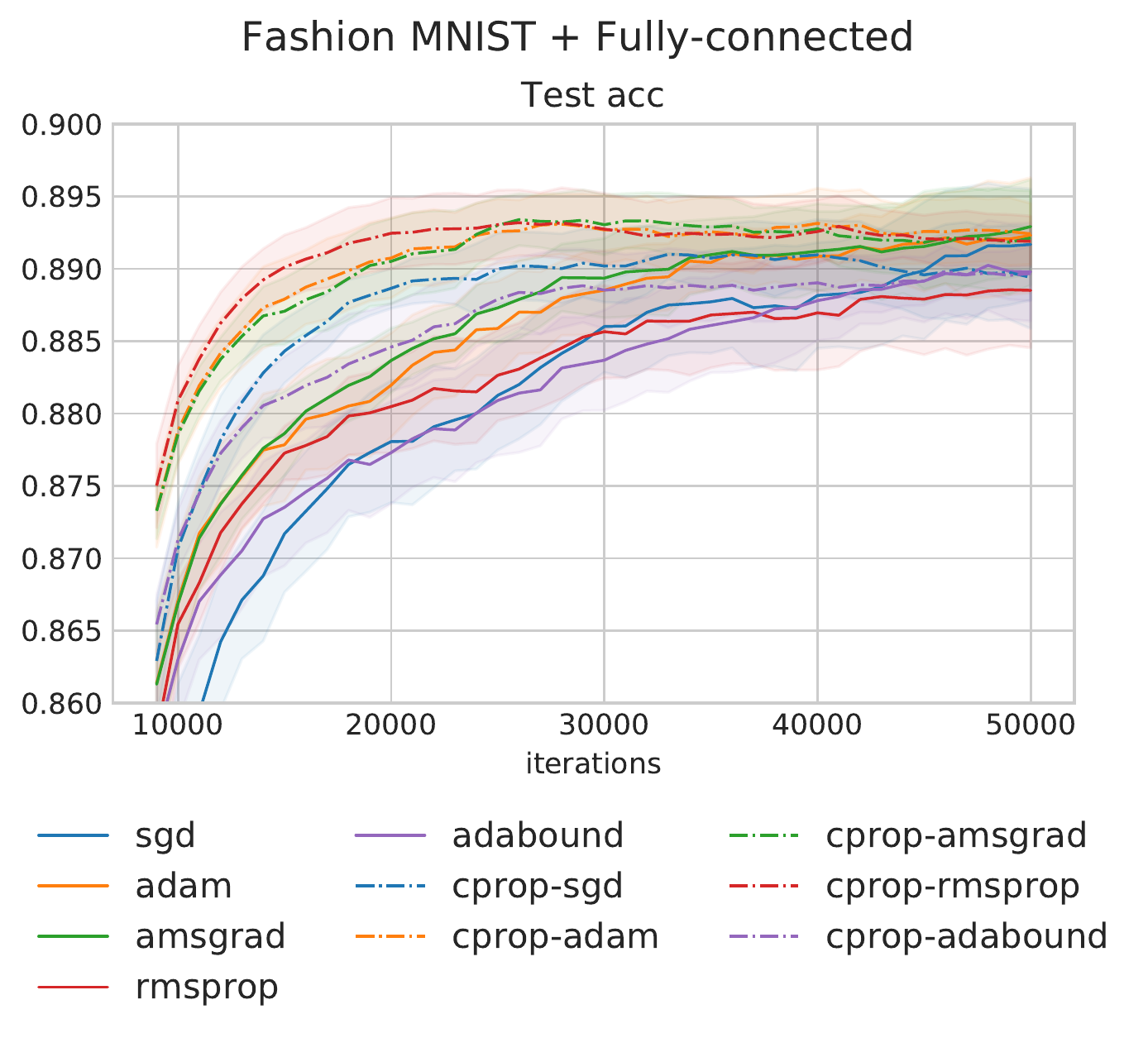}
\includegraphics[width=0.33\textwidth]{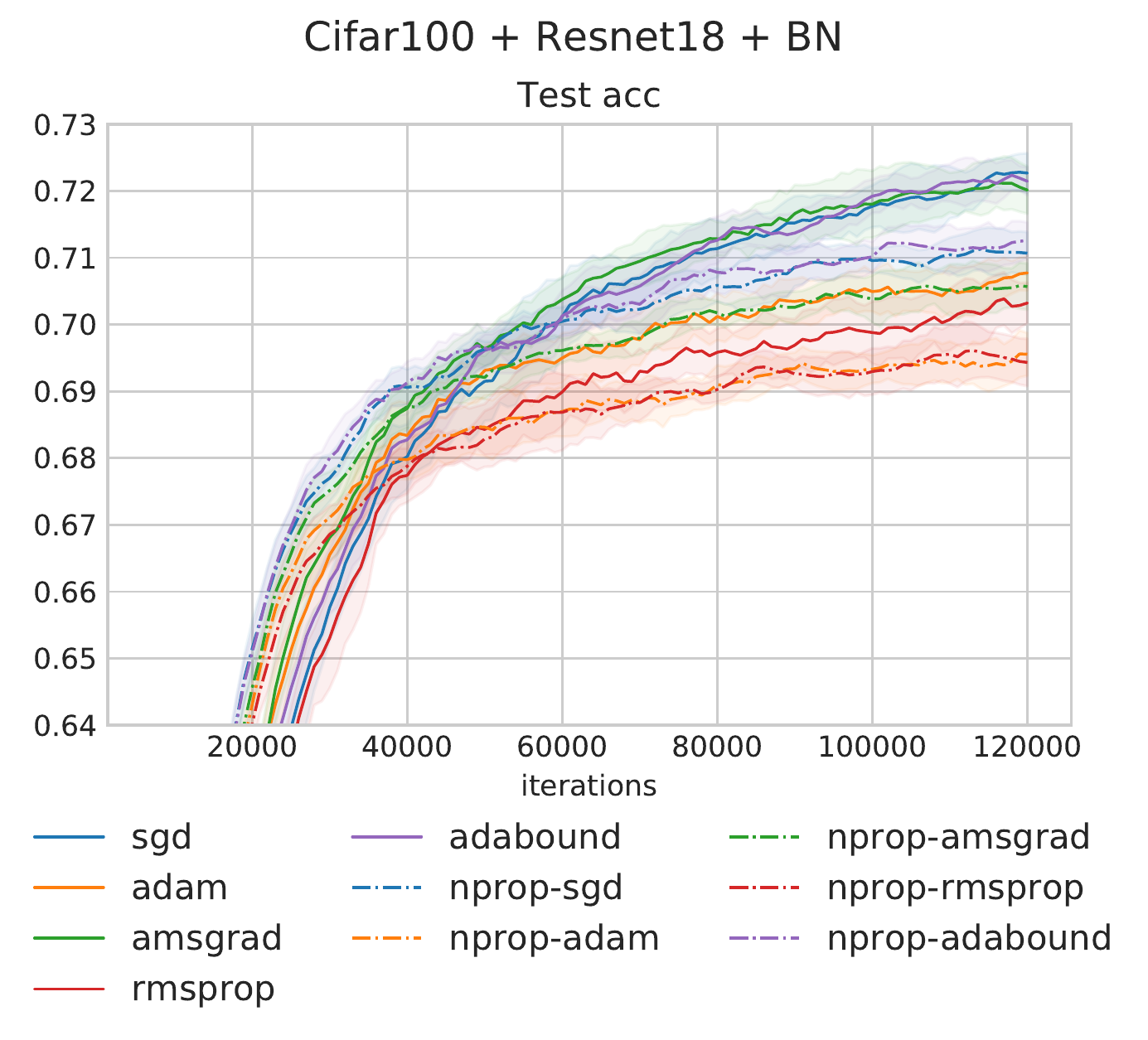}
\includegraphics[width=0.33\textwidth]{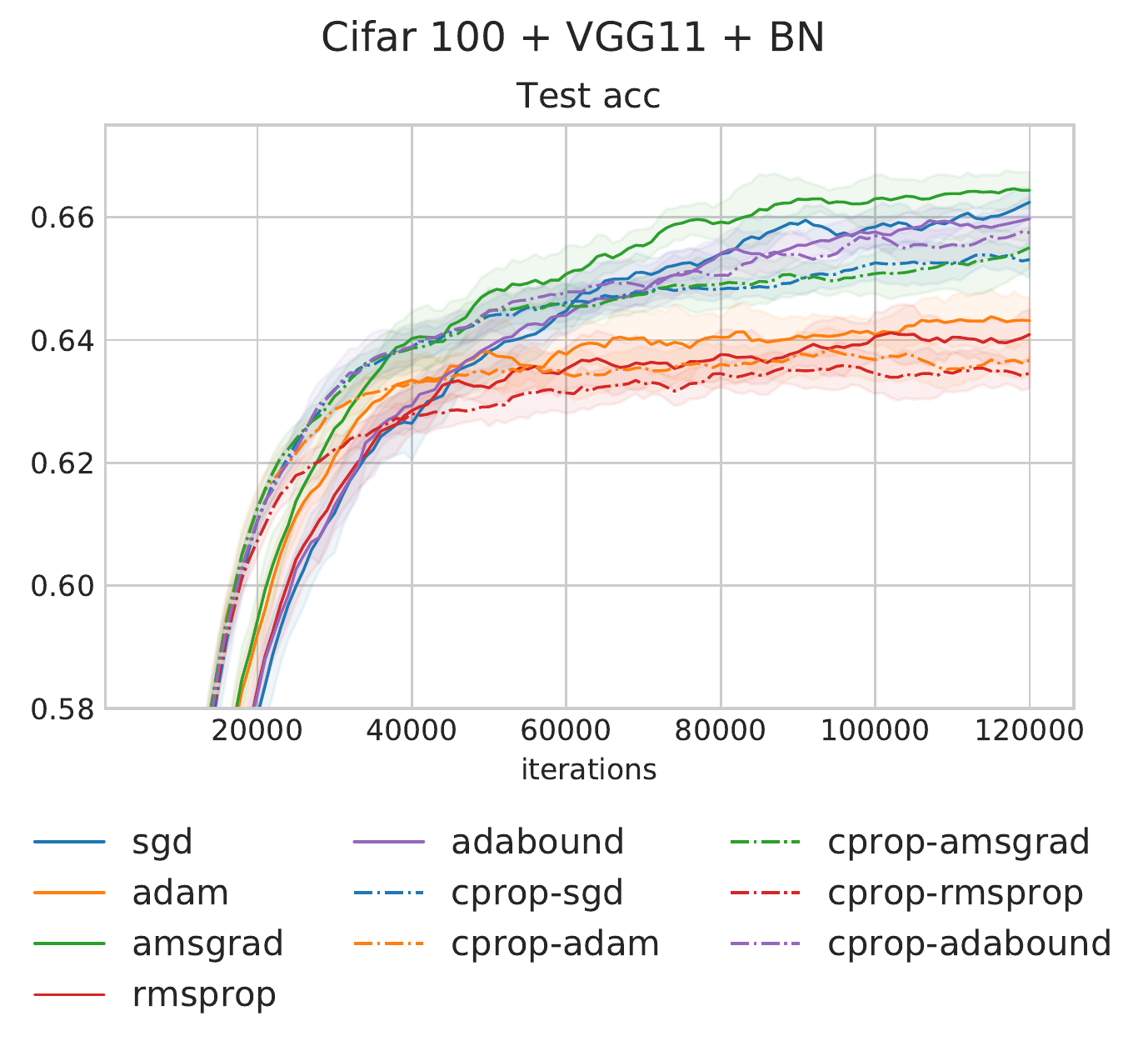}
\end{center}
\vspace{-0.7cm}
\begin{center}
\includegraphics[width=0.33\textwidth]{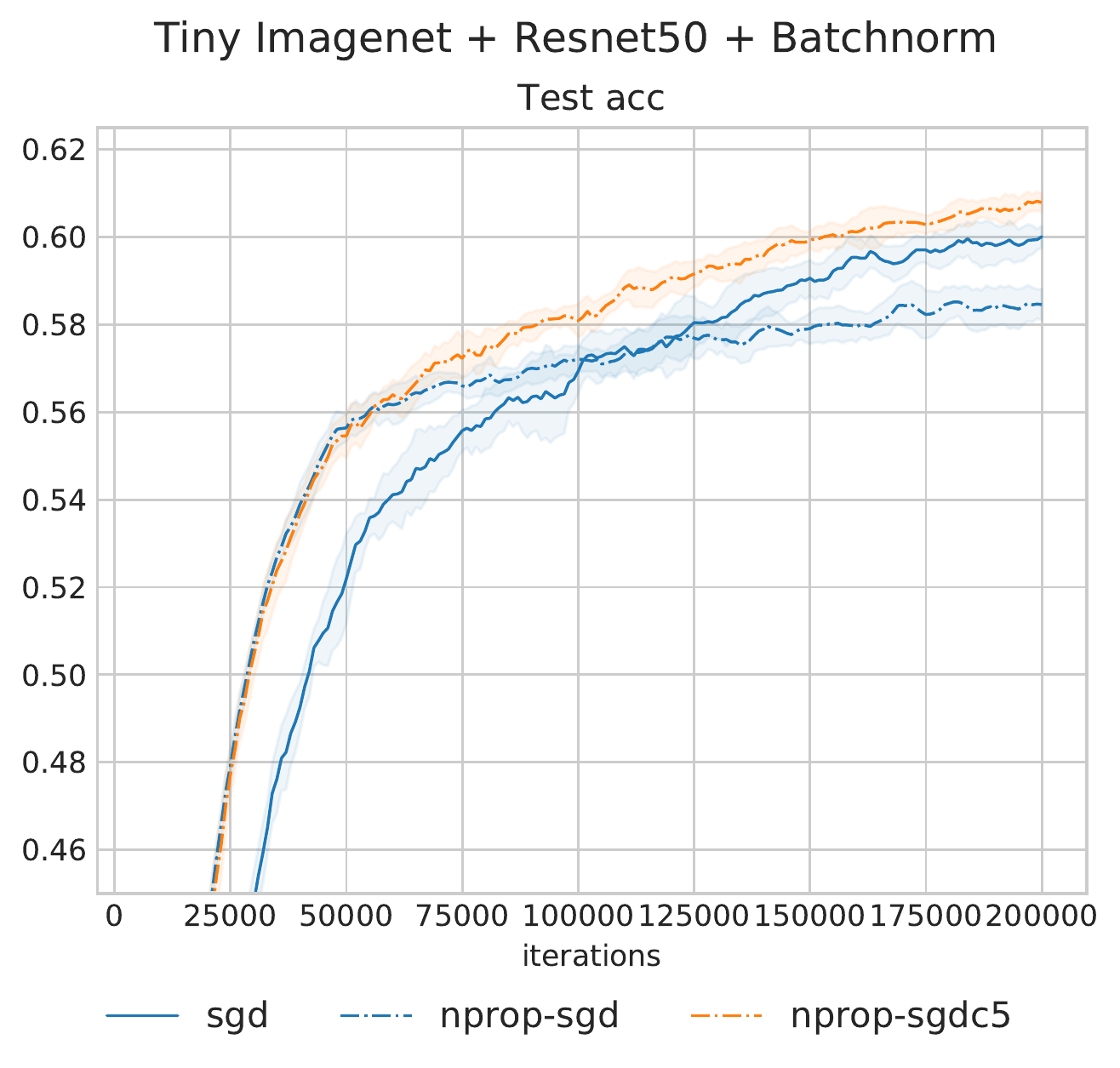}
\includegraphics[width=0.33\textwidth]{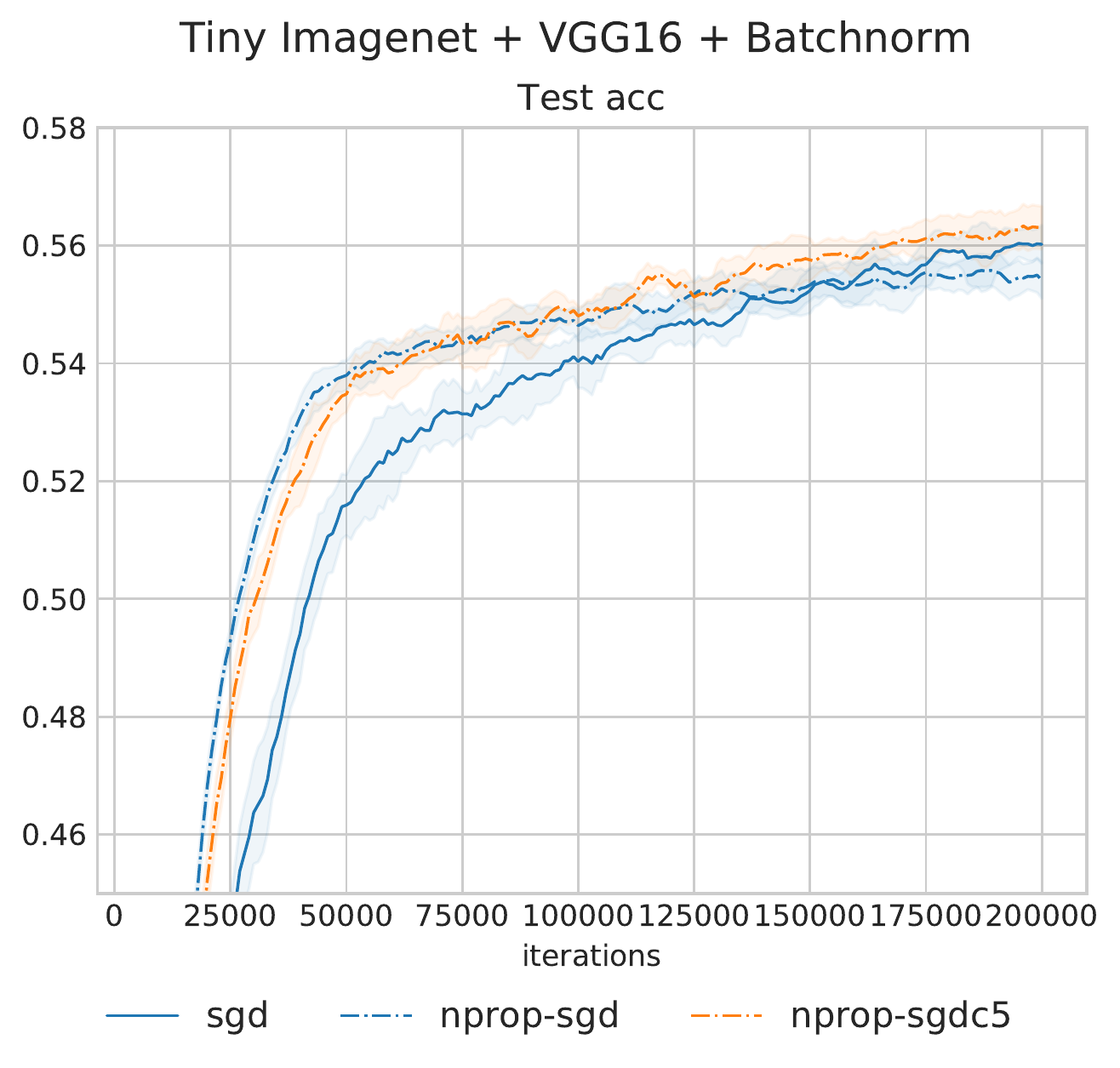}
\includegraphics[width=0.33\textwidth]{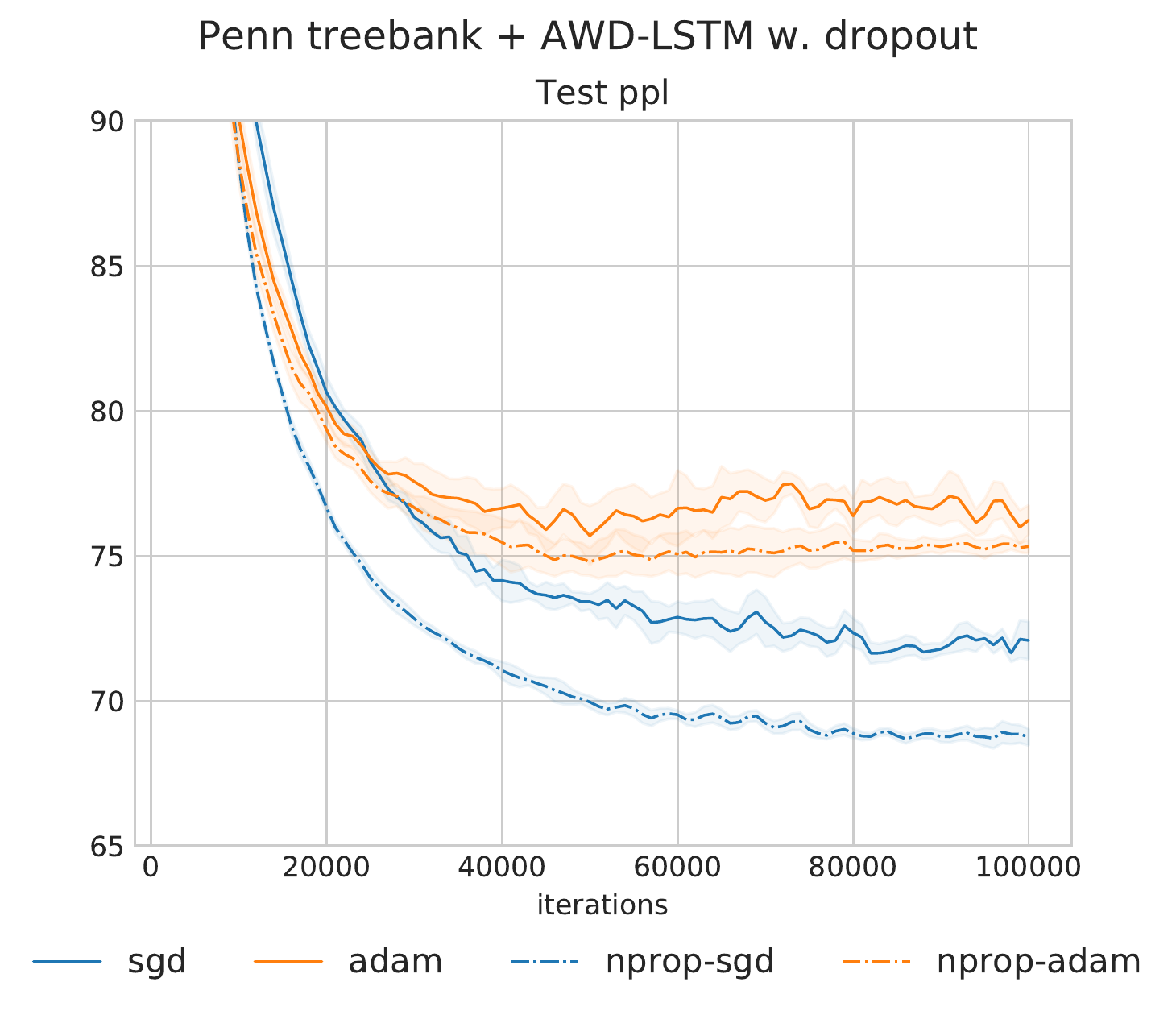}
\end{center}
\vspace{-0.25cm}
\caption{\label{fig:generalization} Generalization performance of CProp across many datasets and architectures. For language modeling, the measure is perplexity (PPL) the lower the better. In general, SGD generalized best given enough training time. With overconfidence $c$, CProp could rival SGD while sacrificing only marginal training speed.}
\vspace{-0.45cm}
\end{figure*}

\begin{figure}[t]
\centerline{
\includegraphics[width=0.25\textwidth]{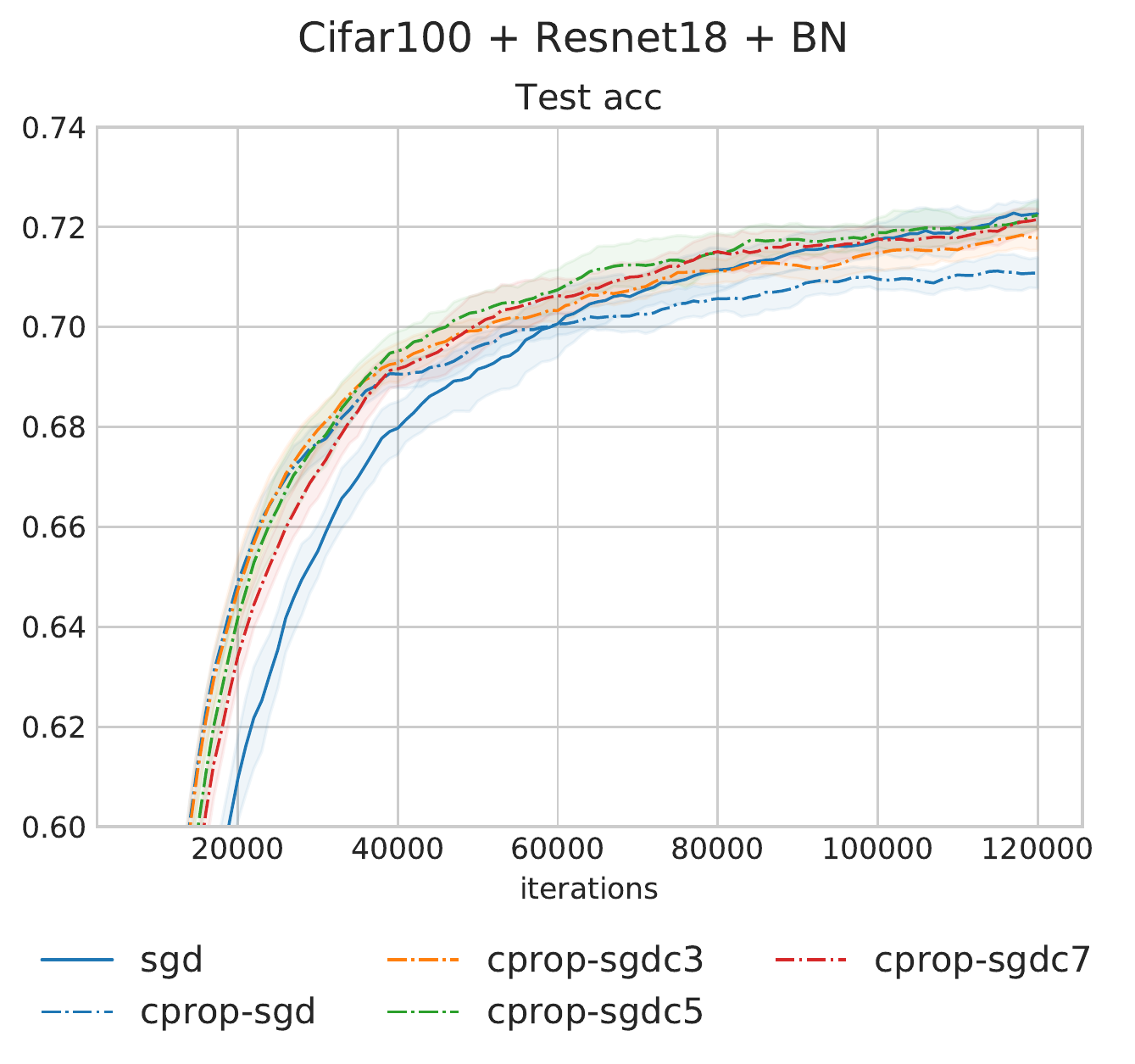}
\includegraphics[width=0.25\textwidth]{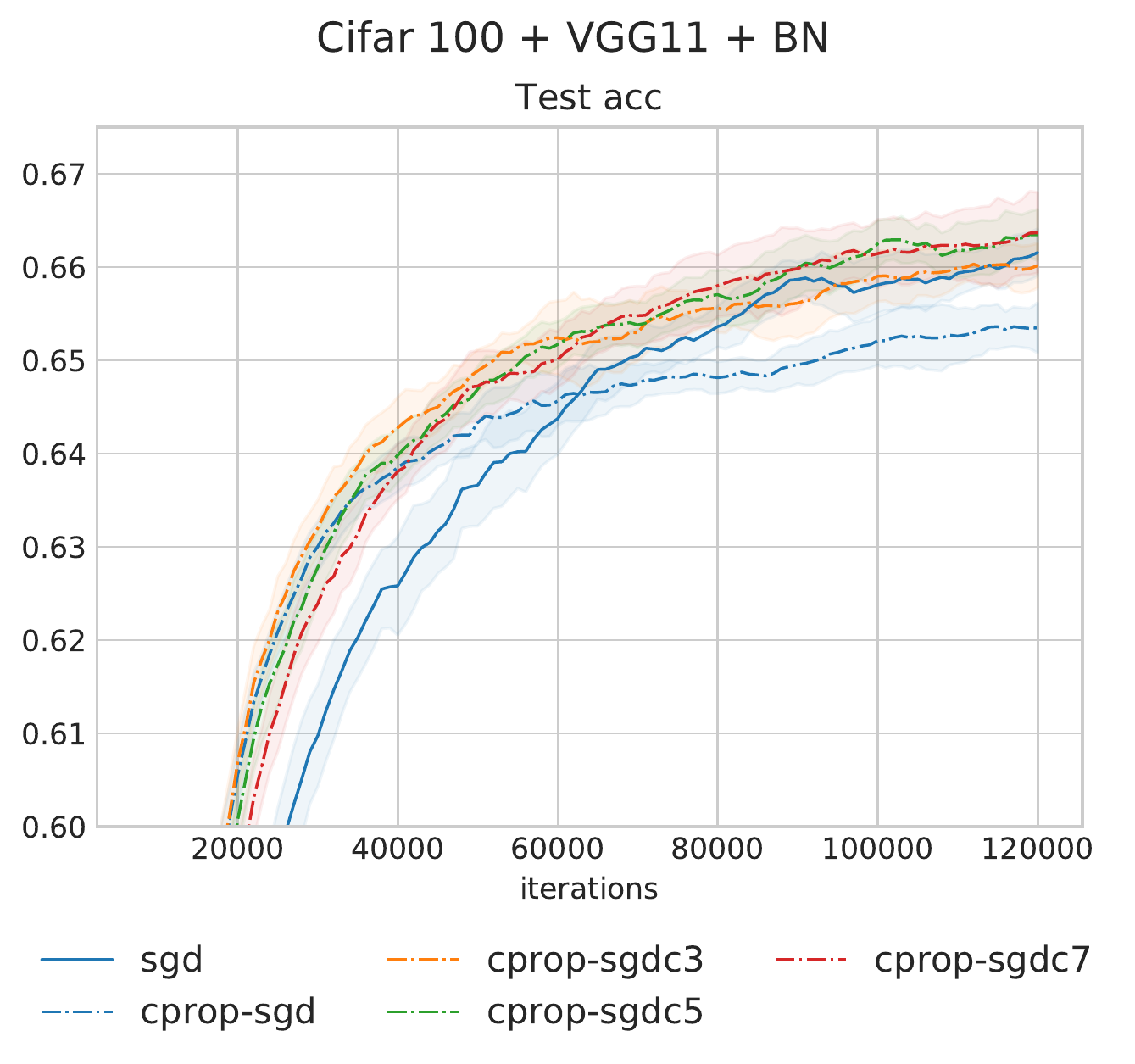}
}
\vspace{-0.1cm}
\caption{\label{fig:overconfidence} Generalization performance of overconfident CProp with varying $c$. Larger $c$'s trade off the training speed with generalization performance improving over a default $c=1$.}
\vspace{-0.25cm}
\end{figure}

Although this work makes no claim on the generalization performance i.e. test loss/accuracy. For completeness, we include those results in figure \ref{fig:generalization}. Vanilla SGD seems to generalize best in most cases while being slower to train. As CProp only scaling down, the result follows many observations that larger learning rates tend to generalize better \cite{Keskar2016-zv, Jastrzebski2018-mz, Xing2018-oa}. Underconfident CProp generalizes poorer. This opens up a possibility of either bumping up base learning rates or using the overconfidence coefficient $c > 1$. In figure \ref{fig:overconfidence}, we show that the generalization performance of CProp could be improved with higher overconfidence coefficients. 

\subsection{Performance with dropout\label{dis:dropout}}

\begin{figure}[t]
\centerline{
\includegraphics[width=0.25\textwidth]{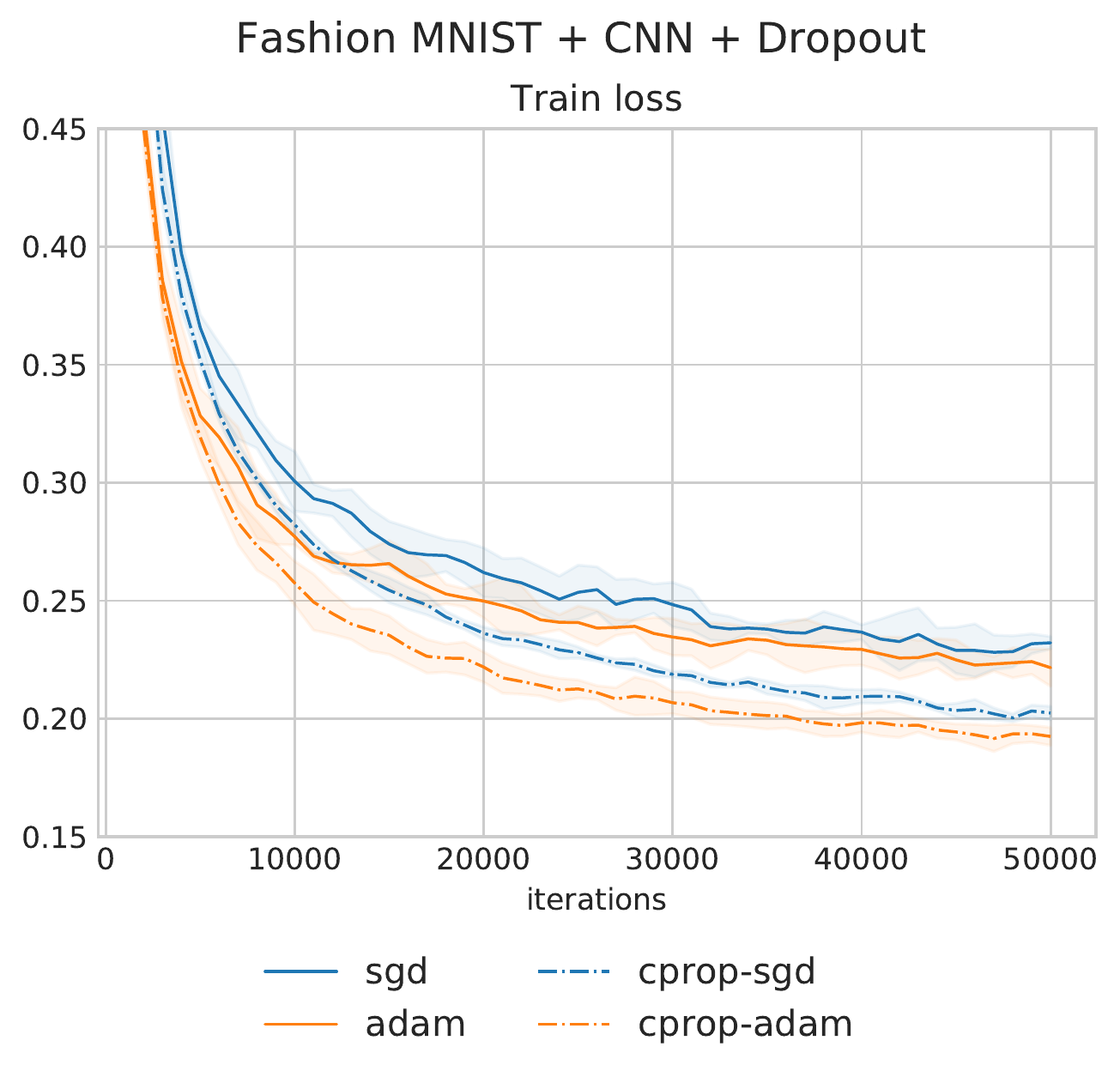}
\includegraphics[width=0.25\textwidth]{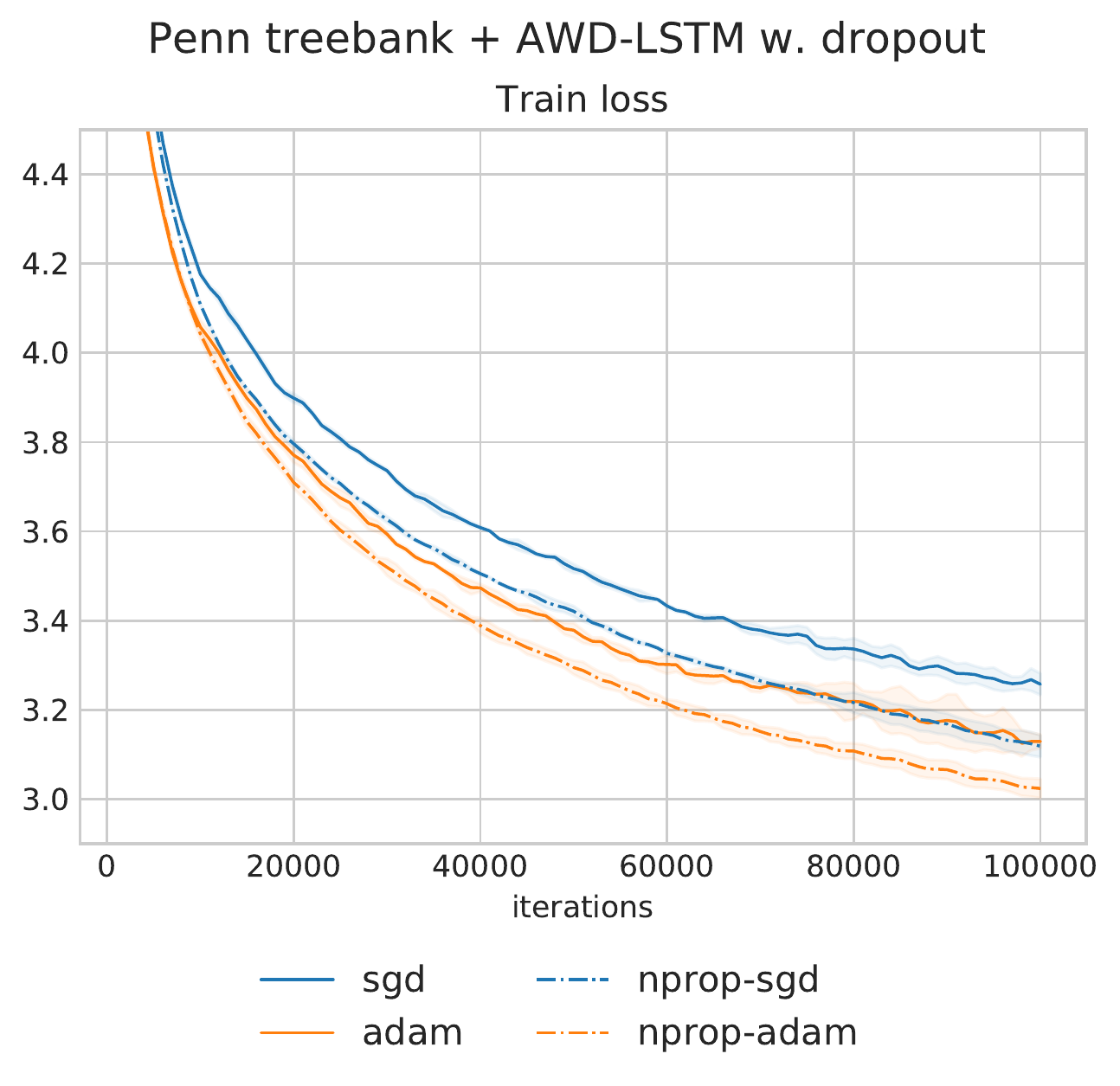}
}
\vspace{-0.1cm}
\caption{\label{fig:dropout} Performance with dropout. (Left) Dropout 0.3 with CNN architecture on Fashion MNIST. (Right) AWD-LSTM with dropout on Penn treebank. The gaps are generally narrower than without dropout counterparts.}
\vspace{-0.45cm}
\end{figure}

From unclear reasons, CProp seems to be less effective under the presence of strong dropout. We observe this effect consistently across many architectures given that the dropout is strong enough. In figure \ref{fig:dropout}, we demonstrate some of this effect. While improvement can still be seen, the gaps are narrower (and even narrower with stronger dropout). Since CProp uses the gradient signals, it follows that under a strong dropout the gradient has some characteristic rendering CProp less useful. 

\section{Related works\label{relatedwork}}

\textbf{Adaptive gradient}. There is a large literature regarding adaptive gradient. AdaGrad \cite{Duchi2011-ie} was developed to deal with sparse gradient problem. The idea is to keep reducing the learning rates of weights updated frequently. This was shown to work well in some specific tasks, but usually underestimates the learning rate in general. RMSProp \cite{Tieleman2012-rf} was inspired by RPROP\cite{Riedmiller1993-ng} but aiming for mini-batch settings. The idea is to update according the gradient signs, +1 or -1, instead of the actual gradient. The signs are derived approximately by dividing the gradients with their square roots of mean second moments. It does not multiplicatively increase/decrease learning rates as in RPROP. Adam \cite{Kingma2014-jt} is a combination of both momentum and RMSProp. Adam has seen wide adaption across many communities owing to its ease of tuning. Later, it was shown that Adam might have a convergence problem. A remedy was proposed in AMSGrad \cite{Reddi2018-mk}, and enjoys much better convergence properties. 

\textbf{Learning rate scheduling.} 
Temporally reducing learning rate is required for convergence in many optimizers. Many traditional schedulers, including inverse iteration count and exponential decay, are well known. However, the hyperparameters come with them are not easy to select. Probably the most commonly used is step decay which reduces the learning rate on predefined iterations, seen in \cite{Simonyan2014-wm,Krizhevsky2012-zm,He2016-wj} to name a few. Usually practitioners will reduce the learning rate when the validation loss plateaus. This process can be automated as in \cite{Merity2017-hd}. 
Increasing learning rate is also possible for a different reason. It has been observed that a learning rate can be drastically increased while being successful to train if it starts small and ends small. Larger learning rate makes faster progress leading to faster convergence. The start small comes from the principle of \textit{warm up} which argues that early gradients are noisy requiring lower learning rates. Combining the idea of warm up and the learning rate decay is the birth of Cyclical learning rate \cite{Smith2017-pm,Smith2018-mo} and Slanted triangular learning rate \cite{Howard2018-ao}. A closely related work is Warm Restart \cite{Loshchilov2017-xc} where the author suggests bumping up the learning rate and decays it slowly many times with different intervals.

\textbf{Optimal learning rate.} 
The optimal learning rate is the one that reduces the loss function the most. It could be derived using a second-order information \cite{Goodfellow2016-ch}. However, computing Hessian and inverting it is compute and memory intensive for large networks, and is also susceptible to noises. A more practical approximation of Hessian was proposed in \cite{Schaul2013-lf}. A pessimistic learning rate could also be derived from the maximum steepness of the loss landscape, the Lipschitz constant. Estimation of this constant is possible but not trivial as proposed in \cite{Yedida2019-my}. 
An iterative approach has been propsed in RPROP \cite{Riedmiller1993-ng}. It takes two consecutive gradients and decide whether to increase/decrease the learning rate. RPROP works in batch settings and was shown to work in most settings with minimal tuning \cite{Igel2003-ts}. To the best of our knowledge, there is no variant that works on mini-batch settings. 

\textbf{Model averaging.} Lookahead \cite{Zhang2019-tv} is an optimizer based on the idea of model averaging. It maintains two sets of parameters, fast and slow. The fast weights are updated according to an optimizer of choice while the slow weights slowly track the fast weights. This allows Lookahead to be less susceptible to instabilities of the fast weight by reducing variance. 

\textbf{Generalization gap.} It has been observed that Adam, including many related adaptive gradient methods, is inferior to SGD in terms of generalization \cite{Wilson2017-gu}. An attempt to mitigate this effect while preserving the convergence speed of Adam is to start with adaptive gradient and end with SGD. The idea was explored by \cite{Keskar2017-rm} with a fixed step transition between Adam and SGD which was later extended to smooth transition seen in AdaBound \cite{Luo2019-kx}.

\section{Conclusion}

We propose a learning rate scaling method, CProp, which takes a look on the past gradients, and derives the scaling from how much the past gradients conform. The conformity is measured using a probabilistic framework by determining how confidence it is that the past gradients collectively have a certain sign, either positive or negative. Via the conformity, CProp can effectively adjust the learning rate to make faster progress than traditional optimizers. This principle is shown to be useful across many tasks and architectures. Since the scaling itself is just another term, it could be used easily with any kind of optimizer. 

% \section{Acknowledgement}
% Many thanks to Nat Dilokthanakul on his careful thoughts on the line of arguments of the paper. We have reflowed the explanation with more clarity. This paper was improved so much owing to his suggestions. 
% Ekapol Chuangsuawanich helped fill in some theoretical grounds on the paper arguments and pointed out where to improve. 
% These people were our pre-reviewer who contributed their precious time to our work. 

{\small
\bibliographystyle{ieee_fullname}
\bibliography{ms}
}

\clearpage

\onecolumn
\section*{Supplementary Materials}
\appendix
\section{Network architectures}

\subsection{Fashion MNIST's Fully-connected architecture\label{arch:fc}}

\textbf{Layers.} 784, 300, 100, 10. After each layer is ReLU activation except the last output layer.

\subsection{Fashion MNIST's CNN architecture\label{arch:cnn}}
\textbf{Convolutional layers.} 16, M, 32, M, 64, M, 128. Each number is the number of output channels from the convolution layer with 3x3 kernels (padding = 1). Following each layer are ReLU activation and dropout \cite{Srivastava2017-xi} (channel-wise) in order. Note that dropouts are optional depending on the experiment. \textit{M} denotes a 2x2 maxpooling layer.

\textbf{Classification layers.} We use a global average pooling layer followed by an 1x1 convolution layer to output the logit of 10 classes.

\subsection{Cifar's VGG11 architecture\label{arch:vgg11}}
This is based on VGG11 only with smaller classification layers for the smaller dataset.

\textbf{Convolutional layers.} 64, M, 128, M, 256, 256, M, 512, 512, M, 512, 512, M. Using the same convention as \ref{arch:cnn}. Note that batch normalization is applied after each layer except the output layer.

\textbf{Classification layers.} A global average pooling layer is used followed by a fully-connected layer to output the logits of 100 classes. 

\subsection{Cifar's Resnet18 architecture\label{arch:resnet18}}

This is based on Resnet18 only with smaller classification layers for the smaller dataset.

\textbf{Resnet layers.}  64, [64, 64], [64, 64], [128D, 128], [128, 128], [256D, 256], [256, 256], [512D, 512], [512, 512]. Each number denotes the number of output channels from 3x3 convolution (padding=1). A bracket denotes a residual block i.e. the input of the block is added to its output. If the input and output are not compatible, 1x1 convolution is applied to the shortcut either by changing the number of channels or by stride=2 to reduce the size.  \textit{D} denotes a stride=2. Batch normalization and ReLU are applied after each convolution layer. 

\textbf{Classification layers.} A global average pooling layer is used followed by a fully-connected layer to output the logits of 100 classes. 

\subsection{Tiny Imagenet's VGG16 architecture\label{arch:vgg16}}

This is based on VGG16 only with smaller classification layers for the smaller dataset.

\textbf{Convolutional layers.} 64, 64, M, 128, 128, M, 256, 256, 256, M, 512, 512, 512, 'M', 512, 512, 512, M. Using the same convention as \ref{arch:cnn}. Note that batch normalization is applied after each layer except the output layer.

\textbf{Classification layers.} A global average pooling layer is used followed by a fully-connected layer to output the logits of 200 classes. 

\subsection{Tiny Imagenet's Resnet50 architecture\label{arch:resnet50}}

This is based on Resnet50 only with smaller classification layers for the smaller dataset.

\textbf{Resnet layers.}  64, [64I, 64, 256I]$\times 3$, [128I, 128D, 512I], [128I, 128, 512I]$\times 2$, [256I, 256D, 1024I], [256I, 256, 1024I]$\times 2$, [512I, 512D, 2048I], [512I, 512, 2048I]$\times 2$. Using the same convention as \ref{arch:resnet18}. \textit{I} denote a 1x1 convolution layer. 

\textbf{Classification layers.} A global average pooling layer is used followed by a fully-connected layer to output the logits of 200 classes. 

\subsection{AWD-LSTM for Penn treebank\label{arch:ptb-awd-lstm}}

We used two kinds of networks: with and without dropout. AWD-LSTM was proposed with strong dropout, we created another version of it by setting dropouts to zero while keeping everything else intact. We call this AWD-LSTM wo. dropout. 

We follow the AWD-LSTM strictly. We refer to the original paper \cite{Merity2017-hd} for more details. We used a three-layer LSTM with 400, 1150, 400 units respectively. The hyperparameters are provided in table \ref{tab:awd-lstm-hyper}. Note that AWD-LSTM wo. dropout is the same with dropout parameters set to zero.

\begin{table}[tbh]
    \centering
    \caption{
    \label{tab:awd-lstm-hyper} AWD-LSTM w. dropout hyperparameters.
    }
    \begin{tabular}{ccc}
         \toprule
         \textbf{Hyperparameter} & \textbf{Value} \\
         \midrule
         Embedding dim. & 400  \\
         LSTM units & 1150  \\
         LSTM layers & 3  \\
         Gradient norm clip & 0.25 \\
         Embedding dropout & 0.1  \\
         Input dropout & 0.4 \\
         LSTM layer dropout & 0.25 \\
         Output dropout & 0.4 \\
         LSTM weight dropout & 0.5 \\
         Alpha & 2 \\
         Beta & 1 \\
         Split cross-entropy & Not used \\
         \bottomrule
    \end{tabular}
\end{table}

% \subsection{Wiki2's LSTM architecture\label{arch:wiki2-lstm}}

% A two-layer LSTM with 300 hidden units. Embedding of size 300. 

% \subsection{Wiki103's LSTM architecture\label{arch:wiki103-lstm}}
\section{Optimizer hyperparameters\label{optimizer_params}}
\begin{itemize}
\item \textbf{SGD.} No momentum applied.
\item \textbf{Adam and AMSGrad.} $\beta_1 = 0.9, \beta_2 = 0.999, \epsilon=10^{-8}$.
\item \textbf{RMSProp.} $\alpha=0.99, \epsilon=10^{-8}$ with no momentum.
% \item \textbf{Lookahead.} $k=5, \alpha=0.5$. $\alpha$ refers to the slow weight learning rate.
\item \textbf{AdaBound.} $\beta_1 = 0.9, \beta_2 = 0.999, \epsilon=10^{-8}$ which are the default parameters. The convergence speed of the bound is determined by $\beta_2$ as a convention to the original paper \cite{Luo2019-kx}.
\item \textbf{CProp variants.} $\beta=0.999, c=1, \epsilon=10^{-8}$. 
\end{itemize}
We did not use weight decay.  
\section{Dataset }

\subsection{Fashion MNIST}

We use the batch size of 32. We did not do any preprocessing for Fashion MNIST. 

\subsection{Cifar 100}
We use the batch size of 32. We applied augmentations to the training data as follows: random crop of size 32x32 with zero padding of size 4, random horizontal flip, and normalize by means and variances channel-wise. 

\subsection{Tiny imagenet}
We use the batch size of 32. We applied augmentations to the training data as follows: random crop of size 64x64 with zero padding of size 8, random horizontal flip, and normalize by means and variances channel-wise. 

\subsection{Penn treebank}
The datasets is chopped into 32 chunks of equal sizes where 32 is our batch size. The datasets is further chopped temporally into chunks of 35 which corresponds to the truncated backpropagation length. This applied to all datasets, both train and test. Note that Penn treebank also has a separate validation dataset which was not used in our experiments.

\textbf{Regarding truncated backpropagation.} A forward pass is done over a sequence of length 35 at a time. The backpropagation is done over that length as well. However, the hidden states of the model are carried over to the next forward pass while not being backpropagated through. This is repeated to the end of dataset. 
\section{Comparing to possible alternatives\label{alternatives}}

\begin{figure}[h]
\centerline{\includegraphics[width=0.8\textwidth]{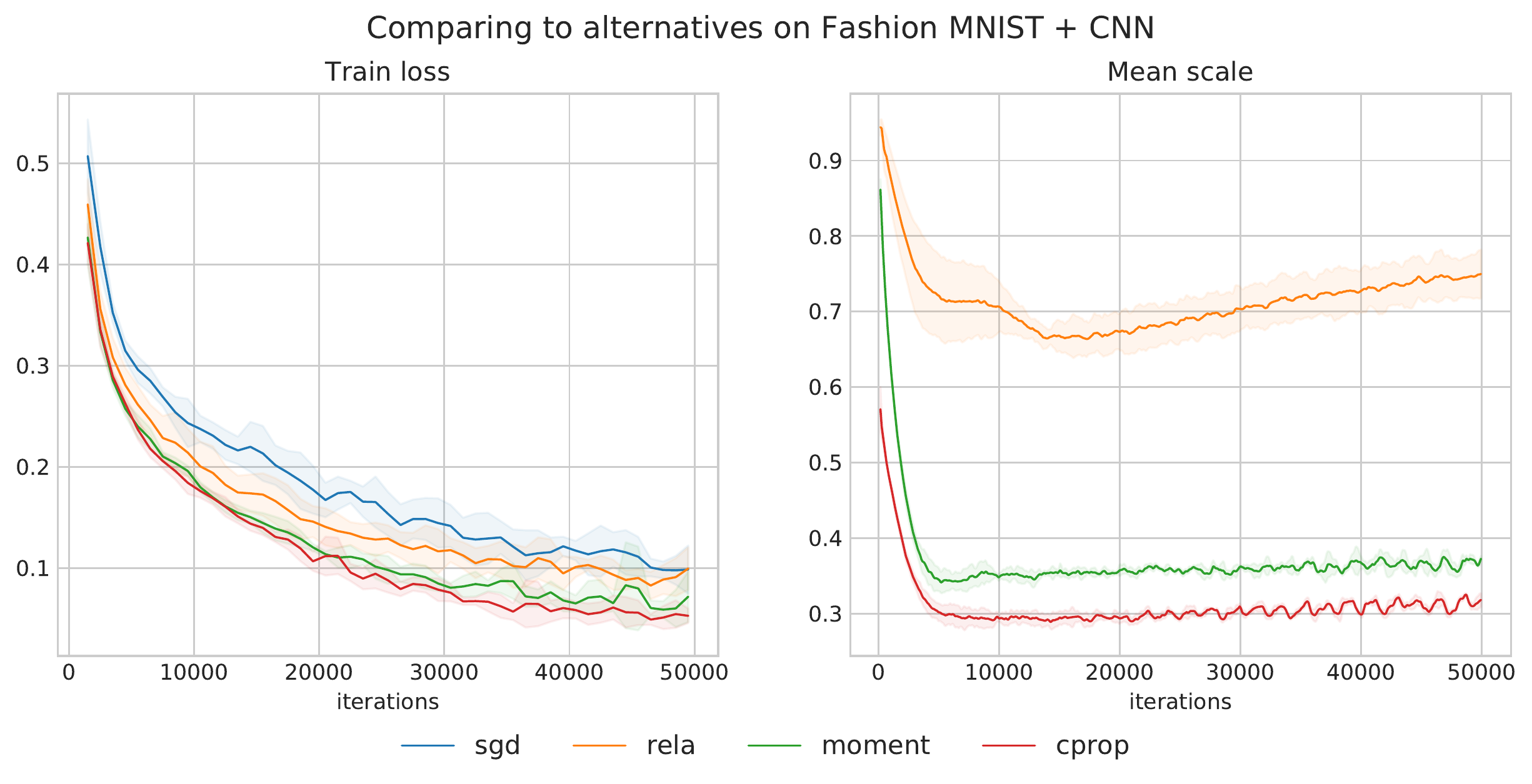}}
\vspace{-0.15cm}
\caption{\label{fig:fmnist-alternatives} Comparing CProp with relative rate and moment rate. Results on Fashion MNIST with CNN architecture. Each plot is from 3 different seeds.}
\vspace{-0.25cm}
\end{figure}

\begin{figure}[h]
\begin{center}
\includegraphics[width=0.8\textwidth]{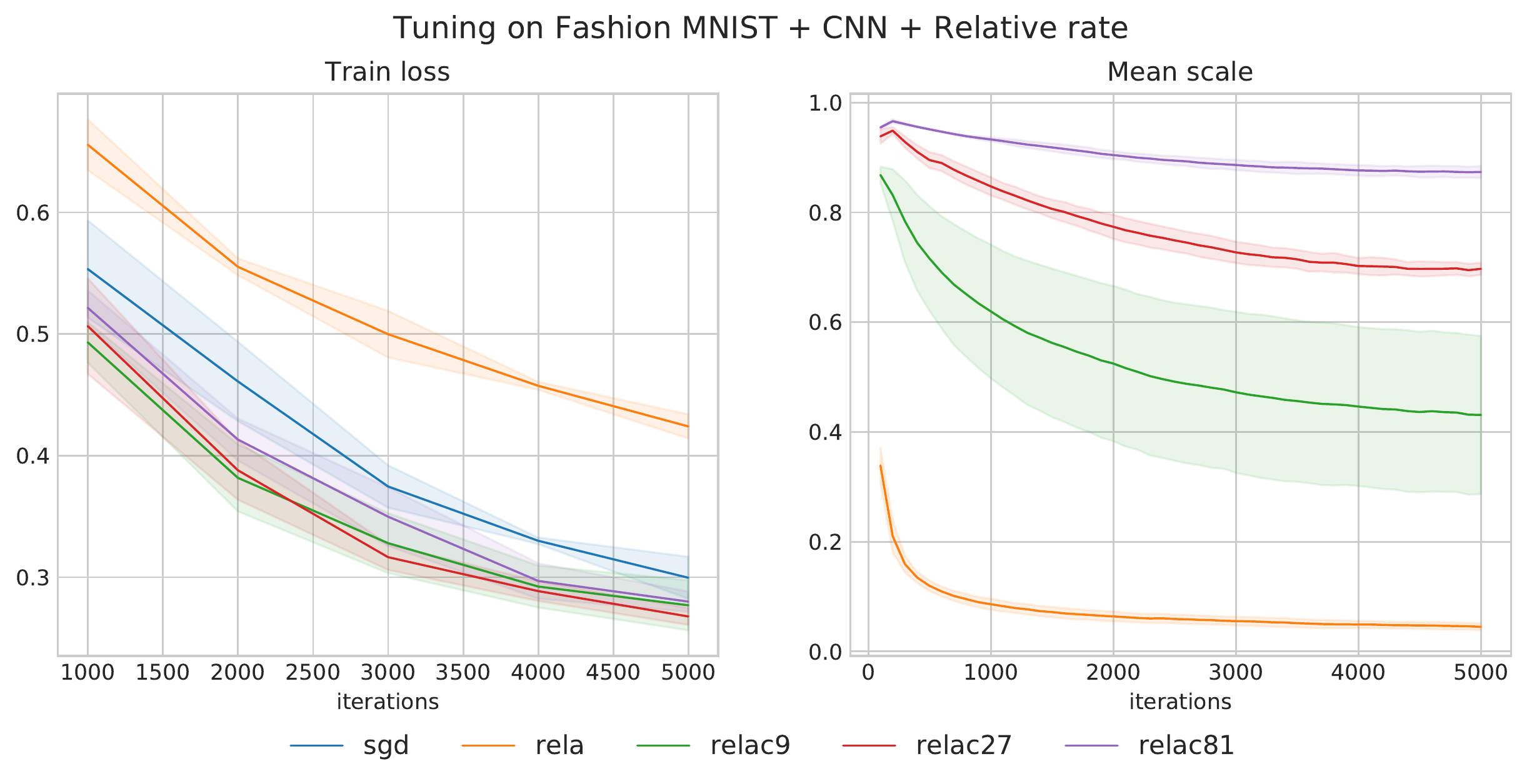}
\includegraphics[width=0.8\textwidth]{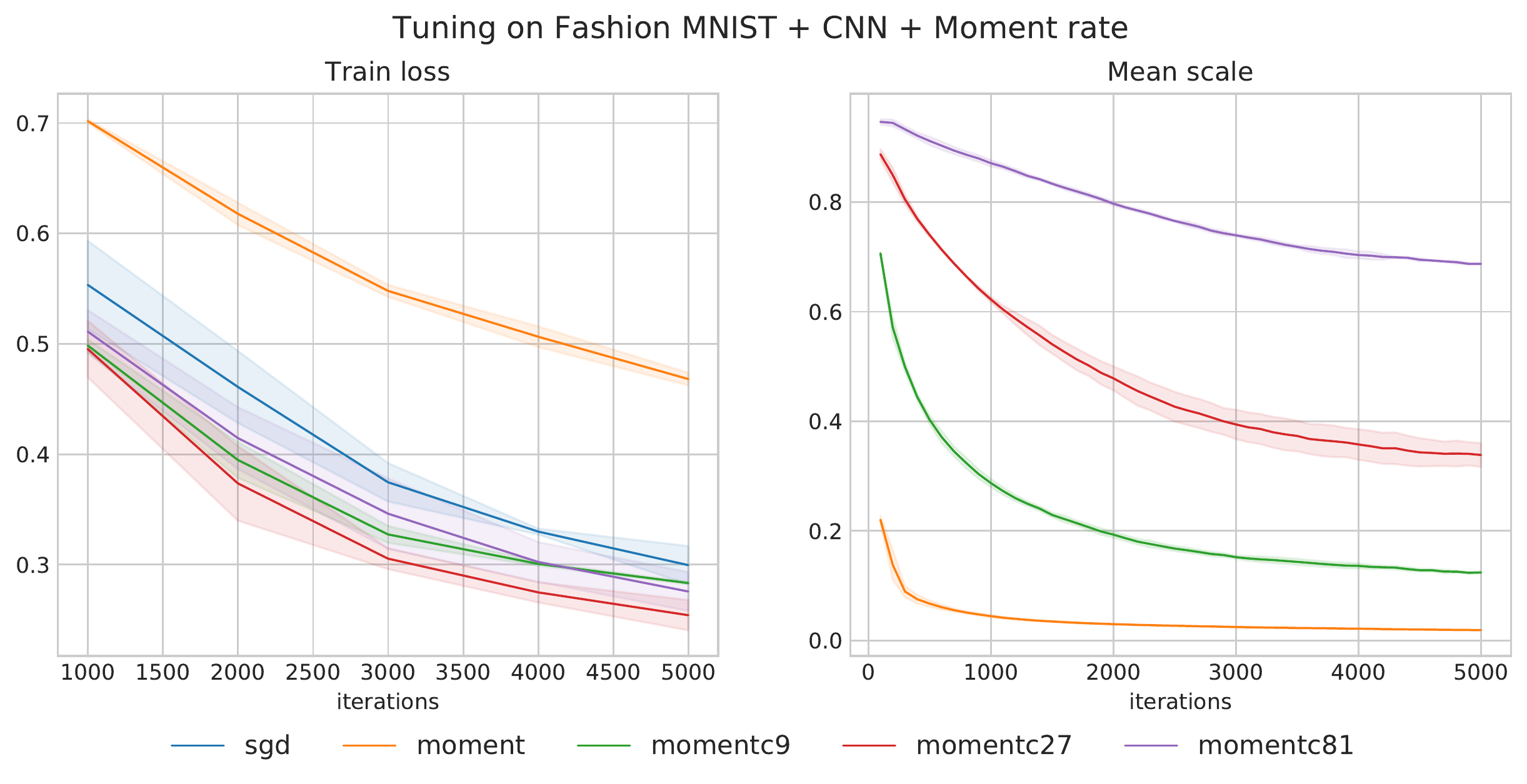}
\end{center}
\vspace{-0.15cm}
\caption{\label{fig:alternatives-tune} Tuning $c$ for relative rate and moment rate. We start from $c=1$ multiplicatively increase it by 3 until it gets worse. Each plot was run for 3 different seeds. We have $c=27$ for both methods although the plots are not very clear. Note that without proper tuning these methods performly poorly even compared to normal SGD.}
\vspace{-0.25cm}
\end{figure}

Here we show the results from alternatives methods mentioned in the main paper, namely \textbf{relative rate} and \textbf{moment rate}. We need to tune $c$ (overconfidence coefficient) for both methods due to their different characteristics e.g. moment rate tends to underestimate the learning rate, a larger learning rate needed to counteract this effect. For CProp, we did not 
tune it, using the default $c=1$. The tuning results are shown in figure \ref{fig:alternatives-tune}. We compared the three methods on Fashion MNIST with CNN architecture in figure \ref{fig:fmnist-alternatives}. Despite CProp not being tuned, it held a leading position while moment rate (with extensive tuning) came close to it.

\section{Learning rate selection\label{lr_selection}}

We performed a systematic tuning for learning rate for each setting (architecture and dataset). The optimal learning rate is one that reduces the training loss fastest, hence we do not seek a learning rate that works well in the long run (which usually underestimates)\footnote{This does not undermine the validity our results because one could do even better by using multiple steps of learning rates which capture the fast win of large learning rate and long-term win from the lower learning rate.}. Our procedure begins by guessing a learning rate, we then both multiplicatively increase and decrease the initial learning rates by the factor of 3 until no further improvement could be made. We run with 3 different seeds to give estimates of the variance. With batch normalization, learning rates are less sensitive our choice might seem a bit subjective. Here we plot the learning rate search results for each architecture and dataset.

We did not perform learning rate selection for AMSGrad and AdaBound because these two easily share learning rate with other optimizers. AMSGrad shares Adam's learning rate. AdaBound shares the start learning rate with Adam and shares the final learning rate with SGD. 

\begin{figure}[t]
\centerline{
\includegraphics[width=0.33\textwidth]{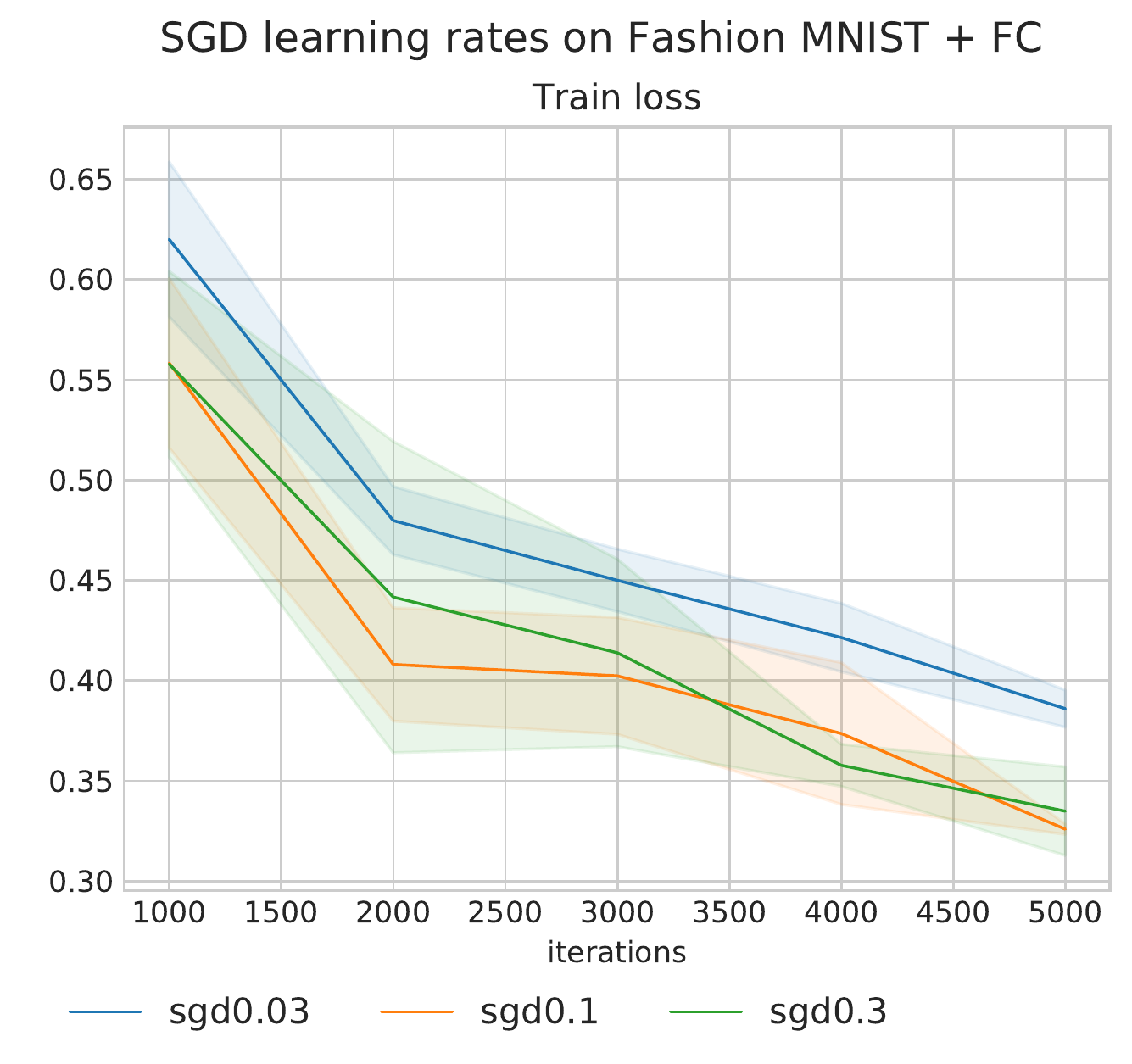}
\includegraphics[width=0.33\textwidth]{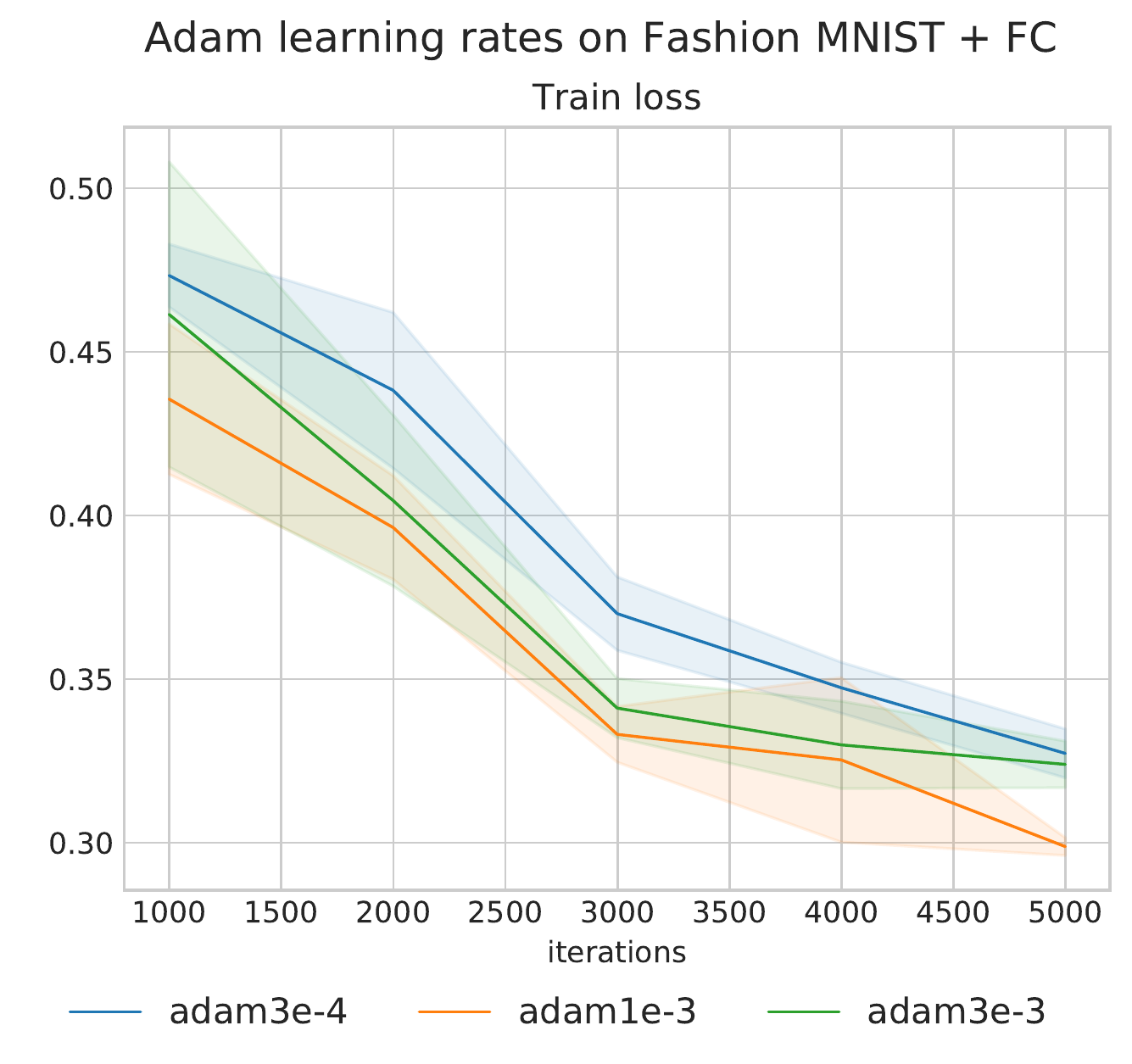}
\includegraphics[width=0.33\textwidth]{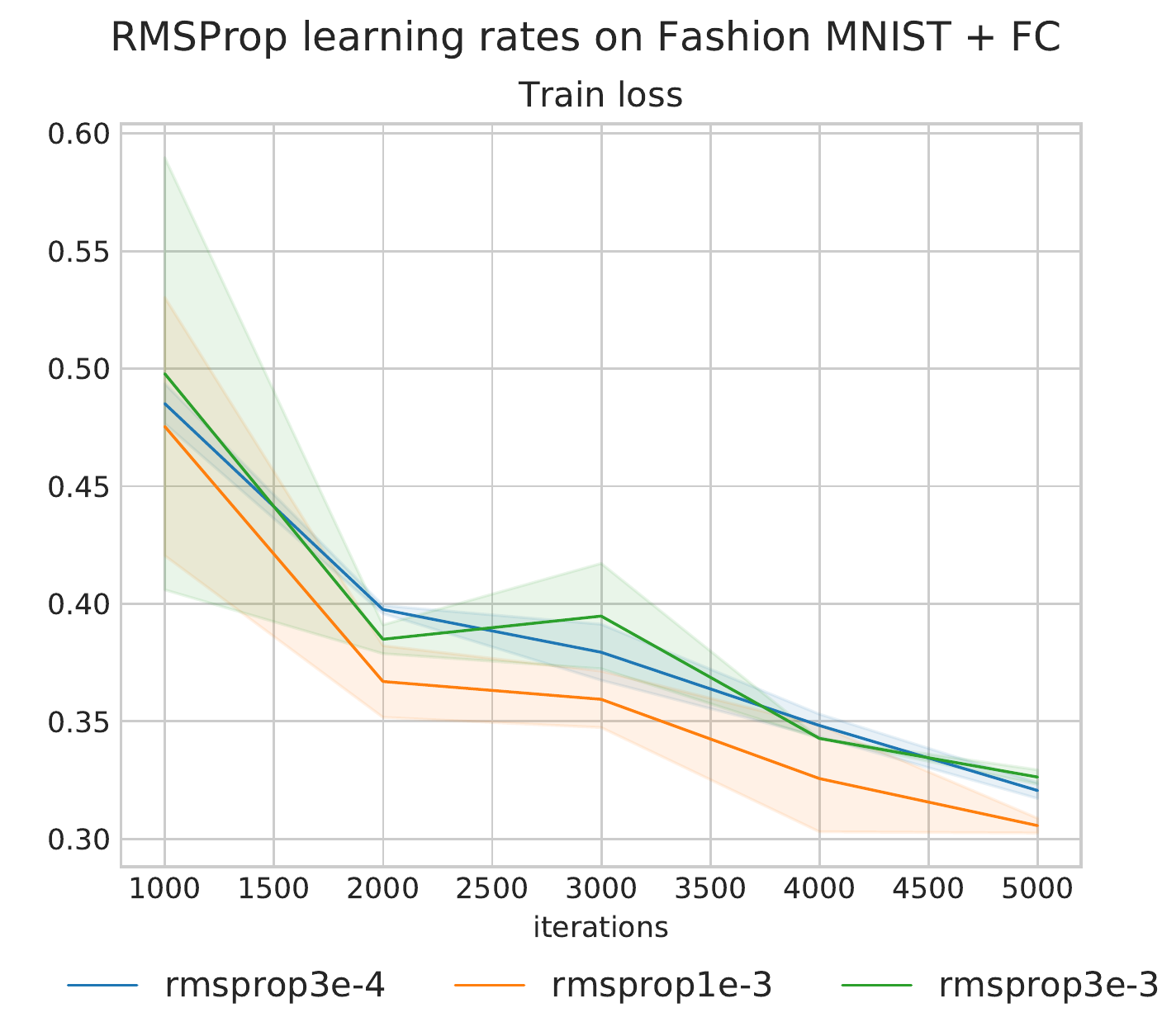}
}
\vspace{-0.15cm}
\caption{\label{fig:lr-fmnist-fc} Learning rate tuning results for Fashion MNIST with Fully-connected. We chose 0.1 for SGD, 1e-3 for Adam, and 1e-3 for RMSProp.}
\vspace{-0.15cm}
\end{figure}

\begin{figure}[t]
\centerline{
\includegraphics[width=0.33\textwidth]{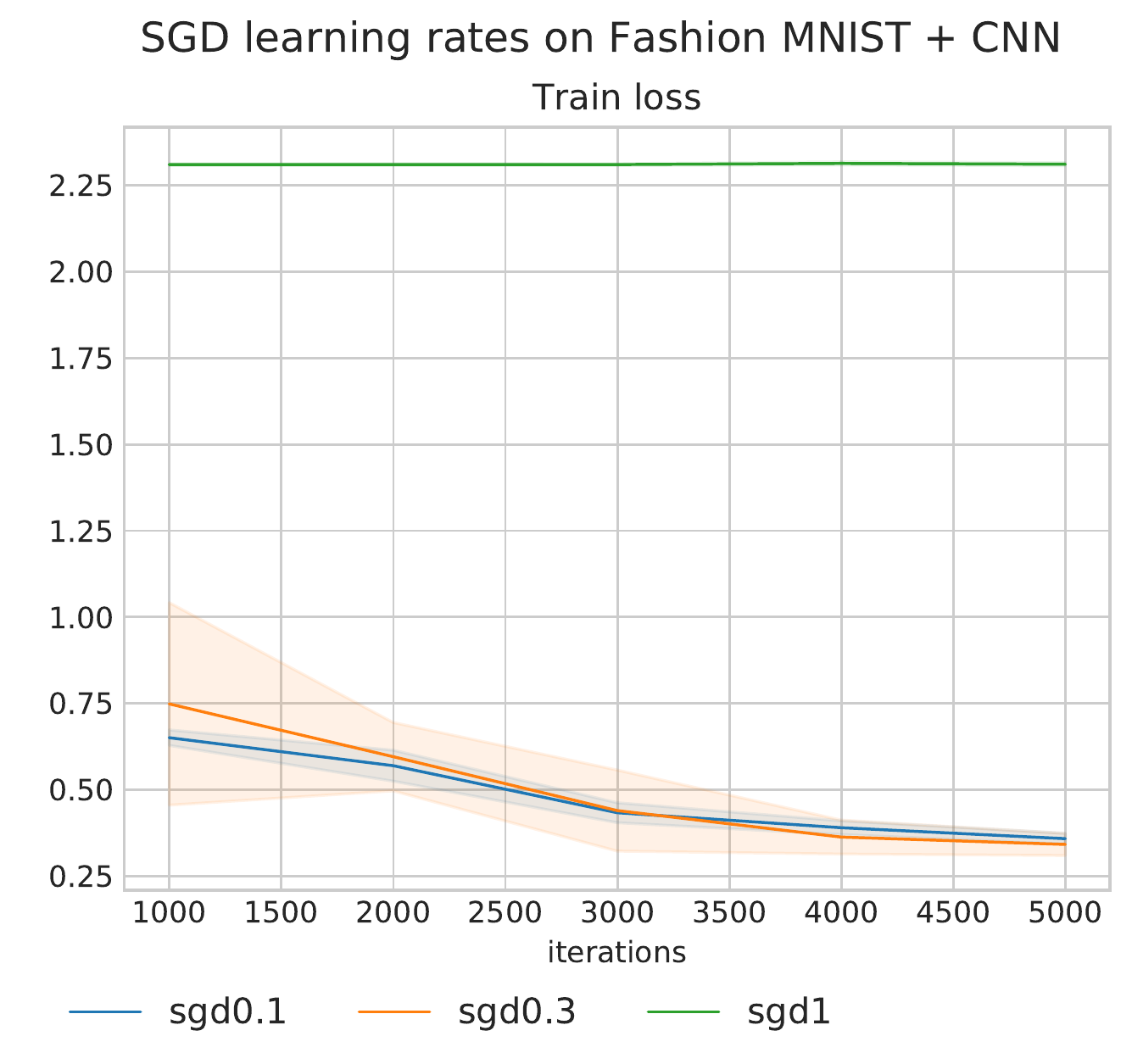}
\includegraphics[width=0.33\textwidth]{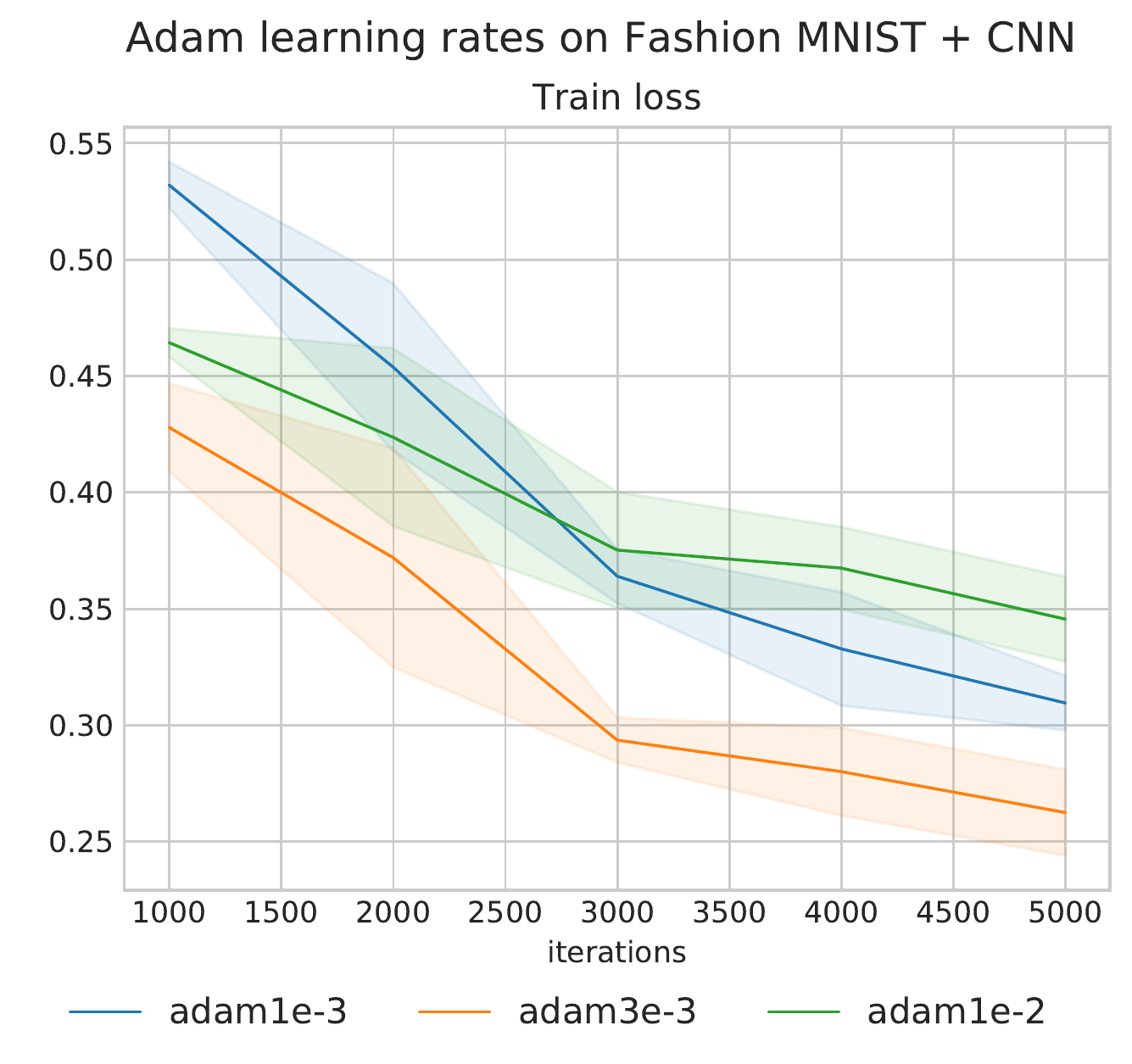}
\includegraphics[width=0.33\textwidth]{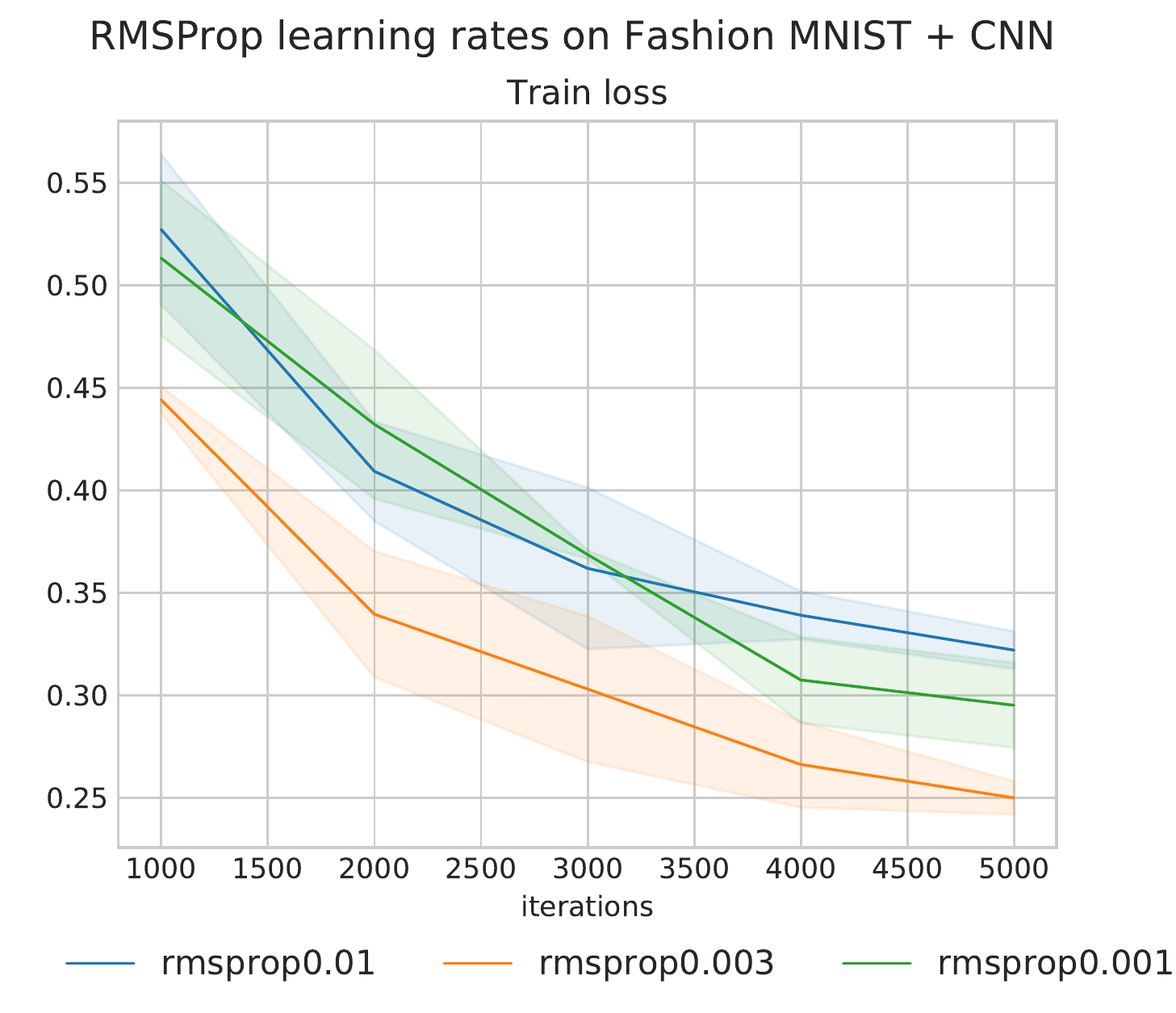}
}
\vspace{-0.15cm}
\caption{\label{fig:lr-fmnist-cnn} Learning rate tuning results for Fashion MNIST with CNN. We chose 0.3 for SGD, 3e-3 for Adam, and 3e-3 for RMSProp.}
\vspace{-0.15cm}
\end{figure}

\textbf{Fashion MNIST}: Fully-connected (see figure \ref{fig:lr-fmnist-fc}), CNN (see figure \ref{fig:lr-fmnist-cnn}. 

\begin{figure}[t]
\centerline{
\includegraphics[width=0.33\textwidth]{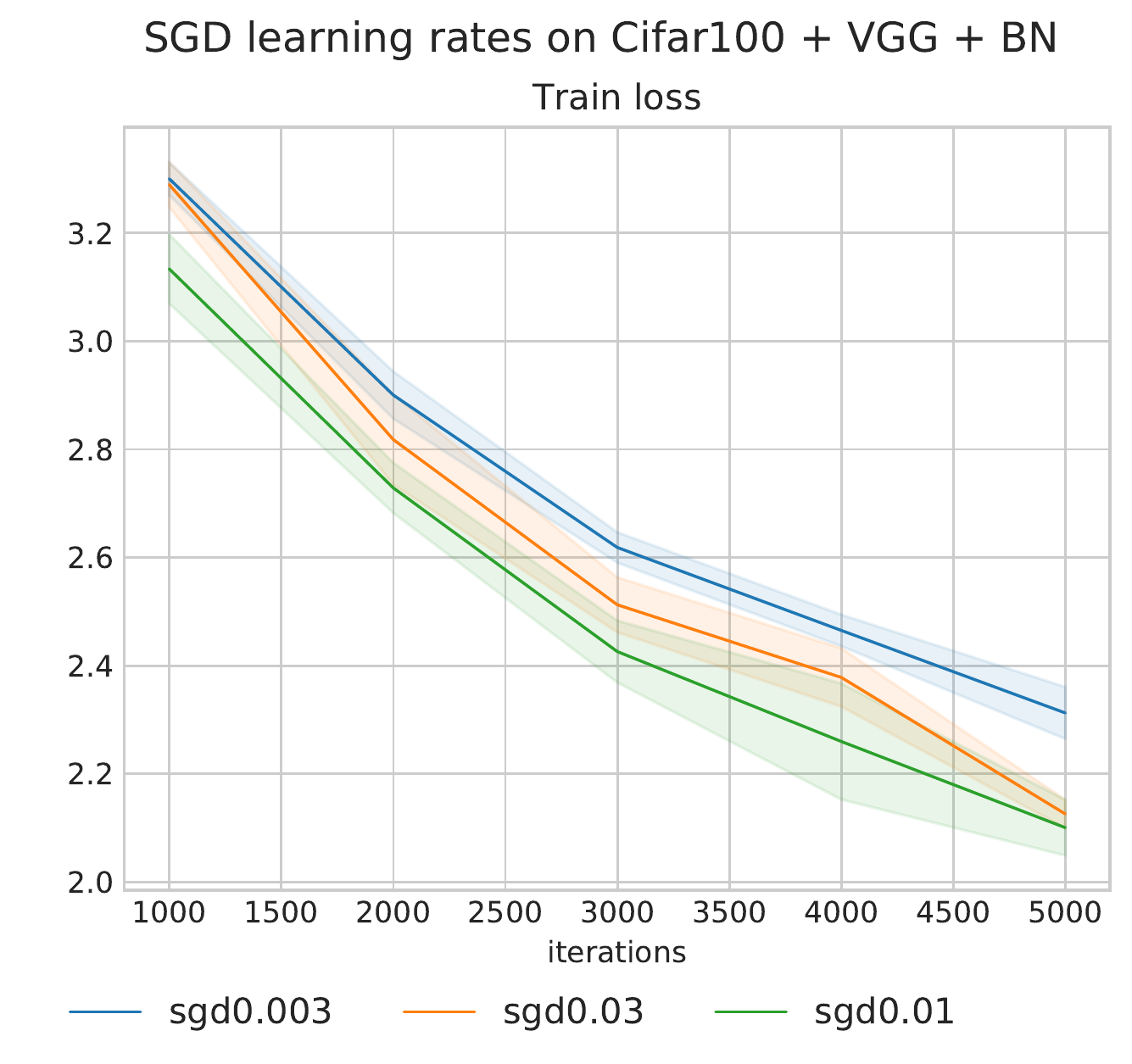}
\includegraphics[width=0.33\textwidth]{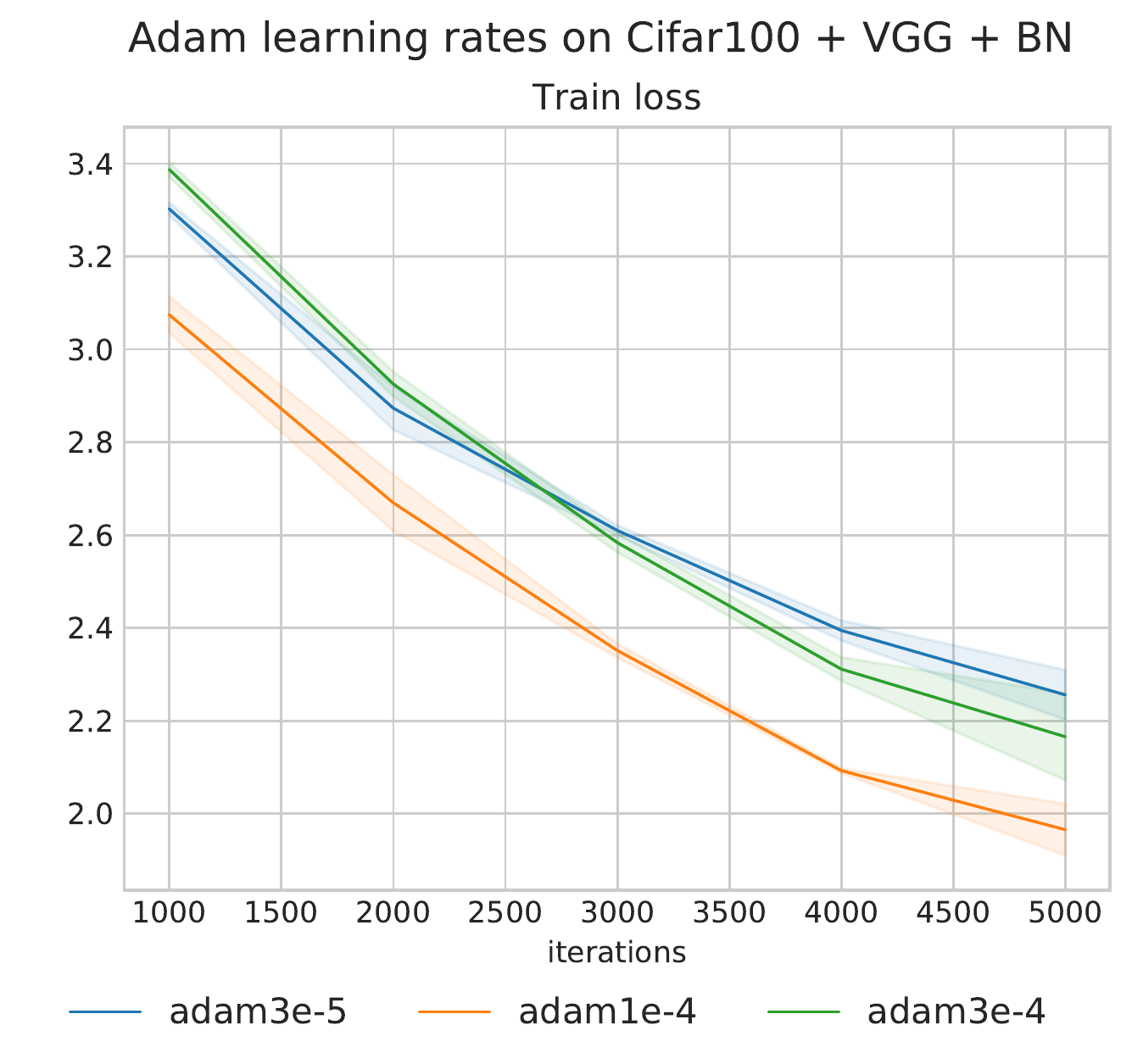}
\includegraphics[width=0.33\textwidth]{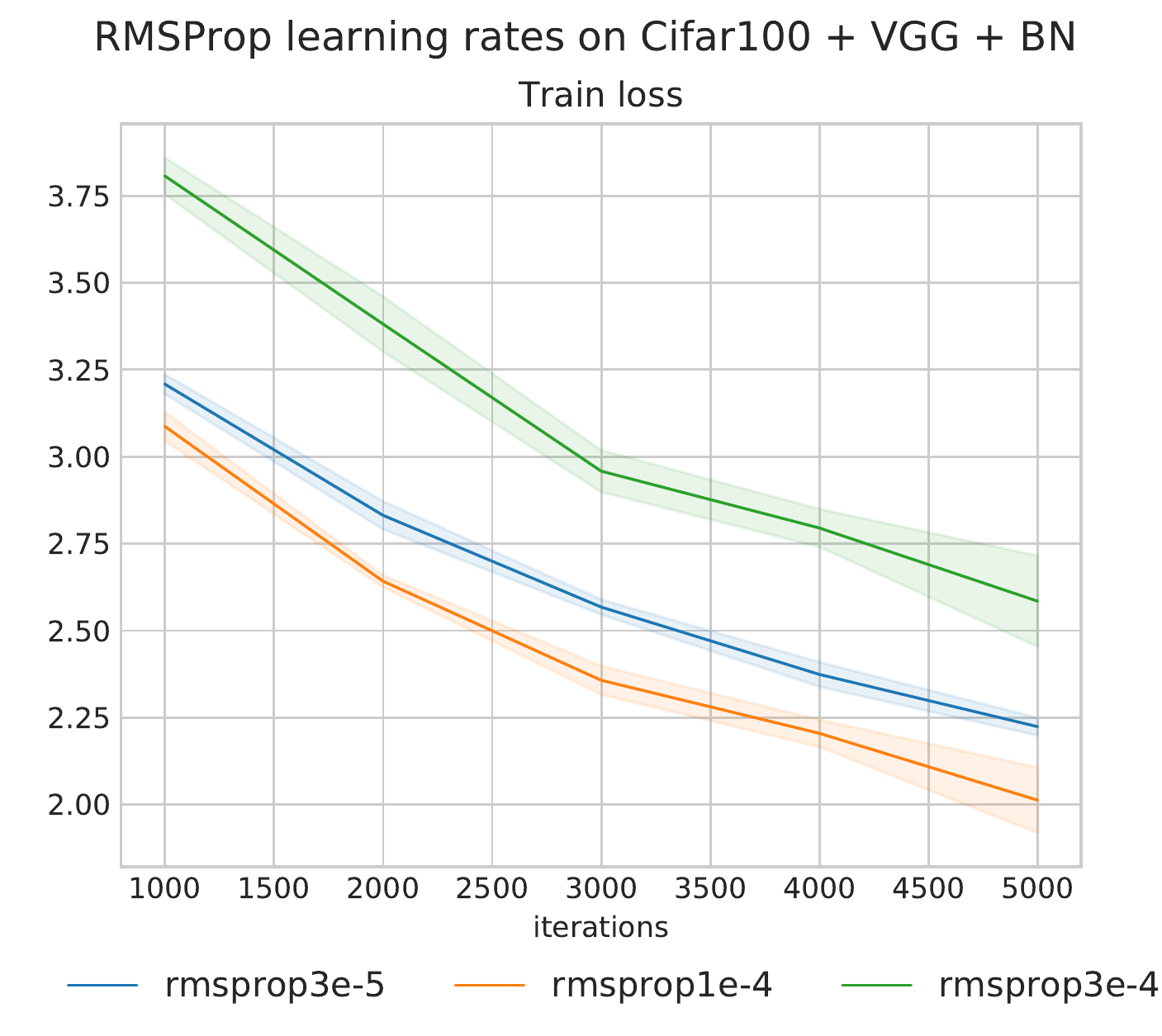}
}
\vspace{-0.15cm}
\caption{\label{fig:lr-cifar-vgg} Learning rate tuning results for Cifar100 with VGG11. We chose 0.01 for SGD, 1e-4 for Adam, and 1e-4 for RMSProp.}
\vspace{-0.15cm}
\end{figure}

\begin{figure}[t]
\centerline{
\includegraphics[width=0.33\textwidth]{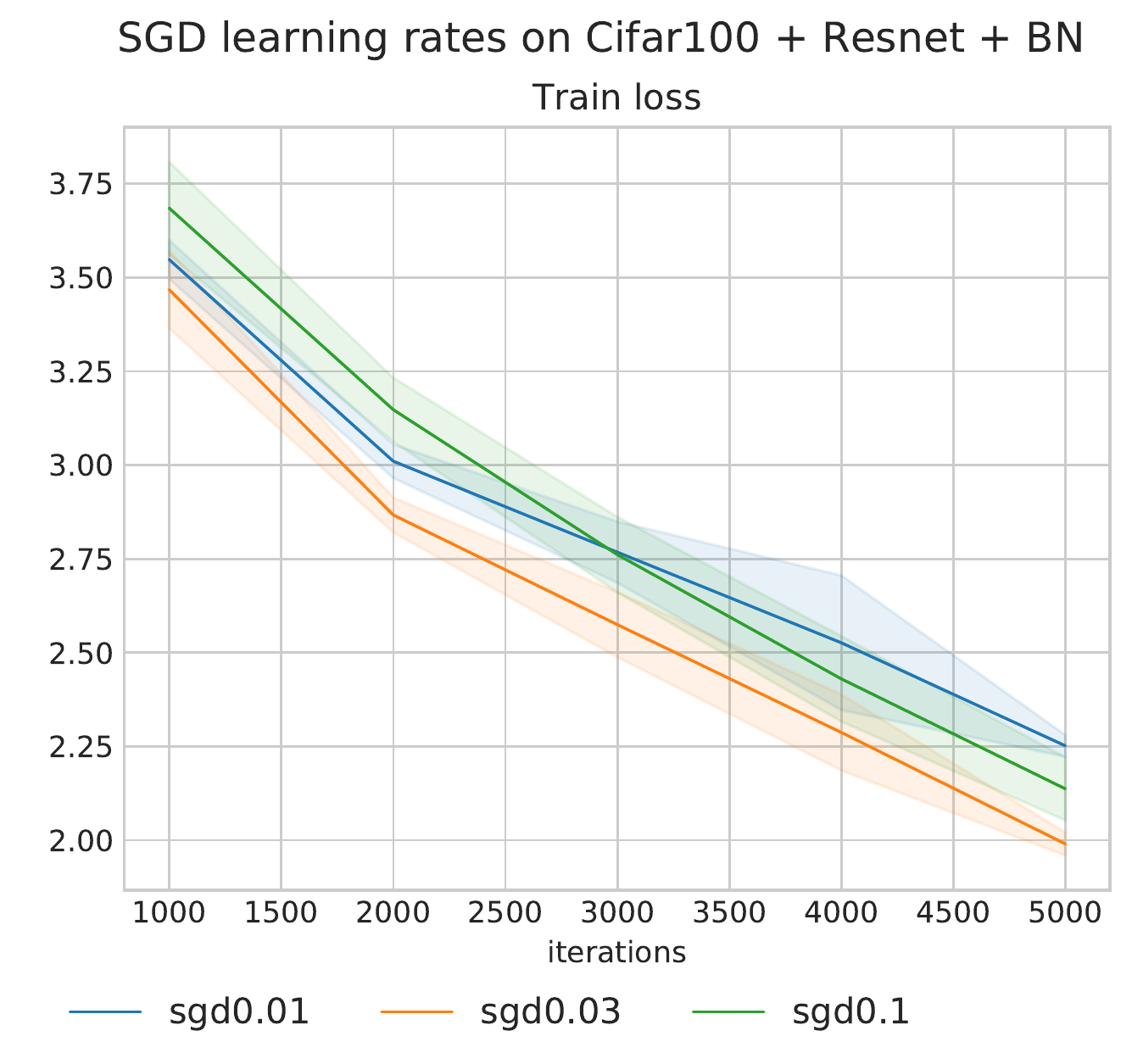}
\includegraphics[width=0.33\textwidth]{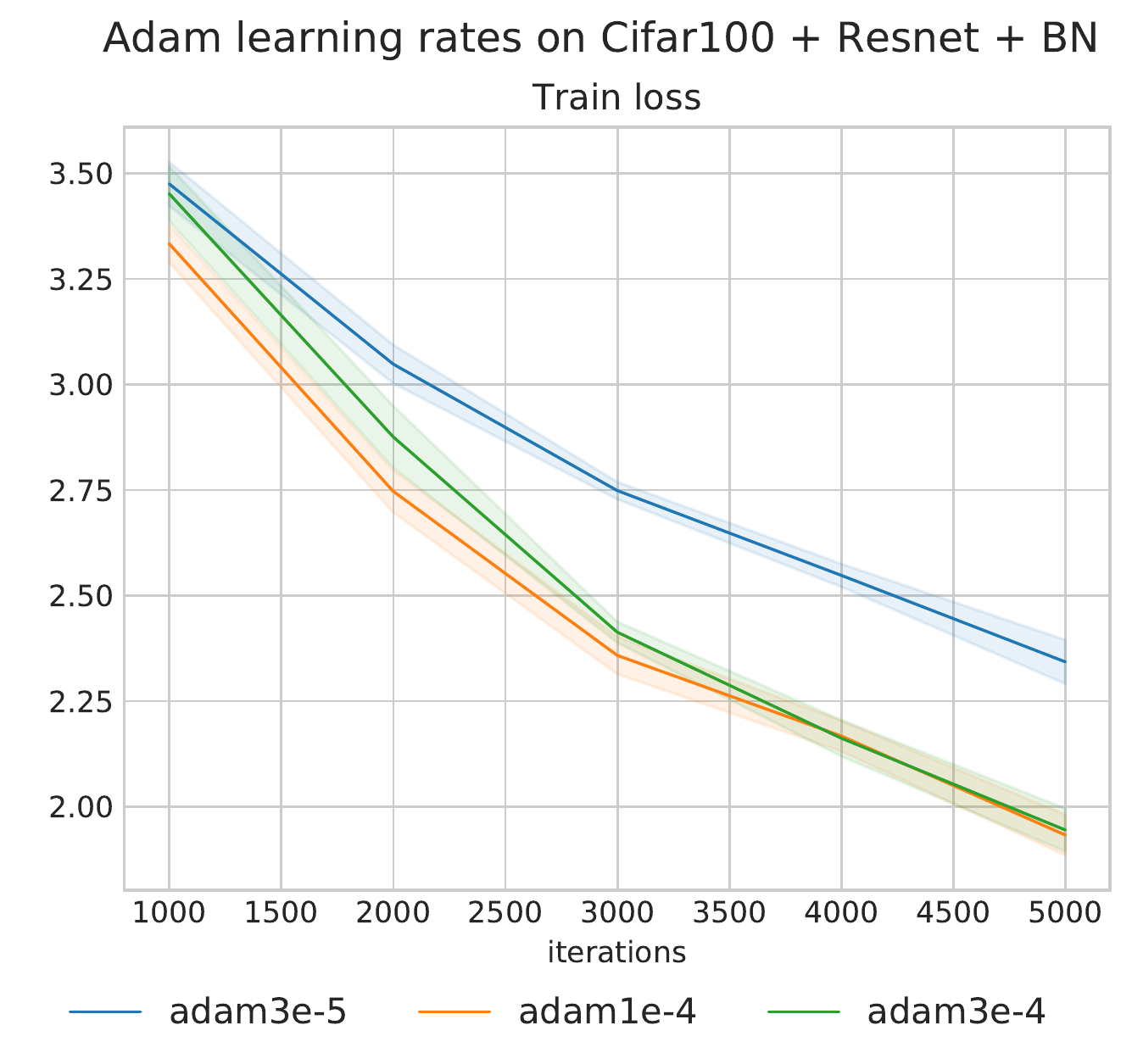}
\includegraphics[width=0.33\textwidth]{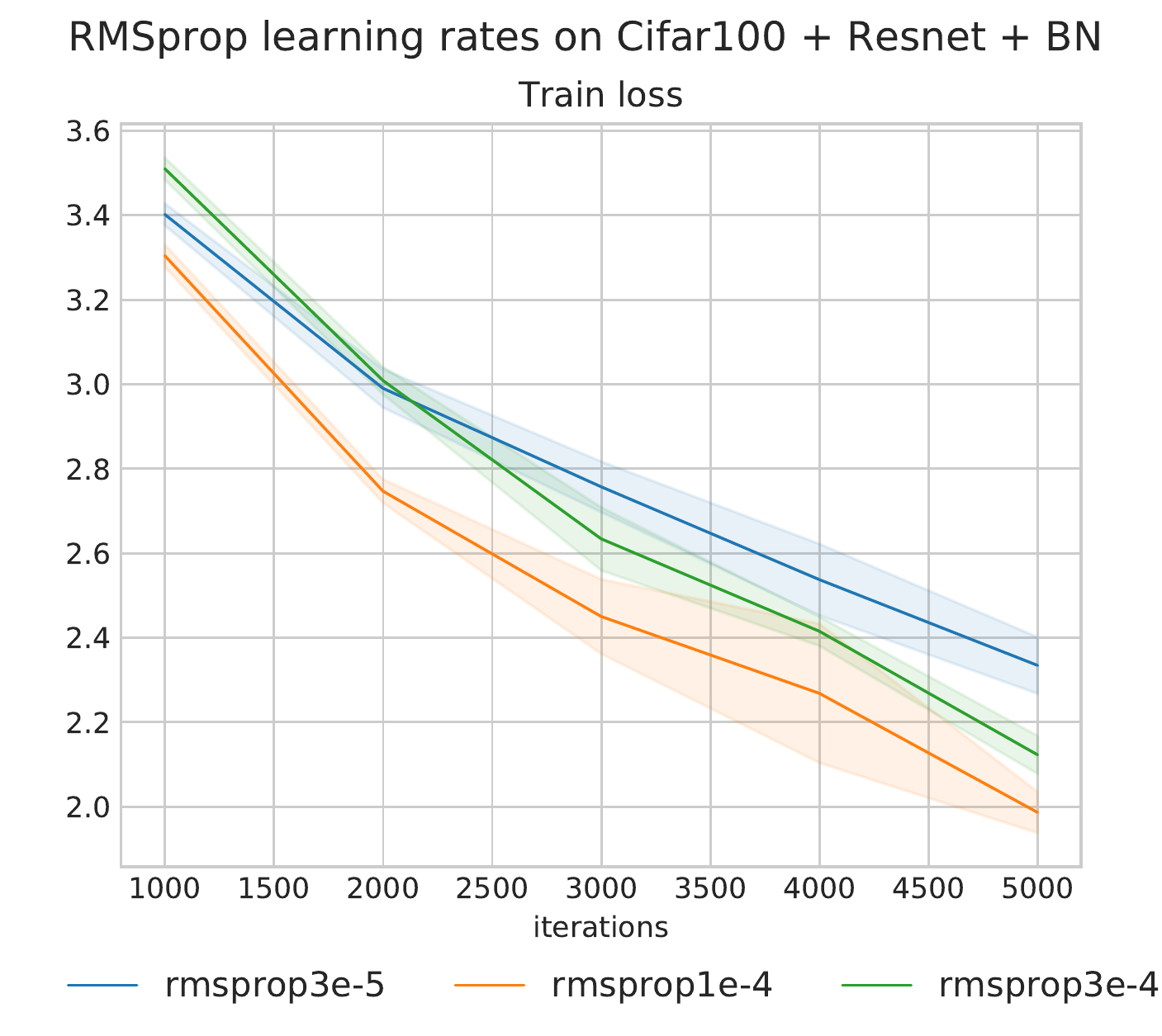}
}
\vspace{-0.15cm}
\caption{\label{fig:lr-cifar-resnet} Learning rate tuning results for Cifar100 with Resnet18. We chose 0.03 for SGD, 1e-4 for Adam, and 1e-4 for RMSProp.}
\vspace{-0.15cm}
\end{figure}

\textbf{Cifar100}: VGG11 + BN (see figure \ref{fig:lr-cifar-vgg}), Resnet18 + BN (see figure \ref{fig:lr-cifar-resnet}).

\begin{figure}[t]
\centerline{
\includegraphics[width=0.33\textwidth]{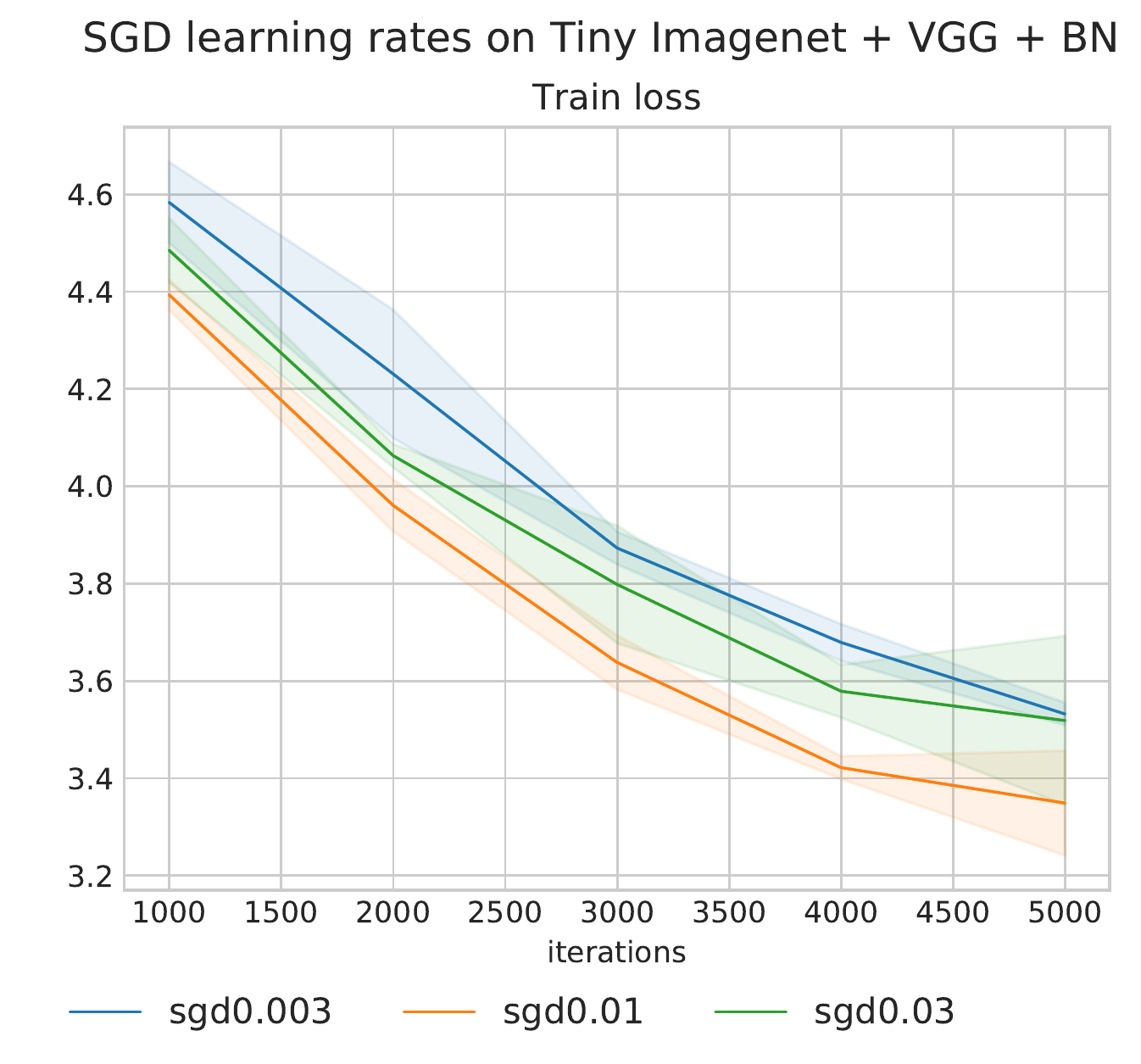}
\includegraphics[width=0.33\textwidth]{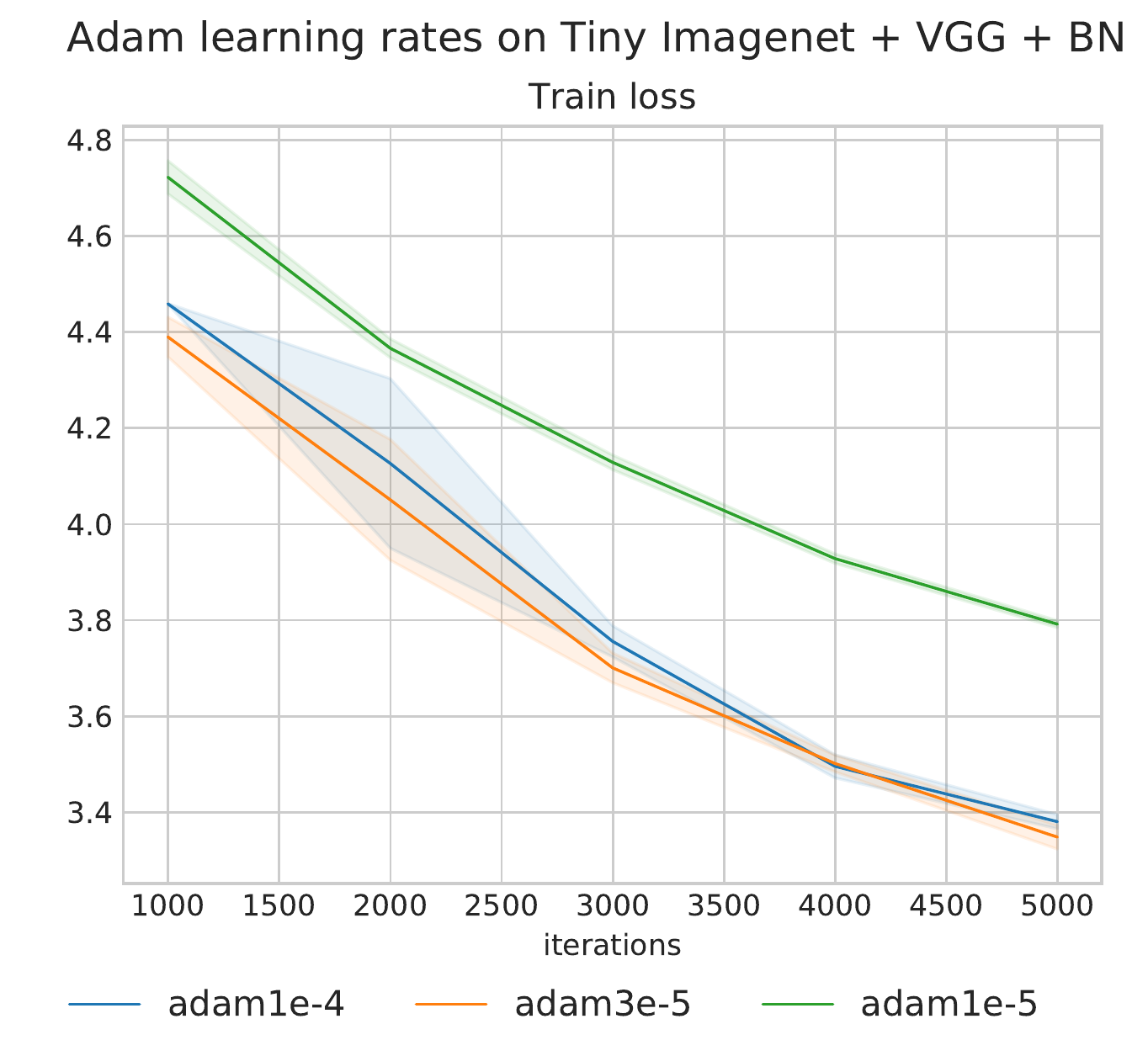}
}
\vspace{-0.15cm}
\caption{\label{fig:lr-tinyimagenet-vgg} Learning rate tuning results for Tiny Imagenet with VGG16. We chose 0.01 for SGD and 3e-5 for Adam.}
\vspace{-0.15cm}
\end{figure}

\begin{figure}[t]
\centerline{
\includegraphics[width=0.33\textwidth]{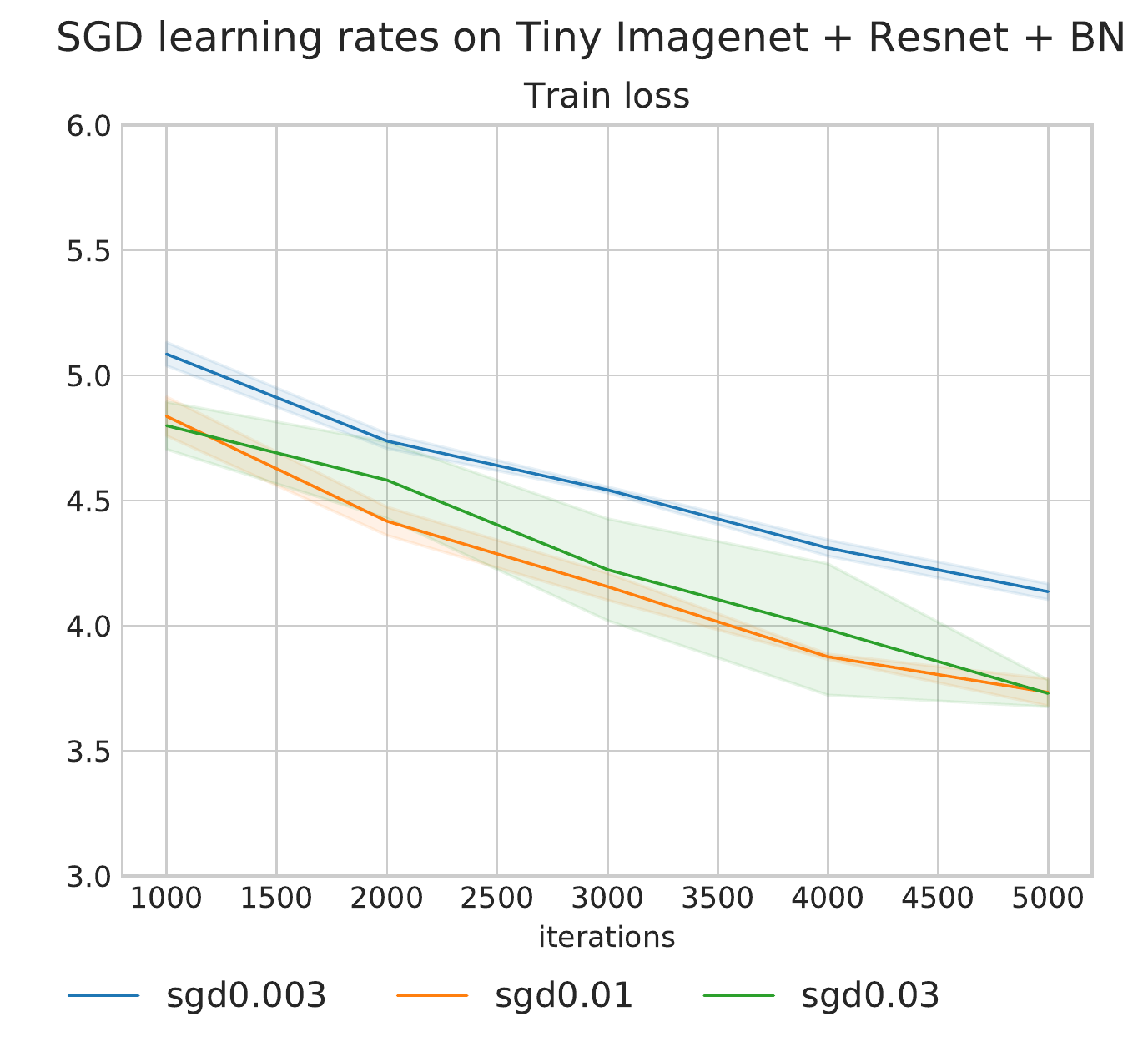}
\includegraphics[width=0.33\textwidth]{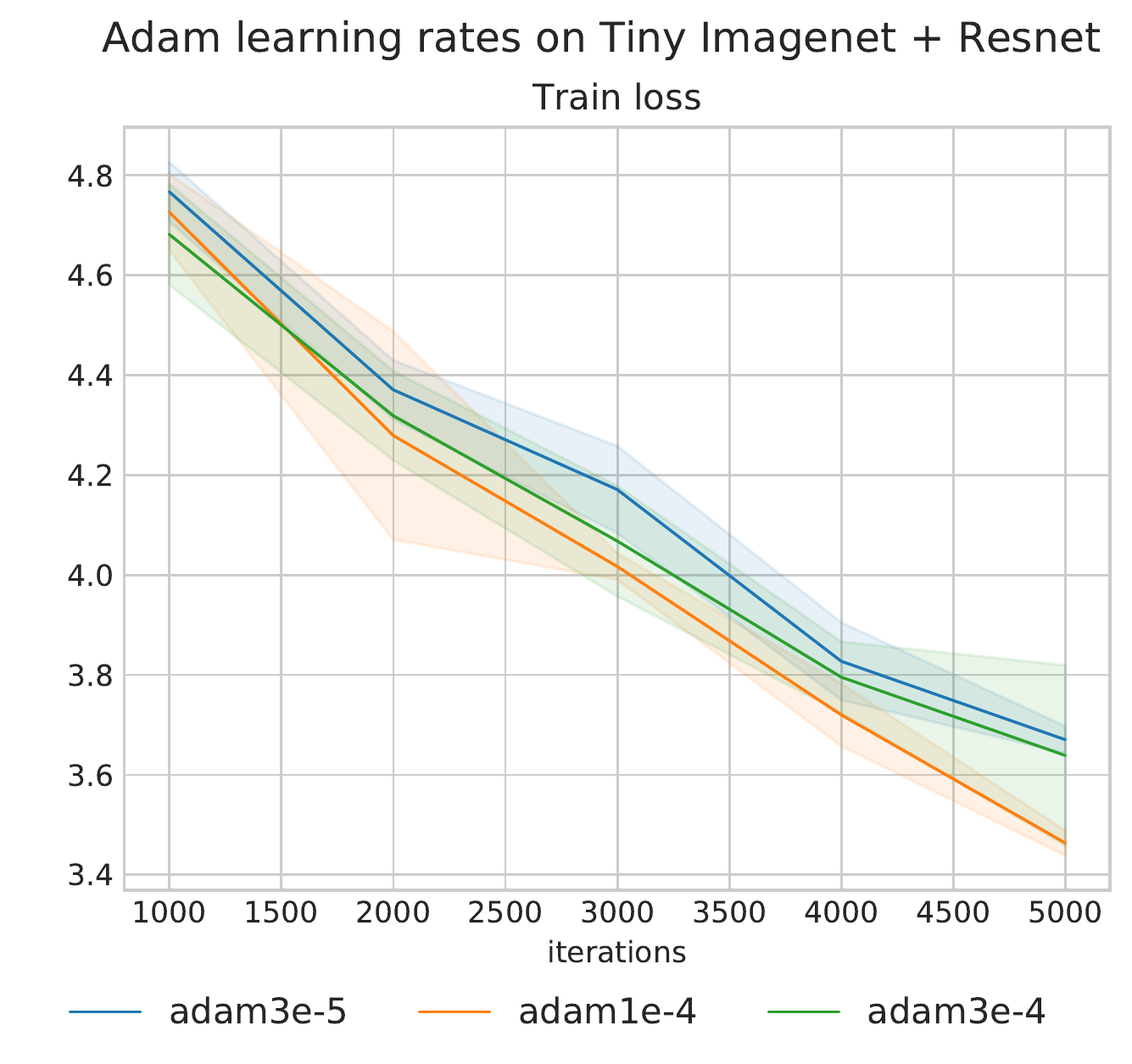}
}
\vspace{-0.15cm}
\caption{\label{fig:lr-tinyimagenet-resnet} Learning rate tuning results for Tiny Imagenet with Resnet50. We chose 0.01 for SGD and 1e-4 for Adam.}
\vspace{-0.15cm}
\end{figure}

\textbf{Tiny Imagenet}: VGG16 + BN (see figure \ref{fig:lr-tinyimagenet-vgg}), Resnet50 + BN (see figure \ref{fig:lr-tinyimagenet-resnet}).

\begin{figure}[t]
\centerline{
\includegraphics[width=0.33\textwidth]{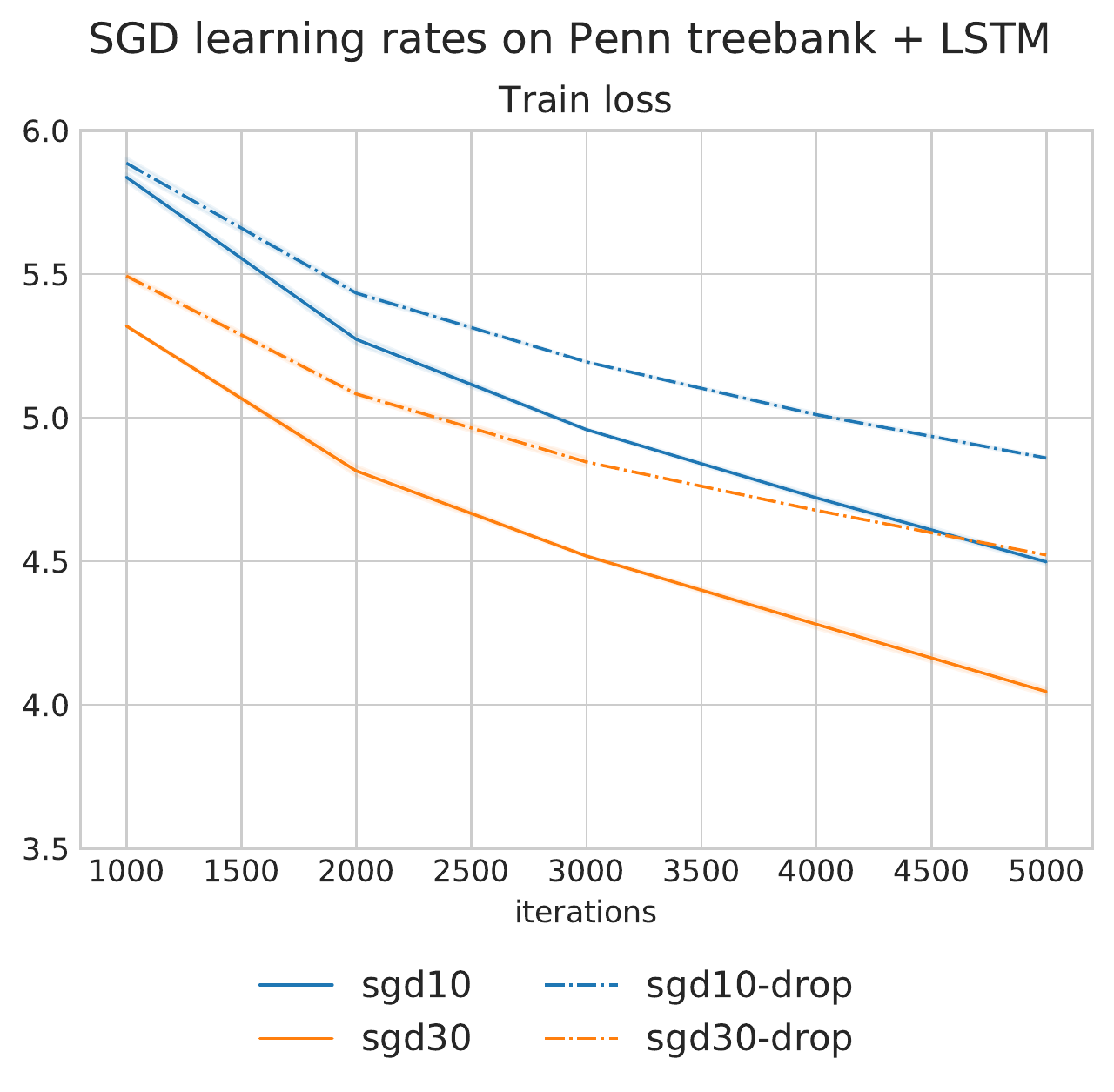}
\includegraphics[width=0.33\textwidth]{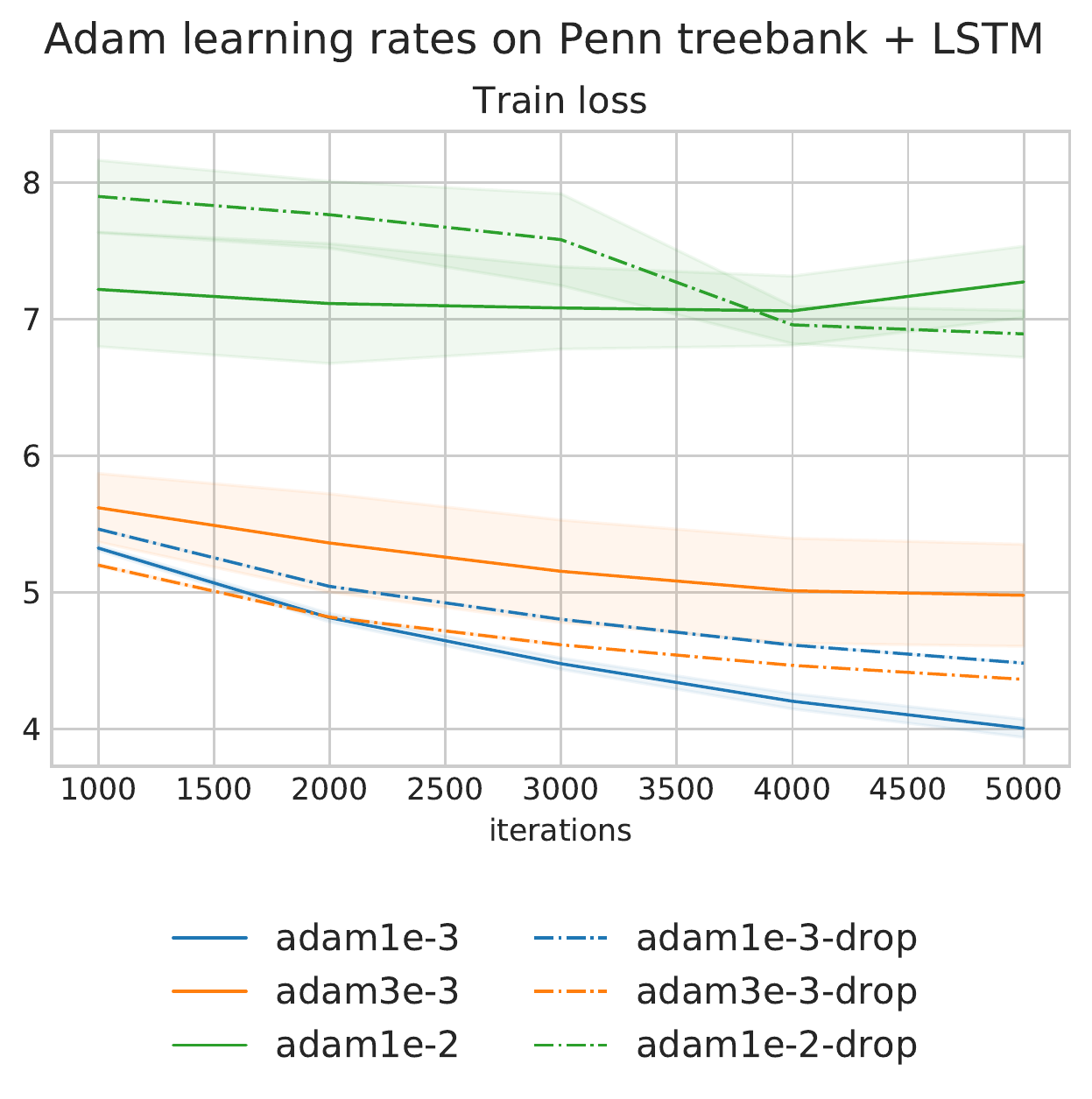}
}
\vspace{-0.15cm}
\caption{\label{fig:lr-ptb-lstm} Learning rate tuning results for Penn treebank with AWD-LSTM. For AWD-LTM wo. dropout, we chose 30 for SGD and 1e-3 for Adam. For AWD-LSTM w. dropout, we chose 30 for SGD and 3e-3 for Adam.}
\vspace{-0.15cm}
\end{figure}

\textbf{Penn treebank}: AWD-LSTM w. and wo. dropout (see figure \ref{fig:lr-ptb-lstm}).

% Fashion MNIST
% FC

% CNN

% CIFAR
% VGG

% Resnet

% Tiny imagenet
% VGG

% Resnet

% PTB with and without dropout

% Batch sizes Adam
\begin{figure*}[t]
\centerline{
\includegraphics[width=0.3\textwidth]{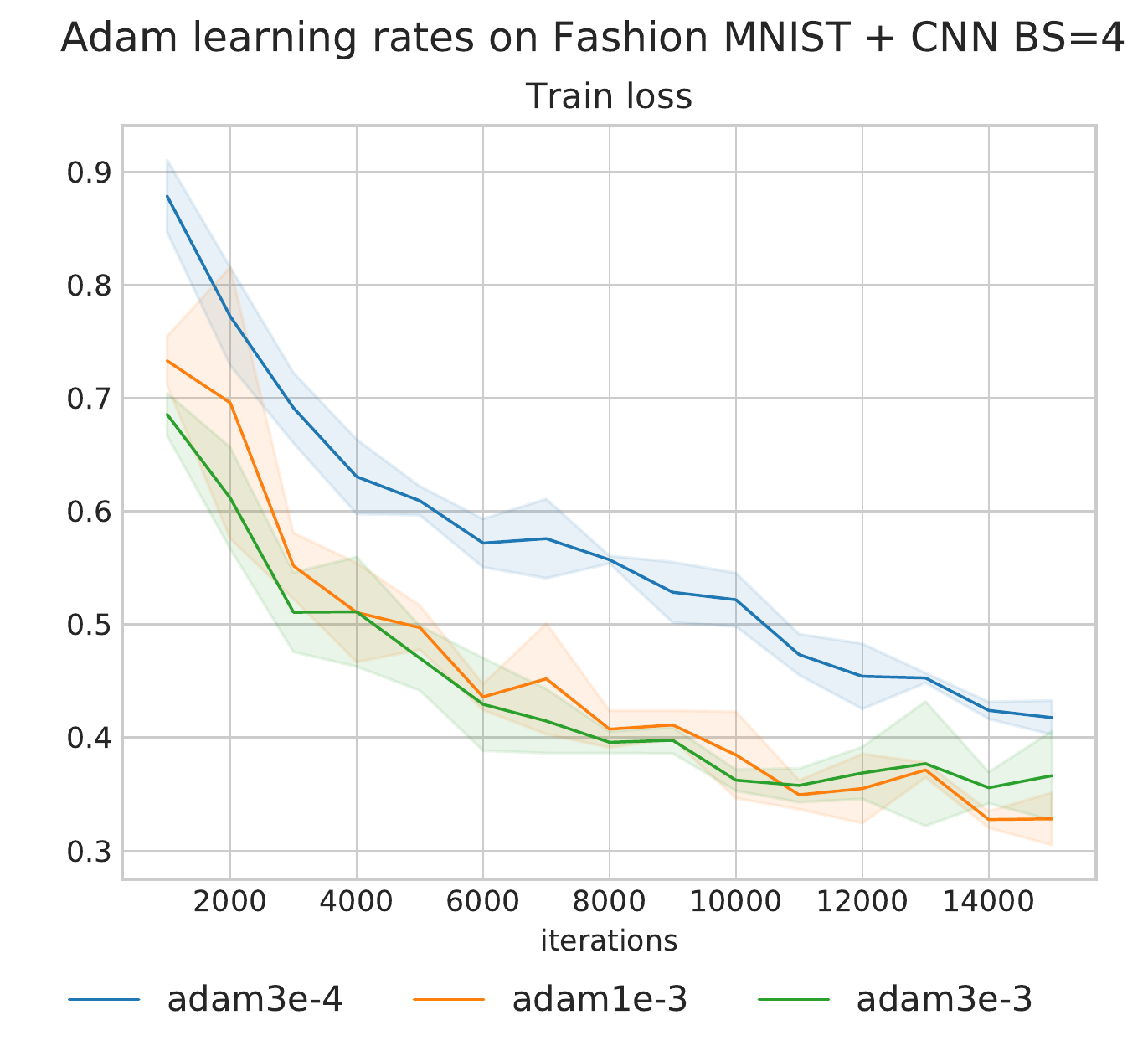}
\includegraphics[width=0.3\textwidth]{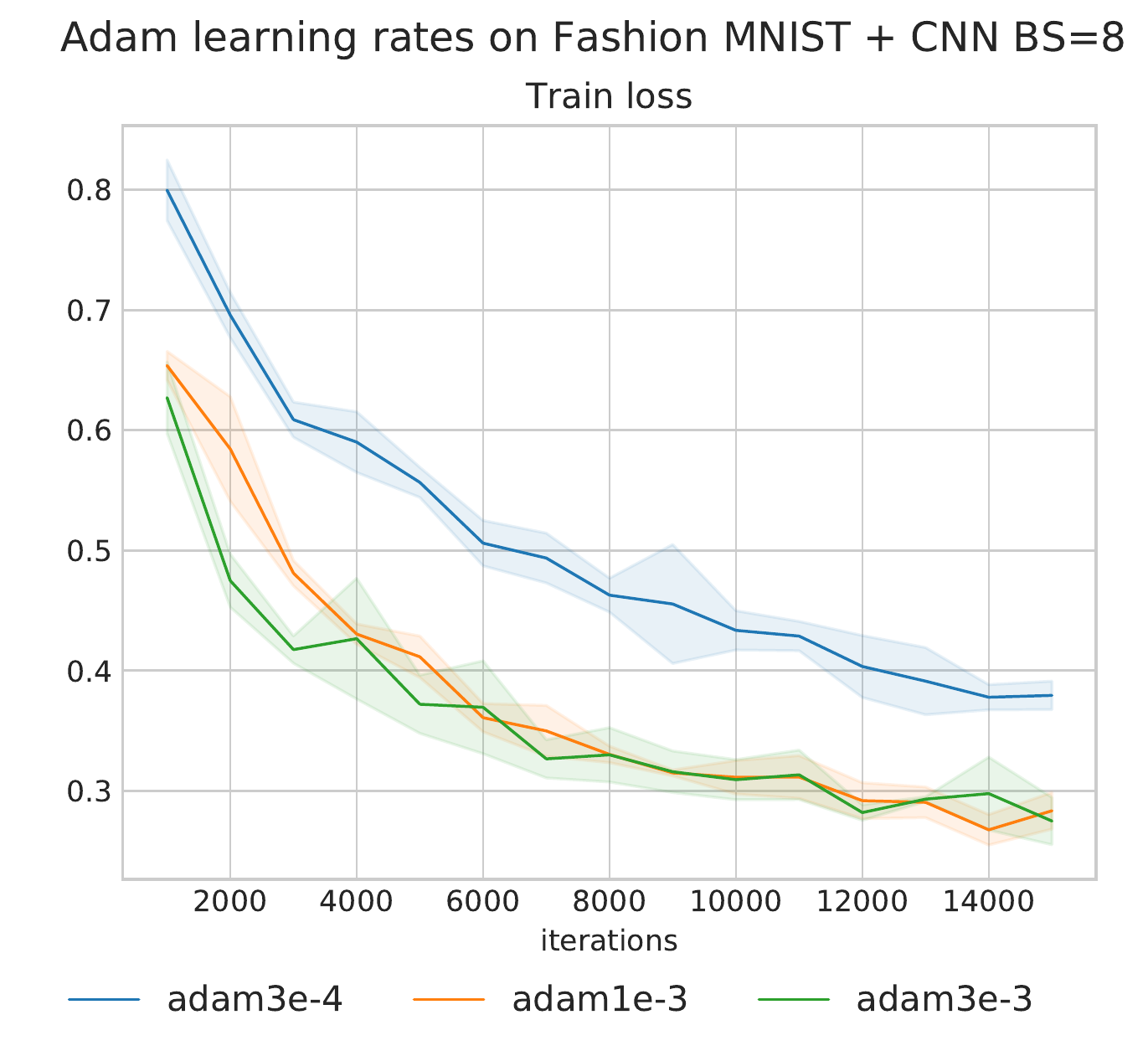}
\includegraphics[width=0.3\textwidth]{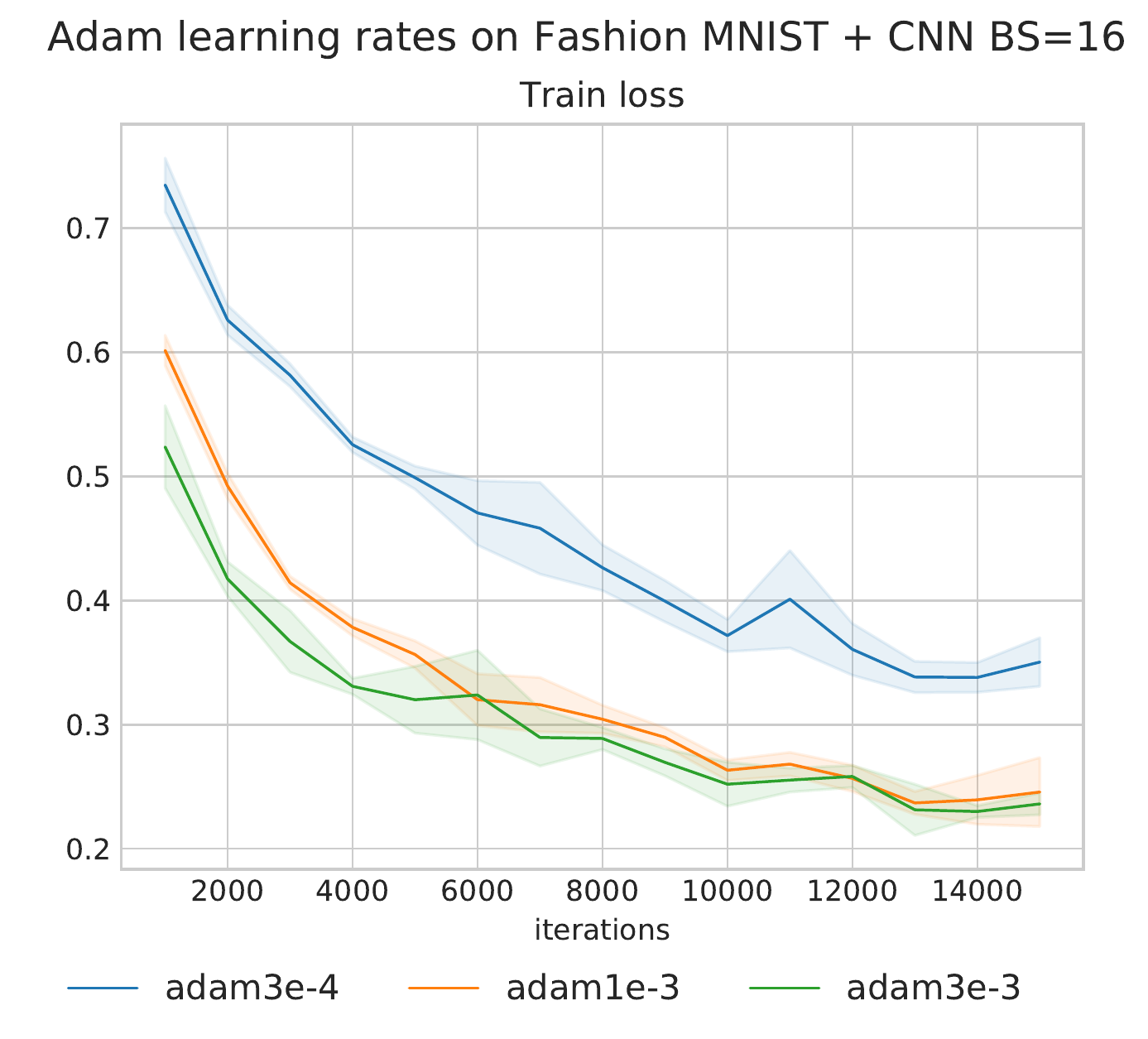}
}
\vspace{-0.15cm}
\caption{\label{fig:lr-smallbs-adam} Adam learning rate tuning results with different batch size for Fashion MNIST with CNN. We chose the learning rates of 1e-3 for batch sizes 4 and 8, and 3e-3 for batch sizes of 16, 32, 64.}
\vspace{-0.15cm}
\end{figure*}

\end{document}